%% file: slac_neurips2020.tex
\documentclass{article}

\PassOptionsToPackage{numbers, compress}{natbib}
\usepackage[final]{neurips_2020}

\usepackage[utf8]{inputenc} %
\usepackage[T1]{fontenc}    %
\usepackage{hyperref}       %
\usepackage{url}            %
\usepackage{booktabs}       %
\usepackage{amsfonts}       %
\usepackage{nicefrac}       %
\usepackage{microtype}      %

\usepackage{amsmath}
\usepackage{amssymb}
\usepackage{mathtools}
\usepackage{bm}
\usepackage{algorithm}
\usepackage{algorithmicx}
\usepackage[noend]{algpseudocode}
\usepackage{etoolbox}
\usepackage{tikz} %
\usetikzlibrary{positioning}  %
\usetikzlibrary{arrows}  %
\usepackage{xspace}
\usepackage{xpatch}
\usepackage{xstring}
\usepackage{graphicx}
\usepackage{enumitem}
\usepackage{tikzscale}
\usepackage{afterpage}
\usepackage{subcaption}
\usepackage{wrapfig}
\usepackage{multirow}
\usepackage{adjustbox}
\usepackage{makecell}
\usepackage{rotating}
\usepackage{arydshln}
\usepackage{standalone}
\usepackage[bottom]{footmisc} %

\usepackage[font=footnotesize,labelfont=bf]{caption}
\def\equationautorefname~#1\null{Equation~(#1)\null}

\makeatletter
\def\adl@drawiv#1#2#3{%
        \hskip.5\tabcolsep
        \xleaders#3{#2.5\@tempdimb #1{1}#2.5\@tempdimb}%
                #2\z@ plus1fil minus1fil\relax
        \hskip.5\tabcolsep}
\newcommand{\cdashlinelr}[1]{%
  \noalign{\vskip\aboverulesep
           \global\let\@dashdrawstore\adl@draw
           \global\let\adl@draw\adl@drawiv}
  \cdashline{#1}
  \noalign{\global\let\adl@draw\@dashdrawstore
           \vskip\belowrulesep}}
\makeatother

\makeatletter
\def\underbracex#1#2{\mathop{\vtop{\m@th\ialign{##\crcr
   $\hfil\displaystyle{#2}\hfil$\crcr
   \noalign{\kern3\p@\nointerlineskip}%
   #1\crcr\noalign{\kern3\p@}}}}\limits}

\def\upbracefilla{$\m@th \setbox\z@\hbox{$\braceld$}%
  \bracelu\leaders\vrule \@height\ht\z@ \@depth\z@\hfill 
\kern\p@\vrule \@width\p@\kern\p@\vrule \@width\p@\kern\p@\vrule \@width\p@
$}

\def\upbracefillb{$\m@th \setbox\z@\hbox{$\braceld$}%
\vrule \@width\p@\kern\p@\vrule \@width\p@\kern\p@\vrule \@width\p@\kern\p@
 \leaders\vrule \@height\ht\z@ \@depth\z@\hfill\bracerd
  \braceld\leaders\vrule \@height\ht\z@ \@depth\z@\hfill
\kern\p@\vrule \@width\p@\kern\p@\vrule \@width\p@\kern\p@\vrule \@width\p@
$}

\def\upbracefillc{$\m@th \setbox\z@\hbox{$\braceld$}%
\vrule \@width\p@\kern\p@\vrule \@width\p@\kern\p@\vrule \@width\p@\kern\p@
\leaders\vrule \@height\ht\z@ \@depth\z@\hfill
\kern\p@\vrule \@width\p@\kern\p@\vrule \@width\p@\kern\p@\vrule \@width\p@
$}

\def\upbracefilld{$\m@th \setbox\z@\hbox{$\braceld$}%
\vrule \@width\p@\kern\p@\vrule \@width\p@\kern\p@\vrule \@width\p@\kern\p@
 \leaders\vrule \@height\ht\z@ \@depth\z@\hfill\braceru$}

\def\upbracefillbd{$\m@th \setbox\z@\hbox{$\braceld$}%
\vrule \@width\p@\kern\p@\vrule \@width\p@\kern\p@\vrule \@width\p@\kern\p@
\bracerd\braceld
 \leaders\vrule \@height\ht\z@ \@depth\z@\hfill\braceru$}
\makeatother

\makeatletter
\patchcmd{\hyper@makecurrent}{%
    \ifx\Hy@param\Hy@chapterstring
        \let\Hy@param\Hy@chapapp
    \fi
}{%
    \iftoggle{inappendix}{%
        \@checkappendixparam{chapter}%
        \@checkappendixparam{section}%
        \@checkappendixparam{subsection}%
        \@checkappendixparam{subsubsection}%
        \@checkappendixparam{paragraph}%
        \@checkappendixparam{subparagraph}%
    }{}%
}{}{\errmessage{failed to patch}}

\newcommand*{\@checkappendixparam}[1]{%
    \def\@checkappendixparamtmp{#1}%
    \ifx\Hy@param\@checkappendixparamtmp
        \let\Hy@param\Hy@appendixstring
    \fi
}
\makeatletter
\def\blfootnote{\xdef\@thefnmark{}\@footnotetext}
\makeatother

\newtoggle{inappendix}
\togglefalse{inappendix}

\apptocmd{\appendix}{\toggletrue{inappendix}}{}{\errmessage{failed to patch}}
\input{variables}

\title{Stochastic Latent Actor-Critic: Deep Reinforcement Learning with a Latent Variable Model}

\newcommand{\berkeley}{University of California, Berkeley}
\newcommand{\deepmind}{DeepMind}
\newcommand{\berkeleysym}{1}
\newcommand{\deepmindsym}{2}

\author{%
  Alex X. Lee\textsuperscript{\textnormal{\berkeleysym{},\deepmindsym{}}} ~~~~~~
  Anusha Nagabandi\textsuperscript{\textnormal{\berkeleysym{}}} ~~~~~~
  Pieter Abbeel\textsuperscript{\textnormal{\berkeleysym{}}} ~~~~~~
  Sergey Levine\textsuperscript{\textnormal{\berkeleysym{}}} \\
  \textnormal{\textsuperscript{\berkeleysym{}}\berkeley{}} \\
  \textnormal{\textsuperscript{\deepmindsym{}}\deepmind{}} \\
  \texttt{\{alexlee\_gk,nagaban2,pabbeel,svlevine\}@cs.berkeley.edu} \\
}

\begin{document}

\maketitle

\begin{abstract}
  Deep reinforcement learning (RL) algorithms can use high-capacity deep networks to learn directly from image observations. However, these high-dimensional observation spaces present a number of  challenges in practice, since the policy must now solve two problems: representation learning and task learning. In this work, we tackle these two problems separately, by explicitly learning latent representations that can accelerate reinforcement learning from images. We propose the stochastic latent actor-critic (SLAC) algorithm: a sample-efficient and high-performing RL algorithm for learning policies for complex continuous control tasks directly from high-dimensional image inputs. SLAC provides a novel and principled approach for unifying stochastic sequential models and RL into a single method, by learning a compact latent representation and then performing RL in the model's learned latent space. Our experimental evaluation demonstrates that our method outperforms both model-free and model-based alternatives in terms of final performance and sample efficiency, on a range of difficult image-based control tasks. Our code and videos of our results are available at our website.\footnote{\url{https://alexlee-gk.github.io/slac/}}
\end{abstract}

\input{00_intro}
\input{01_related}

\input{02_preliminaries}
\input{03_approach}
\input{04_experiments}

\input{05_conclusion}

\input{06_broader_impact}

\begin{ack}
We thank Marvin Zhang, Abhishek Gupta, and Chelsea Finn for useful discussions and feedback, Danijar Hafner for providing timely assistance with PlaNet, and Maximilian Igl for providing timely assistance with DVRL.
This research was supported by the National Science Foundation through IIS-1651843 and IIS-1700697, as well as ARL DCIST CRA W911NF-17-2-0181 and the Office of Naval Research. Compute support was provided by NVIDIA.
\end{ack}

 {\small
 \bibliographystyle{abbrvnat}
 \bibliography{slac_neurips2020}
 }

\newpage
\appendix

\input{07_appendix}

\end{document}

%% file: variables.tex
\newcommand*{\mat}[1]{\ensuremath{\bm{#1}}}
\renewcommand*{\vec}[1]{\ensuremath{\boldsymbol{\mathbf{#1}}}}

\newcommand*{\E}{\ensuremath{\mathop{\mathbb{E}}}}
\newcommand*\diff{\mathop{}\!\mathrm{d}}  

\newcommand*{\s}[1]{\ensuremath{\vec{s}_{#1}}}
\renewcommand*{\a}[1]{\ensuremath{\vec{a}_{#1}}}
\renewcommand*{\r}[1]{\ensuremath{r_{#1}}}

\newcommand*{\z}[1]{\ensuremath{\vec{z}_{#1}}}
\newcommand*{\x}[1]{\ensuremath{\vec{x}_{#1}}}
\newcommand*{\optim}[1]{\ensuremath{\mathcal{O}_{#1}}}

\renewcommand*{\S}{\ensuremath{\mathcal{S}}}
\newcommand*{\A}{\ensuremath{\mathcal{A}}}

\newcommand*{\Z}{\ensuremath{\mathcal{Z}}}
\newcommand*{\X}{\ensuremath{\mathcal{X}}}

\newcommand*{\policy}{\ensuremath{\pi}}
\newcommand*{\Qparams}{\ensuremath{\theta}}  
\newcommand*{\pparams}{\ensuremath{\phi}}  
\newcommand*{\mparams}{\ensuremath{\psi}}  

\newcommand{\kl}[2]{\ensuremath{\mathrm{D_{\textsc{kl}}}\!\left(#1\;\middle\|\;#2\right)}}
\newcommand{\klcompact}[2]{\mathrm{D_{\textsc{kl}}}\!\left(#1\middle\|#2\right)}
\newcommand{\entropy}{\ensuremath{\mathcal{H}}}

\newcommand{\visits}{\ensuremath{\rho}}

%% file: 00_intro.tex
\section{Introduction}
Deep reinforcement learning (RL) algorithms can learn to solve tasks directly from raw, low-level observations such as images. However, such high-dimensional observation spaces present a number of challenges in practice: On one hand, it is difficult to directly learn from these high-dimensional inputs, but on the other hand, it is also difficult to tease out a compact representation of the underlying task-relevant information from which to learn instead. Standard model-free deep RL aims to unify these challenges of representation learning and task learning into a single end-to-end training procedure. However, solving \emph{both} problems together is difficult, since an effective policy requires an effective representation, and an effective representation requires meaningful gradient information to come from the policy or value function, while using only the model-free supervision signal (i.e., the reward function). As a result, learning directly from images with standard end-to-end RL algorithms can in practice be slow, sensitive to hyperparameters, and inefficient.

Instead, we propose to separate representation learning and task learning, by relying on predictive model learning to explicitly acquire a latent representation, and training the RL agent \emph{in that learned latent space}. This alleviates the representation learning challenge because predictive learning benefits from a rich and informative supervision signal even before the agent has made any progress on the task, and thus results in improved sample efficiency of the overall learning process. In this work, our predictive model serves to accelerate task learning by separately addressing representation learning, in contrast to existing model-based RL approaches, which use predictive models either for generating cheap synthetic experience~\citep{sutton1991dyna, gu2016continuous,janner2019mbpo} or for planning into the future~\citep{deisenroth2011pilco, finn2017deep, nagabandi2018neural, chua2018deep, zhang2019solar, hafner2019learning}.

Our proposed stochastic sequential model (\autoref{fig:pgm}) models the high-dimensional observations as the consequence of a latent process, with a Gaussian prior and latent dynamics. This model represents a partially observed Markov decision process (POMDP), where the stochastic latent state enables the model to represent uncertainty about any of the state variables, given the past observations. Solving such a POMDP exactly would be computationally intractable, since it amounts to solving the decision problem in the space of \emph{beliefs}~\citep{astrom1965optimal, kaelbling1998planning}. Recent works approximate the belief as encodings of latent samples from forward rollouts or particle filtering~\citep{buesing2018learning,igl2018deep}, or as learned belief representations in a belief-state forward model~\citep{gregor2019shaping}.
We instead propose a simple approximation, which we derive from the control as inference framework, that trains a Markovian critic on latent state samples and trains an actor on a history of observations and actions, resulting in our stochastic latent actor-critic (SLAC) algorithm. Although this approximation loses some of the benefits of full POMDP solvers (e.g. reducing uncertainty), it is easy and stable to train in practice, achieving competitive results on a range of challenging problems.

The main contribution of this work is a novel and principled approach that integrates learning stochastic sequential models and RL into a single method, performing RL in the model's learned latent space. By formalizing the problem as a control as inference problem within a POMDP, we show that variational inference leads to the objective of our SLAC algorithm. %
We empirically show that SLAC benefits from the good asymptotic performance of model-free RL while also leveraging the improved latent space representation for sample efficiency, by demonstrating that SLAC substantially outperforms both prior model-free and model-based RL algorithms on a range of image-based continuous control benchmark tasks.

%% file: 01_related.tex
\section{Related Work}

\textbf{Representation learning in RL.} End-to-end deep RL can in principle learn representations implicitly as part of the RL process~\citep{mnih2013dqn}. However, prior work has observed that RL has a ``representation learning bottleneck'': a considerable portion of the learning period must be spent acquiring good representations of the observation space~\citep{shelhamer2016loss}. This motivates the use of a distinct representation learning procedure to acquire these representations before the agent has even learned to solve the task. A number of prior works have explored the use of auxiliary supervision in RL to learn such representations~\citep{lange2010deep,finn2016deep,jaderberg2016auxiliary,higgins2017darla,ha2018world,nair2018visual,oord2018cpc,gelada2019deepmdp,dadashi2019polytope}. In contrast to this class of representation learning algorithms, we explicitly learn a latent variable model of the POMDP, in which the latent representation and latent-space dynamics are jointly learned. By modeling covariances between consecutive latent states, we make it feasible for our proposed algorithm to perform Bellman backups directly in the latent space of the learned model.

\textbf{Partial observability in RL.} Our work is also related to prior research on RL under partial observability.
Prior work has studied exact and approximate solutions to POMDPs, but they require explicit models of the POMDP and are only practical for simpler domains~\citep{kaelbling1998planning}.
Recent work has proposed end-to-end RL methods that use recurrent neural networks to process histories of observations and (sometimes) actions, but without constructing a model of the POMDP~\citep{hausknecht2015drqn, foerster2016ddqrn, zhu2018adrqn}.
Other works, however, learn latent-space dynamical system models and then use them to solve the POMDP with model-based RL~\citep{watter2015embed, wahlstrom2015pixels, karl2017deep, karl2017unsupervised, zhang2019solar, hafner2019learning, kim2019vta}.
Although some of these works learn latent variable models that are similar to ours, these methods are often limited by compounding model errors and finite horizon optimization.
In contrast to these works, our approach does not use the model for prediction, and performs infinite horizon policy optimization. Our approach benefits from the good asymptotic performance of model-free RL, while at the same time leveraging the improved latent space representation for sample efficiency.

Other works have also trained latent variable models and used their representations as the inputs to model-free RL algorithms.
They use representations encoded from latent states sampled from the forward model~\citep{buesing2018learning}, belief representations obtained from particle filtering~\citep{igl2018deep}, or belief representations obtained directly from a learned belief-space forward model~\citep{gregor2019shaping}.
Our approach is closely related to these prior methods, in that we also use model-free RL with a latent state representation that is learned via prediction.
However, instead of using belief representations, our method learns a critic directly on latent state samples, which more tractably enables scaling to more complex tasks.
Concurrent to our work, \citet{hafner2020dreamer} proposed to integrate model-free learning with representations from sequence models, as proposed in this paper, with model-based rollouts, further improving on the performance of prior model-based approaches.

\textbf{Sequential latent variable models.}
Several previous works have explored various modeling choices to learn stochastic sequential models~\citep{krishnan2015dkf,archer2015blackbox,karl2017deep,fraccaro2016srnn,fraccaro2017disentangled,doerr2018prssm,gregor2019tdvae}. They vary in the factorization of the generative and inference models, their network architectures, and the objectives used in their training procedures. Our approach is compatible with any of these sequential latent variable models, with the only requirement being that they provide a mechanism to sample latent states from the belief of the learned Markovian latent space.

%% file: 02_preliminaries.tex
\vspace{-1mm}
\section{Preliminaries}
\vspace{-1.5mm}

This work addresses the problem of learning policies from high-dimensional observations in POMDPs, by simultaneously learning a latent representation of the underlying MDP state using variational inference, as well as learning a policy in a maximum entropy RL framework. In this section, we describe maximum entropy RL~\citep{ziebart2010modeling,haarnoja2018soft,levine2018reinforcement} in fully observable MDPs, as well as variational methods for training latent state space models for POMDPs.

\vspace{-1.5mm}
\subsection{Maximum Entropy RL in Fully Observable MDPs}
\vspace{-2mm}
\label{sec:maxent}

Consider a Markov decision process (MDP), with states $\s{t} \in \S$, actions $\a{t} \in \A$, rewards $\r{t}$, initial state distribution $p(\s{1})$, and stochastic transition distribution $p(\s{t+1} | \s{t}, \a{t})$.
Standard RL aims to learn the parameters $\pparams$ of some policy $\policy_\pparams(\a{t} | \s{t})$ such that the expected sum of rewards is maximized under the induced trajectory distribution $\visits_\policy$. This objective can be modified to incorporate an \emph{entropy} term, such that the policy also aims to maximize the expected entropy $\entropy{(\policy_\pparams(\cdot|\s{t}))}$. This formulation has a close connection to variational inference~\citep{ziebart2010modeling,haarnoja2018soft,levine2018reinforcement}, and we build on this in our work. The resulting maximum entropy objective is
$\sum_{t=1}^T \E_{(\s{t}, \a{t}) \sim \visits_\policy}[r(\s{t}, \a{t}) + \alpha \entropy{(\policy_\pparams(\cdot | \s{t}))}]$,
where $r$ is the reward function, and $\alpha$ is a temperature parameter that trades off between maximizing for the reward and for the policy entropy. Soft actor-critic (SAC)~\citep{haarnoja2018soft} uses this maximum entropy RL framework to derive soft policy iteration, which alternates between policy evaluation and policy improvement within the described maximum entropy framework. SAC then extends this soft policy iteration to handle continuous action spaces by using parameterized function approximators to represent both the Q-function $Q_\Qparams$ (critic) and the policy $\policy_\pparams$ (actor). The soft Q-function parameters $\Qparams$ are optimized to minimize the soft Bellman residual,
\vspace{-2mm}
\begin{align}
  J_Q(\Qparams) &= \tfrac{1}{2} \left( Q_\Qparams(\s{t}, \a{t}) - \left( \vphantom{\sum} \right. \right. \r{t} + \gamma \E_{\a{t+1} \sim \policy_\pparams} \left[ Q_{\bar{\Qparams}}(\s{t+1}, \a{t+1}) - \alpha \log \policy_\pparams(\a{t+1} | \s{t+1}) \right] \left. \left. \vphantom{\sum} \right) \right)^2,
\end{align}\\[-2.5mm]
where $\gamma$ is the discount factor, and $\bar{\Qparams}$ are delayed parameters. The policy parameters $\pparams$ are optimized to update the policy towards the exponential of the soft Q-function, resulting in the policy loss
\vspace{-1mm}
\begin{equation}
  J_\policy(\pparams) = \E_{\a{t} \sim \policy_\pparams} \left[ \alpha \log (\policy_\pparams(\a{t} | \s{t})) - Q_\Qparams(\s{t}, \a{t}) \right].
  \label{eq:sac_policy}
\end{equation}\\[-3.5mm]
SLAC builds on top of this maximum entropy RL framework, by further integrating explicit representation learning and handling partial observability.

\vspace{-1.5mm}
\subsection{Sequential Latent Variable Models and Amortized Variational Inference in POMDPs}
\vspace{-2mm}

To learn representations for RL, we use latent variable models trained with amortized variational inference. The learned model must be able to process a large number of pixels that are present in the entangled image $\x{}$, and it must tease out the relevant information into a compact and disentangled representation $\z{}$. To learn such a model, we can consider maximizing the probability of each observed datapoint $\x{}$ from some training set under the entire generative process $p(\x{})= \int p(\x{}|\z{})p(\z{}) \diff \z{}$. This objective is intractable to compute in general due to the marginalization of the latent variables $\z{}$. In amortized variational inference, we utilize the evidence lower bound for the log-likelihood~\citep{kingma2013auto}:
\vspace{-1mm}
\begin{equation}
  \log p(\x{}) \geq E_{\z{}\sim q}\left[\log p(\x{}|\z{})  \right] - \kl{q(\z{}|\x{})}{p(\z{})}.
\label{eqn:vae}
\end{equation}\\[-4.5mm]
We can maximize the probability of the observed datapoints (i.e., the left hand side of \autoref{eqn:vae}) by learning an encoder $q(\z{}|\x{})$ and a decoder $p(\x{}|\z{})$, and then directly performing gradient ascent on the right hand side of the equation. In this setup, the distributions of interest are the prior $p(\z{})$, the observation model $p(\x{}|\z{})$, and the variational approximate posterior $q(\z{}|\x{})$.

In order to extend such models to sequential decision making settings, we must incorporate actions and impose temporal structure on the latent state.
Consider a partially observable MDP (POMDP), with latent states $\z{t} \in \Z$ and its corresponding observations $\x{t} \in \X$.
We make an explicit distinction between an observation $\x{t}$ and the underlying latent state $\z{t}$, to emphasize that the latter is unobserved and its distribution is unknown. Analogous to the MDP, the initial and transition distributions are $p(\z{1})$ and $p(\z{t+1} | \z{t}, \a{t})$, and the reward is $\r{t}$. In addition, the observation model is given by $p(\x{t} | \z{t})$.

As in the case for VAEs, a generative model of these observations $\x{t}$ can be learned by maximizing the log-likelihood. In the POMDP setting, however, we note that $\x{t}$ alone does not provide all necessary information to infer $\z{t}$, and prior observations must be taken into account during inference. This brings us to the discussion of sequential latent variable models.
The distributions of interest are $p(\z{1})$ and $p(\z{t+1} | \z{t}, \a{t})$, the observation model $p(\x{t} | \z{t})$, and the approximate variational posteriors $q(\z{1} | \x{1})$ and $q(\z{t+1} | \x{t+1}, \z{t}, \a{t})$.
The log-likelihood of the observations can then be bounded,
\vspace{-1.5mm}
\begin{equation}
  \log p(\x{1:\tau+1} | \a{1:\tau}) \geq \! \E_{\z{1:\tau+1} \sim q} \!\left[ \sum_{t=0}^{\tau} \log p(\x{t+1} | \z{t+1}) - \kl{q(\z{t+1} | \x{t+1}, \z{t}, \a{t})}{p(\z{t+1} | \z{t}, \a{t})} \!\right]\!\!.
  \label{eq:seq_vae}
\end{equation}
For notational convenience, we define $q(\z{1} | \x{1}, \z{0}, \a{0}) \coloneqq q(\z{1} | \x{1})$ and $p(\z{1} | \z{0}, \a{0}) \coloneqq p(\z{1})$.
Prior work~\citep{buesing2018learning,igl2018deep,gregor2019shaping,hafner2019learning,gregor2019tdvae,kim2019vta,doerr2018prssm,zhang2019solar} has explored modeling such non-Markovian observation sequences, using methods such as recurrent neural networks with deterministic hidden state, as well as probabilistic state-space models. In this work, we enable the effective training of a fully stochastic sequential latent variable model, and bring it together with a maximum entropy actor-critic RL algorithm to create SLAC: a sample-efficient and high-performing RL algorithm for learning policies for complex continuous control tasks directly from high-dimensional image inputs.

%% file: 03_approach.tex
\vspace{-2mm}
\section{Joint Modeling and Control as Inference}
\vspace{-2mm}
\label{sec:derivation}

For a fully observable MDP, the control problem can be embedded into a graphical model by introducing a binary random variable $\optim{t}$, which indicates if time step $t$ is optimal.
When its distribution is chosen to be ${p(\optim{t} = 1 | \s{t}, \a{t}) = \exp(r(\s{t}, \a{t}))}$, then maximization of $p(\optim{1:T})$ via approximate inference in that model yields the optimal policy for the maximum entropy objective~\citep{levine2018reinforcement}.

\begin{wrapfigure}{r}{0.35\columnwidth}
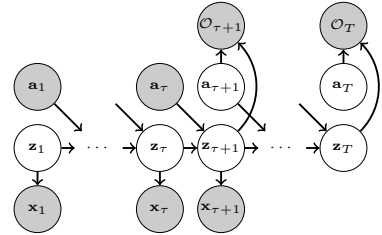

  \centering
  \vspace{-4mm}
  \includestandalone[scale=0.89]{figures/pgm}
  \vspace{-5mm}
  \caption{Graphical model of POMDP with optimality variables for $t \geq \tau+1$.}
  \label{fig:pgm}
  \vspace{-4mm}
\end{wrapfigure}
In this paper, we extend this idea to the POMDP setting, where the probabilistic graphical model includes latent variables, as shown in \autoref{fig:pgm}, and the distribution can analogously be given by $p(\optim{t} = 1 | \z{t}, \a{t}) = \exp(r(\z{t}, \a{t}))$. Instead of maximizing the likelihood of the optimality variables alone, we jointly model the observations (including the observed rewards of the past time steps) and learn maximum entropy policies by maximizing the marginal likelihood $p(\x{1:\tau+1}, \optim{\tau+1:T} | \a{1:\tau})$. This objective represents both the likelihood of the observed data from the past $\tau+1$ steps, as well as the optimality of the agent's actions for future steps, effectively combining both representation learning and control into a single graphical model.
We factorize our variational distribution into a product of \emph{recognition} terms $q(\z{t+1} | \x{t+1}, \z{t}, \a{t})$, \emph{dynamics} terms $p(\z{t+1} | \z{t}, \a{t})$, and \emph{policy} terms $\policy(\a{t} | \x{1:t}, \a{1:t-1})$:
\vspace{-1mm}
\begin{equation}
  q(\z{1:T}, \a{\tau+1:T} | \x{1:\tau+1}, \a{1:\tau}) \!=\! \prod_{t=0}^{\tau} q(\z{t+1} | \x{t+1}, \z{t}, \a{t}) \!\!\!\! \prod_{t=\tau+1}^{T-1} \!\!\!\! p(\z{t+1} | \z{t}, \a{t}) \!\!\!\! \prod_{t=\tau+1}^{T} \!\!\!\! \policy(\a{t} | \x{1:t}, \a{1:t-1}).\!\!
  \label{eq:our_posterior}
\end{equation}\\[-2mm]
The variational distribution uses the dynamics for future time steps to prevent the agent from controlling the transitions and from choosing optimistic actions, analogously to the fully observed MDP setting described by \citet{levine2018reinforcement}.
The posterior over the actions represents the policy \policy{}.
We use the posterior from \autoref{eq:our_posterior} to obtain the evidence lower bound (ELBO) of the likelihood,
\vspace{0.5mm}
\begin{align}
  \MoveEqLeft \log p(\x{1:\tau+1}, \optim{\tau+1:T} | \a{1:\tau}) \nonumber\\
  &\!\!\!\!\!\geq \! \E_{(\z{1:T}, \a{\tau+1:T}) \sim q} \! \left[ \vphantom{\sum} \log p(\x{1:\tau+1}, \optim{\tau+1:T}, \z{1:T}, \a{\tau+1:T} | \a{1:\tau}) - \log q(\z{1:T}, \a{\tau+1:T} | \x{1:\tau+1}, \a{1:\tau}) \right] \nonumber\\[-1.5mm]
  &\!\!\!\!\!= \begin{multlined}[t]
    \E_{(\z{1:T}, \a{\tau+1:T}) \sim q} \left[ \vphantom{\sum_t^\tau}
      \smash{\underbrace{ 
        \sum_{t=0}^{\tau} \left( \vphantom{\sum} \log p(\x{t+1} | \z{t+1}) 
        - \kl{q(\z{t+1} | \x{t+1}, \z{t}, \a{t})}{p(\z{t+1} | \z{t}, \a{t})} \right)
      }_{\text{model objective terms}}}
    \right. \\[3.5mm]
    \left. \vphantom{\sum_t^\tau} {}+ 
      \smash{\underbrace{ 
        \sum_{t=\tau+1}^T \left( \vphantom{\sum} r(\z{t}, \a{t}) + \log p(\a{t}) - \log \policy(\a{t} | \x{1:t}, \a{1:t-1}) \right)
      }_{\text{policy objective terms}}}
    \right] \!,
  \end{multlined}
  \label{eq:elbo}
\end{align} \\[1mm]
where $r(\z{t}, \a{t}) = \log p(\optim{t} = 1 | \z{t}, \a{t})$ by construction and $p(\a{t})$ is the action prior.
The full derivation of the ELBO is given in \autoref{app:elbo_derivation}.
\section{Stochastic Latent Actor Critic}
\label{sec:slac}
\vspace{-2mm}

We now describe our stochastic latent actor critic (SLAC) algorithm, which maximizes the ELBO using function approximators to model the prior and posterior distributions. The ELBO objective in \autoref{eq:elbo} can be split into a model objective and a maximum entropy RL objective. The model objective can be optimized directly, while the maximum entropy RL objective can be optimized via approximate message passing, with messages corresponding to the Q-function. We can rewrite the RL objective to express it in terms of these messages, yielding an actor-critic algorithm analogous to SAC. Additional details of the derivation of the SLAC objectives are given in \autoref{app:elbo_derivation}.

\vspace{-0.5mm}
\textbf{Latent variable model.} The first part of the ELBO corresponds to training the latent variable model to maximize the likelihood of the observations, analogous to the ELBO in \autoref{eq:seq_vae} for the sequential latent variable model.
The generative model is given by $p_\mparams(\z{1})$, $p_\mparams(\z{t+1} | \z{t}, \a{t})$, and $p_\mparams(\x{t} | \z{t})$, and the inference model is given by $q_\mparams(\z{1} | \x{1})$ and $q_\mparams(\z{t+1} | \x{t+1}, \z{t}, \a{t})$.
These distributions are diagonal Gaussian, where the means and variances are given by outputs of neural networks. Further details of our specific model architecture are given in \autoref{app:network_architectures}.
The distribution parameters \mparams{} are optimized with respect to the ELBO in \autoref{eq:elbo}, where the only terms that depend on \mparams{}, and therefore constitute the model objective, are given by
\vspace{-2mm}
\begin{equation}
  \hspace{-.2mm} J_M(\mparams) = \! \E_{{\z{1:\tau+1} \sim q_\mparams}} \! \left[ \sum_{t=0}^{\tau} -\log p_\mparams(\x{t+1} | \z{t+1}) + \klcompact{q_\mparams(\z{t+1} | \x{t+1}, \z{t}, \a{t})}{p_\mparams(\z{t+1} | \z{t}, \a{t})} \right]\!\!,\hspace{-2mm}
  \label{eq:our_model_loss}
\end{equation} \\[-2mm]
where we define $q_{\mparams}(\z{1} | \x{1}, \z{0}, \a{0}) \coloneqq q_{\mparams}(\z{1} | \x{1})$ and $p_{\mparams}(\z{1} | \z{0}, \a{0}) \coloneqq p_{\mparams}(\z{1})$. We use the reparameterization trick to sample from the filtering distribution $q_\mparams(\z{1:\tau+1} | \x{1:\tau+1}, \a{1:\tau})$.

\vspace{-0.5mm}
\textbf{Actor and critic.} The second part of the ELBO corresponds to the maximum entropy RL objective. As in the fully observable case from \autoref{sec:maxent} and as described by \citet{levine2018reinforcement}, this optimization can be solved via message passing of soft Q-values. However, in our method, we must use the latent states $\z{}$, since the true state is unknown. The messages are approximated by minimizing the soft Bellman residual, which we use to train our soft Q-function parameters $\Qparams$,  %
\vspace{-2mm}
\begin{align}
  J_Q(\Qparams) &= \E_{\z{1:\tau+1} \sim q_\mparams}  \left[ \tfrac{1}{2}  \left( Q_\Qparams(\z{\tau}, \a{\tau}) - \left( \r{\tau} + \gamma V_{\bar{\Qparams}}(\z{\tau+1}) \right) \right)^{2} \right], \\
  V_{\Qparams}(\z{\tau+1}) &= \E_{\a{\tau+1} \sim \policy_\pparams} \left[ Q_{\Qparams}(\z{\tau+1}, \a{\tau+1}) - \alpha \log \policy_\pparams(\a{\tau+1} | \x{1:\tau+1}, \a{1:\tau}) \right],
  \label{eq:our_critic_loss}
\end{align}\\[-3mm]
where $V_{\Qparams}$ is the soft state value function and $\bar{\Qparams}$ are delayed target network parameters, obtained as exponential moving averages of $\Qparams$.
Notice that the latents $\z{\tau}$ and $\z{\tau+1}$, which are used in the Bellman backup, are sampled from the same filtering distribution, i.e. $\z{\tau+1} \sim q_\mparams(\z{\tau+1} | \x{\tau+1}, \z{\tau}, \a{\tau})$.
The RL objective, which corresponds to the second part of the ELBO, can then be rewritten in terms of the soft Q-function. The policy parameters $\pparams$ are optimized to maximize this objective, resulting in a policy loss analogous to soft actor-critic~\citep{haarnoja2018soft}:
\vspace{-2mm}
\begin{equation}
  J_\policy(\pparams) = \E_{\z{1:\tau+1} \sim q_\mparams} \left[ \E_{\a{\tau+1} \sim \policy_\pparams} \left[ \vphantom{\sum} \alpha \log \policy_\pparams(\a{\tau+1} | \x{1:\tau+1}, \a{1:\tau}) - Q_\Qparams(\z{\tau+1}, \a{\tau+1}) \right] \vphantom{\E_{\a{\tau+1}}} \right].\hspace{-2mm}
  \label{eq:our_actor_loss}
\end{equation} \\[-6mm]
\algrenewcommand\algorithmicindent{4mm}%
\begin{wrapfigure}{r}{0.46\columnwidth}
  \vspace{-7.5mm}
  \begin{minipage}{0.46\columnwidth}
  \captionsetup{type=algorithm}
\begin{algorithm}[H]
  \footnotesize
  \caption{Stochastic Latent Actor-Critic (SLAC)}
  \label{alg:SLAC}
  \begin{algorithmic}
    \Require Environment $E$ and initial parameters ${\mparams, \pparams, \Qparams_1, \Qparams_2}$ for the model, actor, and critics.
    \State $\x{1} \sim E_{\text{reset}}()$  %
    \State $\mathcal{D} \leftarrow (\x{1})$  %
    \For{each iteration}
      \For{each environment step}
          \State $\a{t} \sim \policy_\pparams(\a{t} | \x{1:t}, \a{1:t-1})$  %
          \State $\r{t}, \x{t+1} \sim E_{\text{step}}(\a{t})$  %
          \State $\mathcal{D} \leftarrow \mathcal{D} \cup (\a{t}, \r{t}, \x{t+1})$  %
      \EndFor
      \For{each gradient step}
          \State $\x{1:\tau+1}, \a{1:\tau}, \r{\tau} \sim \mathcal{D}$  %
          \State $\z{1:\tau+1} \sim q_\mparams(\z{1:\tau+1} | \x{1:\tau+1}, \a{1:\tau})$  %
          \State $\mparams \leftarrow \mparams - \lambda_M \nabla_\mparams J_M(\mparams)$  %
          \State $\Qparams_i \leftarrow \Qparams_i - \lambda_Q \nabla_{\Qparams_i} J_Q(\Qparams_i)$ for $i\in\{1, 2\}$  %
          \State $\pparams \leftarrow \pparams - \lambda_\policy \nabla_\pparams J_\policy(\pparams)$  %
          \State $\bar{\Qparams}_i \leftarrow \nu \Qparams_i + (1-\nu)\bar{\Qparams}_i$ for $i\in\{1,2\}$  %
      \EndFor
    \EndFor
  \end{algorithmic}
\end{algorithm}
  \end{minipage}
  \vspace{-2mm}
\end{wrapfigure}
We assume a uniform action prior, so $\log p(\a{t})$ is a constant term that we omit from the policy loss. This loss only uses the last sample $\z{\tau+1}$ of the sequence for the critic, and we use the reparameterization trick to sample from the policy. 
Note that the policy is not conditioned on the latent state, as this can lead to over-optimistic behavior since the algorithm would learn Q-values for policies that have perfect access to the latent state. Instead, the learned policy in our algorithm is conditioned directly on the past observations and actions. This has the additional benefit that the learned policy can be executed at run time without requiring inference of the latent state.
Finally, we note that for the expectation over latent states in the Bellman residual in \autoref{eq:our_critic_loss}, rather than sampling latent states for all $\z{} \sim \Z$, we sample latent states from the filtering distribution $q_\mparams(\z{1:\tau+1} | \x{1:\tau+1}, \a{1:\tau})$. This design choice allows us to minimize the critic loss for samples that are most relevant for $Q_\Qparams$, while also allowing the critic loss to use the Q-function in the same way as implied by the policy loss in \autoref{eq:our_actor_loss}.

\begin{figure*}
  \centering
  \includegraphics[width=0.119\textwidth]{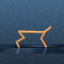}
  \includegraphics[width=0.119\textwidth]{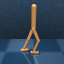}
  \includegraphics[width=0.119\textwidth]{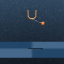}
  \includegraphics[width=0.119\textwidth]{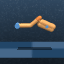}
  \includegraphics[width=0.119\textwidth]{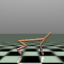}
  \includegraphics[width=0.119\textwidth]{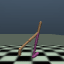}
  \includegraphics[width=0.119\textwidth]{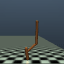}
  \includegraphics[width=0.119\textwidth]{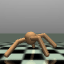}
  \caption{
  Example image observations for our continuous control benchmark tasks: DeepMind Control's cheetah run, walker walk, ball-in-cup catch, and finger spin, and OpenAI Gym's half cheetah, walker, hopper, and ant (left to right).
  These images, which are rendered at a resolution of $64 \times 64$ pixels, are the observation inputs to our algorithm, i.e. to the latent variable model and to the policy.}
  \label{fig:tasks}
\end{figure*}

SLAC is outlined in \autoref{alg:SLAC}. The actor-critic component follows prior work, with automatic tuning of the temperature $\alpha$ and two Q-functions to mitigate overestimation~\citep{fujimoto2018addressing,haarnoja2018soft,haarnoja2018applications}.
SLAC can be viewed as a variant of SAC~\citep{haarnoja2018soft} where the critic is trained on the stochastic latent state of our sequential latent variable model. The backup for the critic is performed on a tuple $(\z{\tau}, \a{\tau}, \r{\tau}, \z{\tau+1})$, sampled from the filtering distribution $q_\mparams(\z{\tau+1}, \z{\tau} | \x{1:\tau+1}, \a{1:\tau})$. The critic can, in principle, take advantage of the perfect knowledge of the state $\z{t}$, which makes learning easier. However, the policy does not have access to $\z{t}$, and must make decisions based on a history of observations and actions.
SLAC is not a model-based algorithm, in that in does not use the model for prediction, but we see in our experiments that SLAC can achieve similar sample efficiency as a model-based algorithm.

%% file: figures/pgm.tex
\begin{tikzpicture}[
  node distance=2mm, auto,
  empty/.style={shape=circle, minimum size=7mm, inner sep=0mm},
  obs/.style={shape=circle, draw=black, fill=black!20, minimum size=7mm, inner sep=0mm},
  unobs/.style={shape=circle, draw=black, minimum size=7mm, inner sep=0mm},
  gen/.style={->, draw=black, thick},
  bendgen/.style={->, bend right=45, draw=black, thick},
]
{\tiny
  \node[unobs] (z1) {$\z{1}$};
  \node[empty, right=of z1] (z2) {$\cdots$};
  \node[unobs, right=of z2] (zt) {$\z{\tau}$};
  \node[unobs, right=of zt] (ztp1) {$\z{\tau+1}$};
  \node[empty, right=of ztp1] (ztp2) {$\cdots$};
  \node[unobs, right=of ztp2] (zT) {$\z{T}$};

  \node[obs, above=of z1] (a1) {$\a{1}$};
  \node[empty, above=of z2] (a2) {};
  \node[obs, above=of zt] (at) {$\a{\tau}$};
  \node[unobs, above=of ztp1] (atp1) {$\a{\tau+1}$};
  \node[empty, above=of ztp2] (atp2) {};
  \node[unobs, above=of zT] (aT) {$\a{T}$};

  \node[obs, below=of z1] (x1) {$\x{1}$};
  \node[obs, below=of zt] (xt) {$\x{\tau}$};
  \node[obs, below=of ztp1] (xtp1) {$\x{\tau+1}$};

  \node[obs, above=of atp1] (otp1) {$\optim{\tau+1}$};
  \node[obs, above=of aT] (oT) {$\optim{T}$};
}
  \path (z1) edge[gen] node {} (z2);
  \path (z2) edge[gen] node {} (zt);
  \path (zt) edge[gen] node {} (ztp1);
  \path (ztp1) edge[gen] node {} (ztp2);
  \path (ztp2) edge[gen] node {} (zT);

  \path (a1) edge[gen] node {} (z2);
  \path (a2) edge[gen] node {} (zt);
  \path (at) edge[gen] node {} (ztp1);
  \path (atp1) edge[gen] node {} (ztp2);
  \path (atp2) edge[gen] node {} (zT);

  \path (z1) edge[gen] node {} (x1);
  \path (zt) edge[gen] node {} (xt);
  \path (ztp1) edge[gen] node {} (xtp1);

  \path (atp1) edge[gen] node {} (otp1);
  \path (ztp1) edge[bendgen] node {} (otp1);
  \path (aT) edge[gen] node {} (oT);
  \path (zT) edge[bendgen] node {} (oT);
\end{tikzpicture}

%% file: 04_experiments.tex
\section{Experimental Evaluation}

We evaluate SLAC on multiple image-based continuous control tasks from both the DeepMind Control Suite~\citep{tassa2018deepmind} and OpenAI Gym~\citep{brockman2016openai}, as illustrated in~\autoref{fig:tasks}.
Full details of SLAC's network architecture are described in \autoref{app:network_architectures}.
Training and evaluation details are given in \autoref{app:train_eval}, and image samples from our model for all tasks are shown in \autoref{app:samples}. Additionally, visualizations of our results and code are available on the project website.\footnote{\url{https://alexlee-gk.github.io/slac/}}
\subsection{Comparative Evaluation on Continuous Control Benchmark Tasks}

To provide a comparative evaluation against prior methods, we evaluate SLAC on four tasks (cheetah run, walker walk, ball-in-cup catch, finger spin) from the DeepMind Control Suite~\citep{tassa2018deepmind}, and four tasks (cheetah, walker, ant, hopper) from OpenAI Gym~\citep{brockman2016openai}. Note that the Gym tasks are typically used with low-dimensional state observations, while we evaluate on them with raw image observations. We compare our method to the following state-of-the-art model-based and model-free algorithms:

\textbf{SAC}~\citep{haarnoja2018soft}: This is an off-policy actor-critic algorithm, which represents a comparison to state-of-the-art model-free learning. We include experiments showing the performance of SAC based on true state (as an upper bound on performance) as well as directly from raw images.

\textbf{D4PG}~\citep{barth2018distributed}: This is also an off-policy actor-critic algorithm, learning directly from raw images. The results reported in the plots below are the performance after $10^8$ training steps, as stated in the benchmarks from~\citet{tassa2018deepmind}.

\textbf{MPO}~\citep{abdolmaleki2018mpo,abdolmaleki2018mpo_b}: This is an off-policy actor-critic algorithm that performs an expectation maximization form of policy iteration, learning directly from raw images.

\textbf{DVRL}~\citep{igl2018deep}: This is an on-policy model-free RL algorithm that trains a partially stochastic latent-variable POMDP model. DVRL uses the \emph{full belief} over the latent state as input into both the actor and critic, as opposed to our method, which trains the critic with the latent state and the actor with a history of actions and observations.

\textbf{PlaNet}~\citep{hafner2019learning}: This is a model-based RL method for learning from images, which uses a partially stochastic sequential latent variable model, but without explicit policy learning. Instead, the model is used for planning with model predictive control (MPC), where each plan is optimized with the cross entropy method (CEM).

\textbf{DrQ}~\citep{kostrikov2020drq}: This is the same as the SAC algorithm, but combined with data augmentation on the image inputs.

\begin{figure}
  \centering
  \includegraphics[scale=0.37, trim={2mm 2.5mm 2.5mm 2.5mm}, clip]{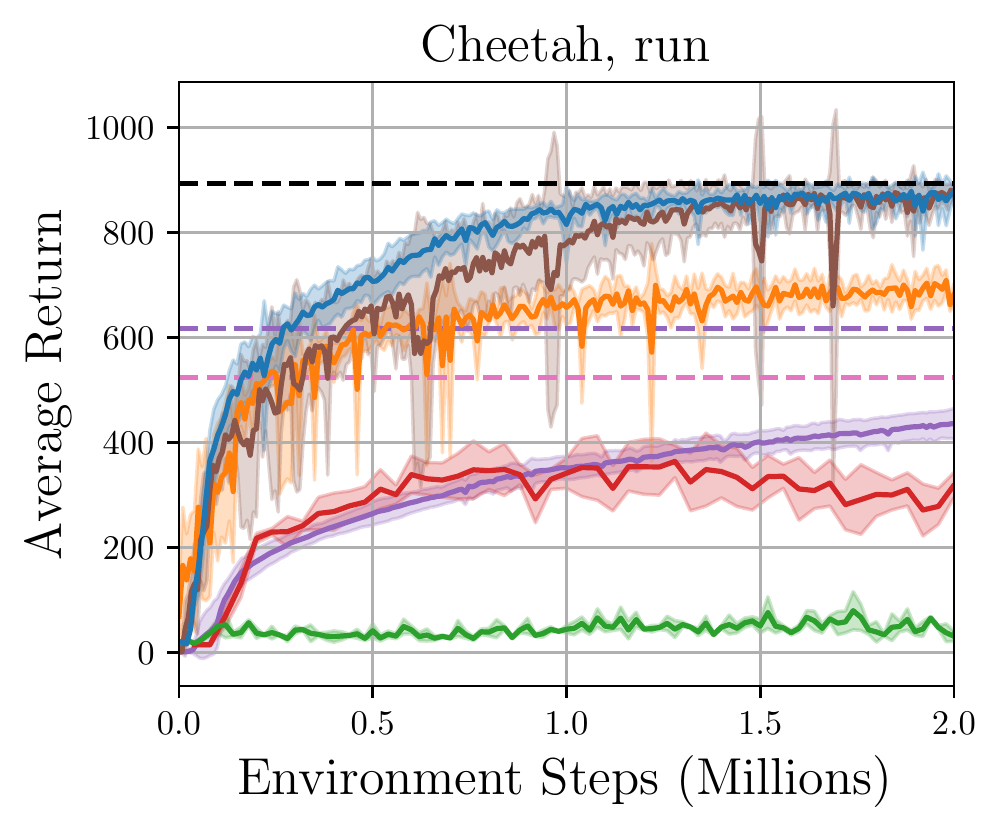} \hfill
  \includegraphics[scale=0.37, trim={8.5mm 2.5mm 2.5mm 2.5mm}, clip]{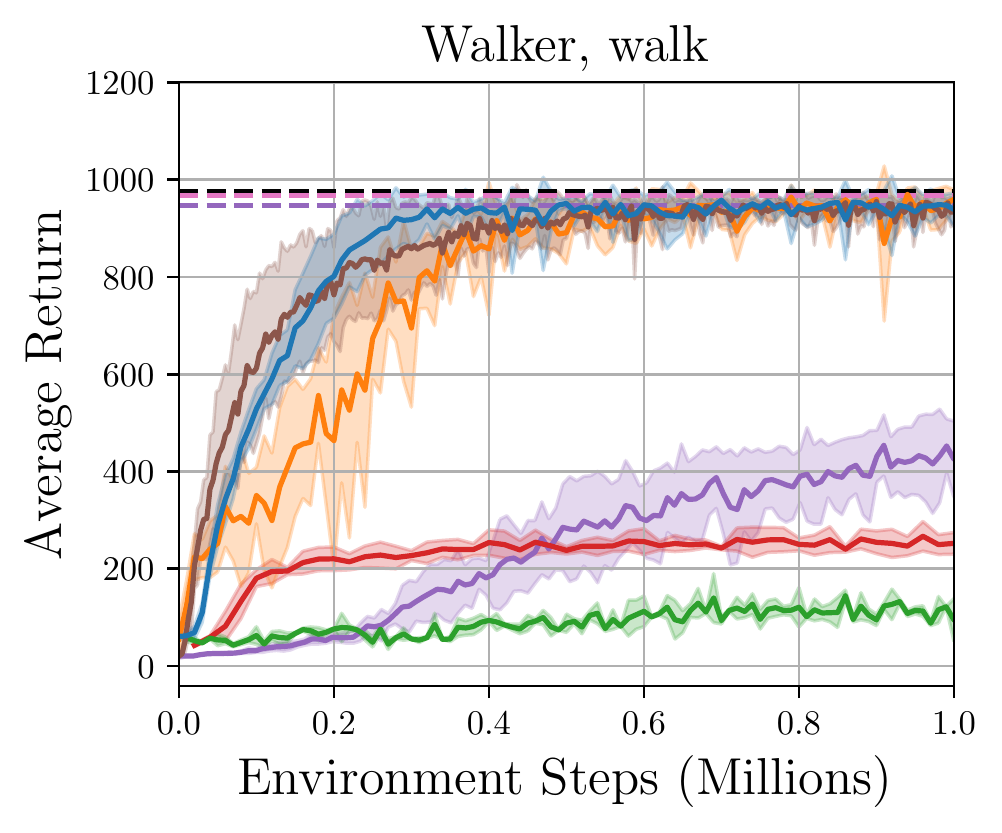} \hfill
  \includegraphics[scale=0.37, trim={8.5mm 2.5mm 2.5mm 2.5mm}, clip]{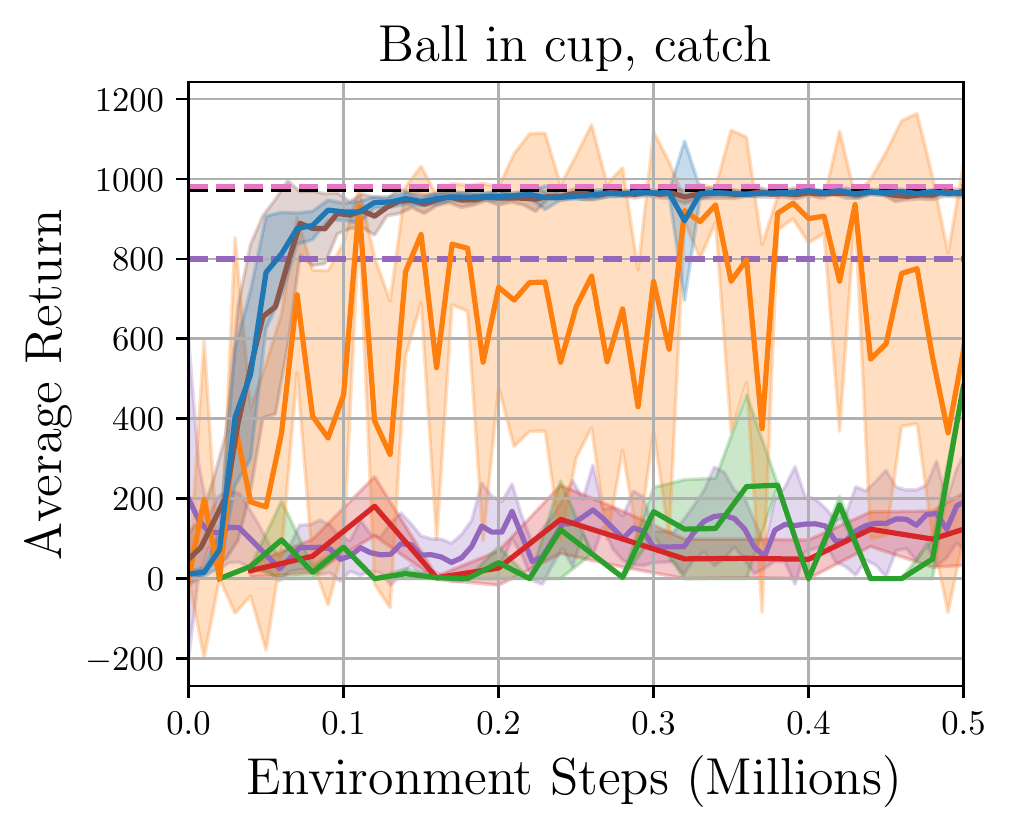} \hfill
  \includegraphics[scale=0.37, trim={8.5mm 2.5mm 2.5mm 2.5mm}, clip]{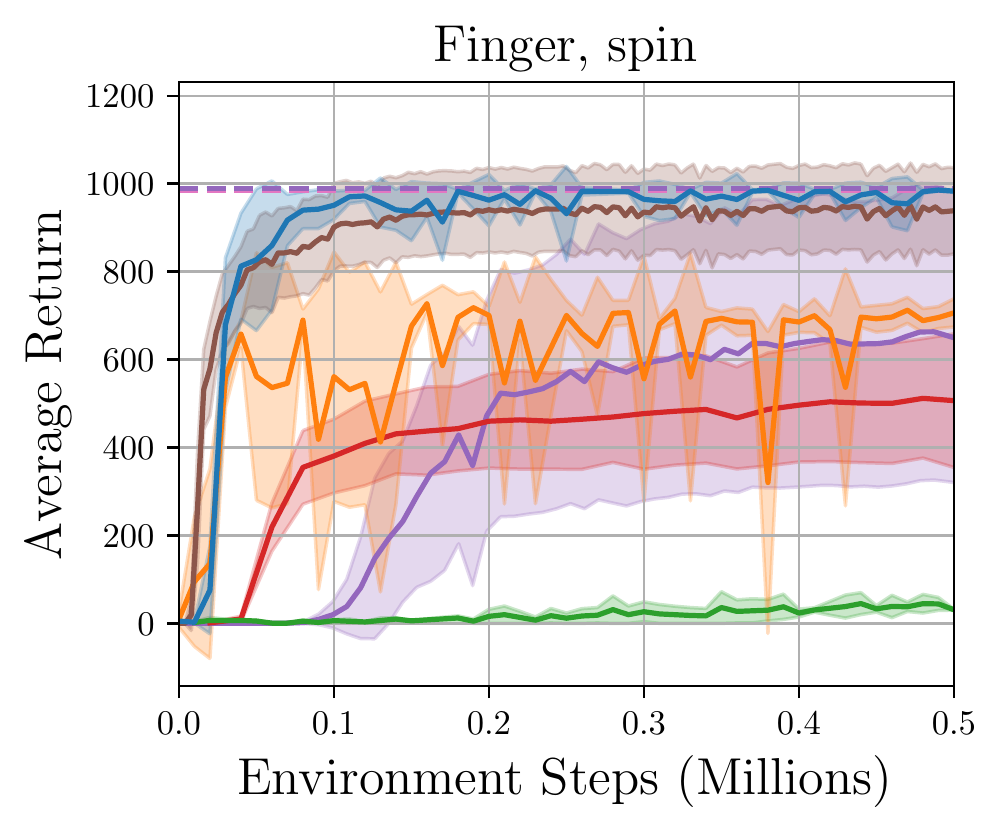} \\
  \includegraphics[scale=0.35]{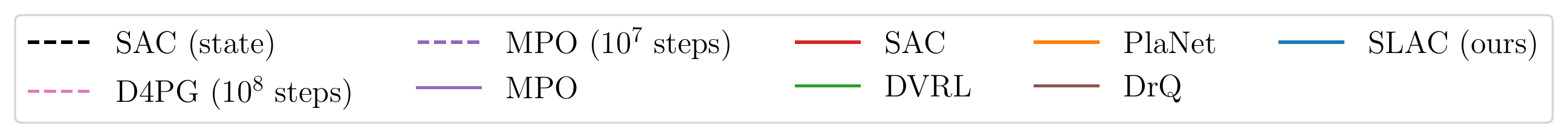}
  \vspace{-2mm}
  \caption{Experiments on the DeepMind Control Suite from images (unless otherwise labeled as ``state").
  SLAC (ours) converges to similar or better final performance than the other methods, while almost always achieving reward as high as the upper bound SAC baseline that learns from true state. Note that for these experiments, 1000 environments steps corresponds to 1 episode.
  }
  \label{fig:dm_control_plots}
\end{figure}
\begin{figure}
  \centering
  \includegraphics[scale=0.37, trim={2mm 2.5mm 2.5mm 2.5mm}, clip]{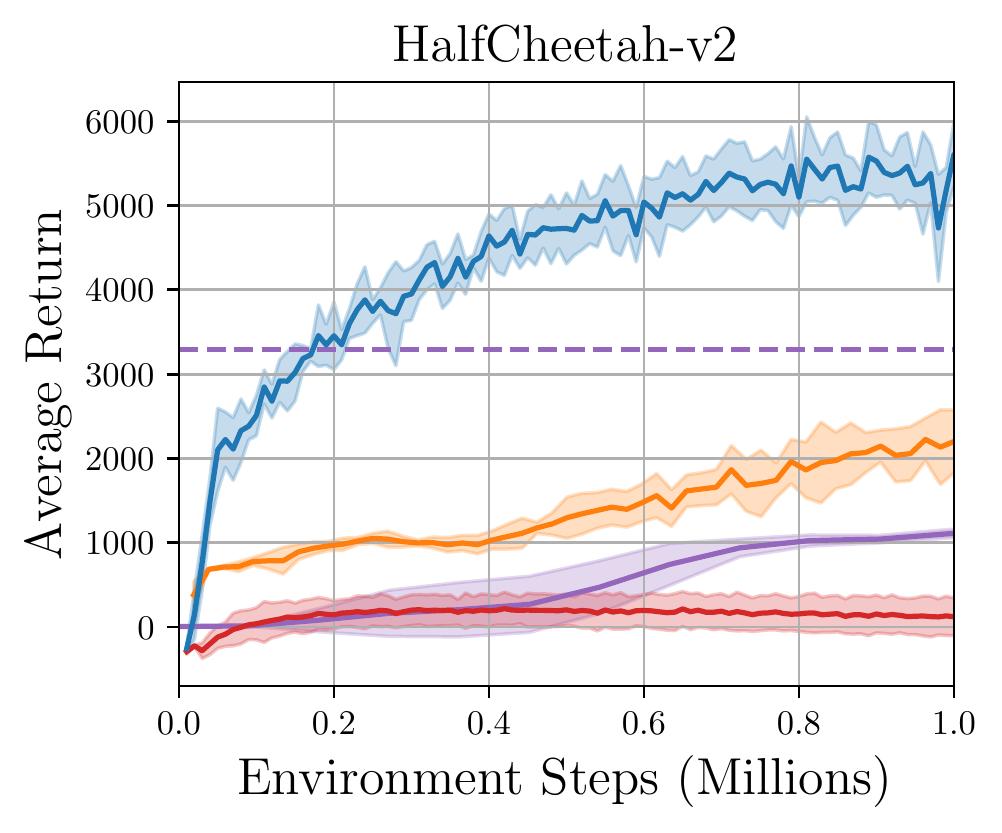} \hfill
  \includegraphics[scale=0.37, trim={8.5mm 2.5mm 2.5mm 2.5mm}, clip]{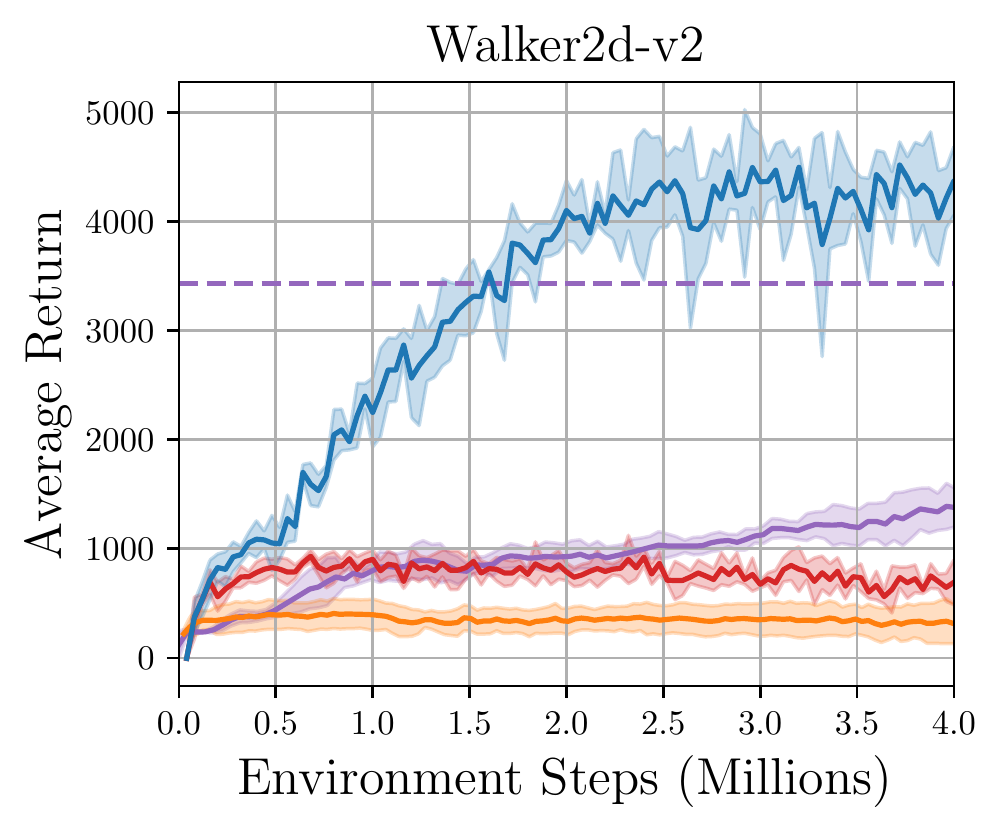} \hfill
  \includegraphics[scale=0.37, trim={8.5mm 2.5mm 2.5mm 2.5mm}, clip]{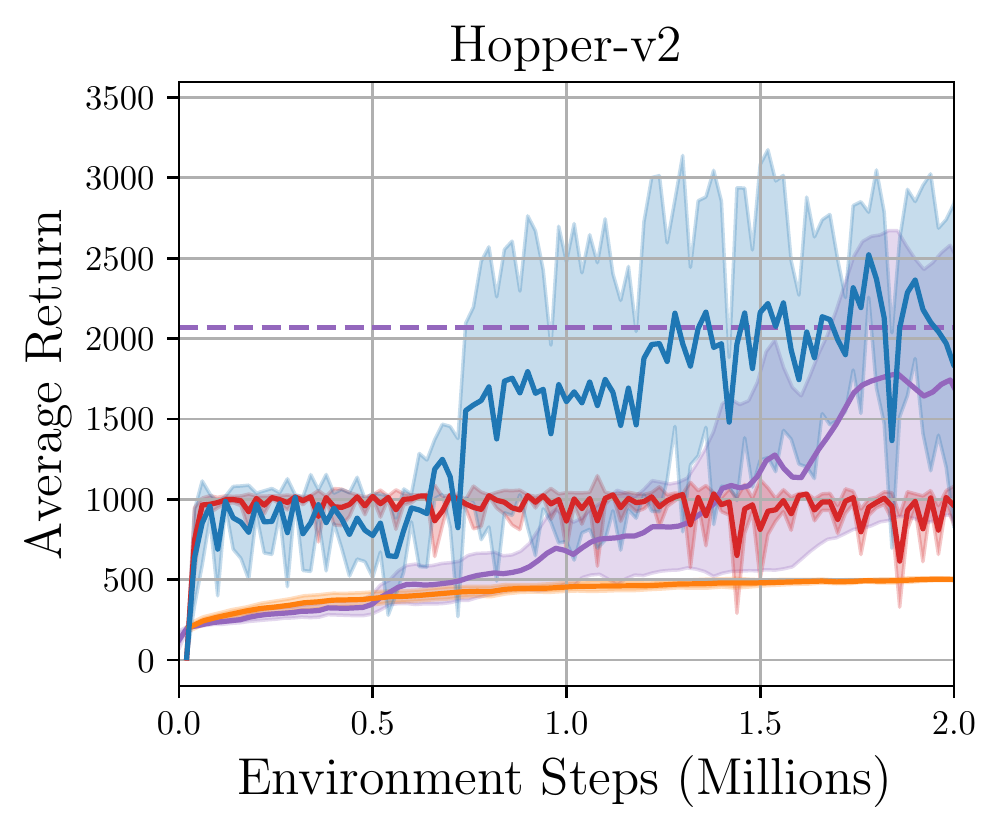} \hfill
  \includegraphics[scale=0.37, trim={8.5mm 2.5mm 2.5mm 2.5mm}, clip]{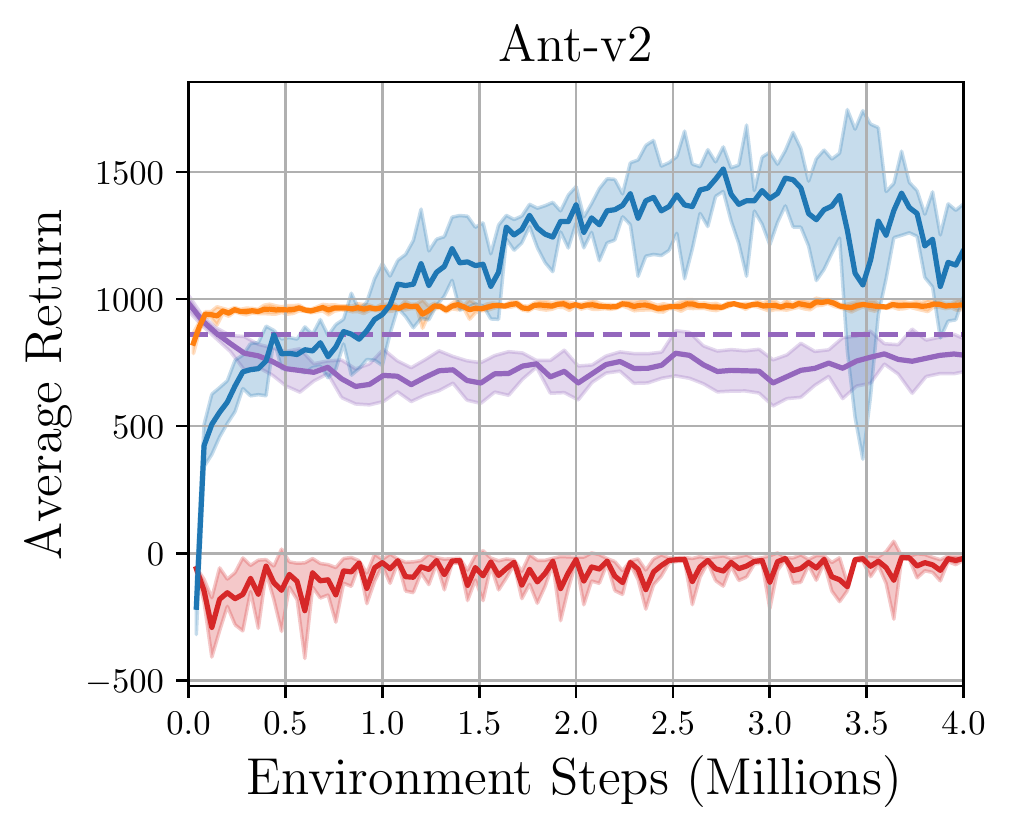} \\
  \includegraphics[scale=0.35]{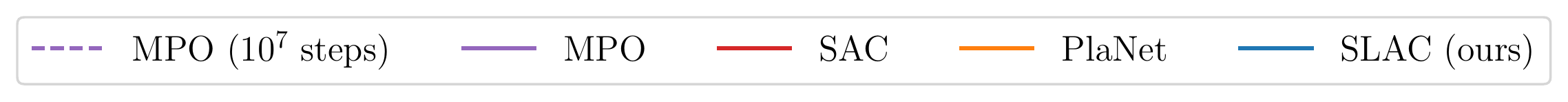}
  \vspace{-2mm}
  \caption{Experiments on the OpenAI Gym benchmark tasks from images. SLAC (ours) converges to higher performance than both PlaNet and SAC on all four of these tasks. The number of environments steps in each episode is variable, depending on the termination.
  }
  \label{fig:gym_plots}
  \vspace{-1mm}
\end{figure}

Our experiments on the DeepMind Control Suite in \autoref{fig:dm_control_plots} show that the sample efficiency of SLAC is comparable or better than \emph{both} model-based and model-free alternatives. This indicates that overcoming the representation learning bottleneck, coupled with efficient off-policy RL, provides for fast learning similar to model-based methods, while attaining final performance comparable to fully model-free techniques that learn from state. SLAC also substantially outperforms DVRL. This difference can be explained in part by the use of an efficient off-policy RL algorithm, which can better take advantage of the learned representation. SLAC achieves comparable or slightly better performance than subsequent work DrQ, which also uses the efficient off-policy SAC algorithm.

We also evaluate SLAC on continuous control benchmark tasks from OpenAI Gym in \autoref{fig:gym_plots}.
We notice that these tasks are more challenging than the DeepMind Control Suite tasks, because the rewards are not as shaped and not bounded between 0 and 1, the dynamics are different, and the episodes terminate on failure (e.g., when the hopper or walker falls over).
PlaNet is unable to solve the last three tasks, while for the cheetah task, it learns a suboptimal policy that involves flipping the cheetah over and pushing forward while on its back. %
To better understand the performance of fixed-horizon MPC on these tasks, we also evaluated with the ground truth dynamics (i.e., the true simulator), and found that even in this case, MPC did not achieve good final performance, suggesting that infinite horizon policy optimization, of the sort performed by SLAC and model-free algorithms, is important to attain good results on these tasks.

Our experiments show that SLAC successfully learns complex continuous control benchmark tasks from raw image inputs. On the DeepMind Control Suite, SLAC exceeds the performance of prior work PlaNet on the four tasks, and SLAC achieves comparable or slightly better performance than subsequence work DrQ. However, on the harder image-based OpenAI Gym tasks, SLAC outperforms PlaNet by a large margin. We note that the prior methods that we tested generally performed poorly on the image-based OpenAI Gym tasks, despite considerable hyperparameter tuning.

\vspace{-1mm}
\subsection{Ablation Experiments}
\label{sec:ablations}
\vspace{-1mm}
We investigate how SLAC is affected by the choice of latent variable model, the inputs given to the actor and critic, the model pretraining, and the number of training updates relative to the number of agent interactions. 
Additional results are given in \autoref{app:ablations}, including experiments that compare the effect of the decoder output variance and using random cropping for data augmentation.

\begin{figure}
  \centering
  \begin{subfigure}[t]{0.26\textwidth}
    \centering
    \includegraphics[scale=0.35, trim={2mm 2.4mm 0mm 2.5mm}, clip]{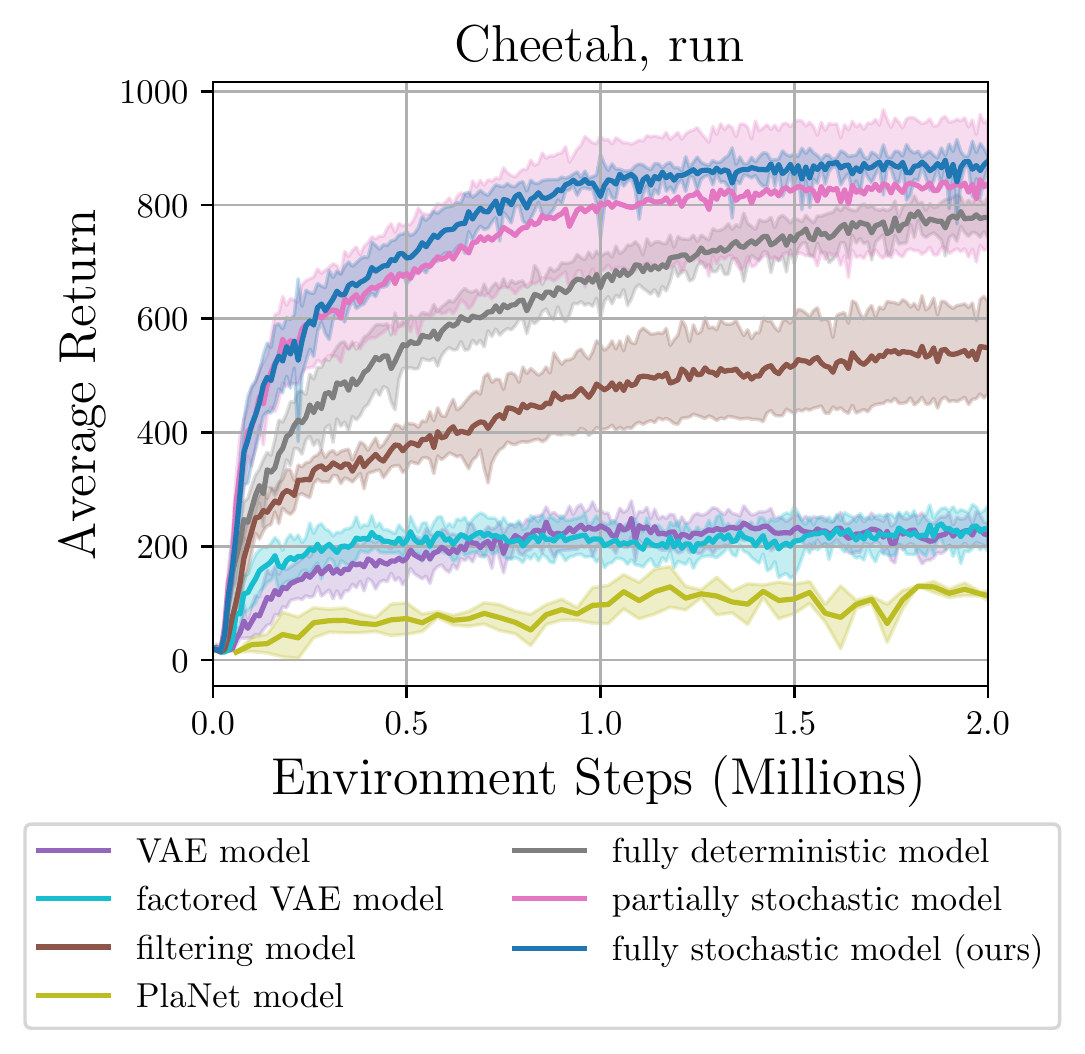}
    \caption{Latent variable model}
    \label{fig:model_ablation}
  \end{subfigure}
  \hfill
  \begin{subfigure}[t]{0.24\textwidth}
    \centering
    \includegraphics[scale=0.35, trim={2mm 2.4mm 2.5mm 2.5mm}, clip]{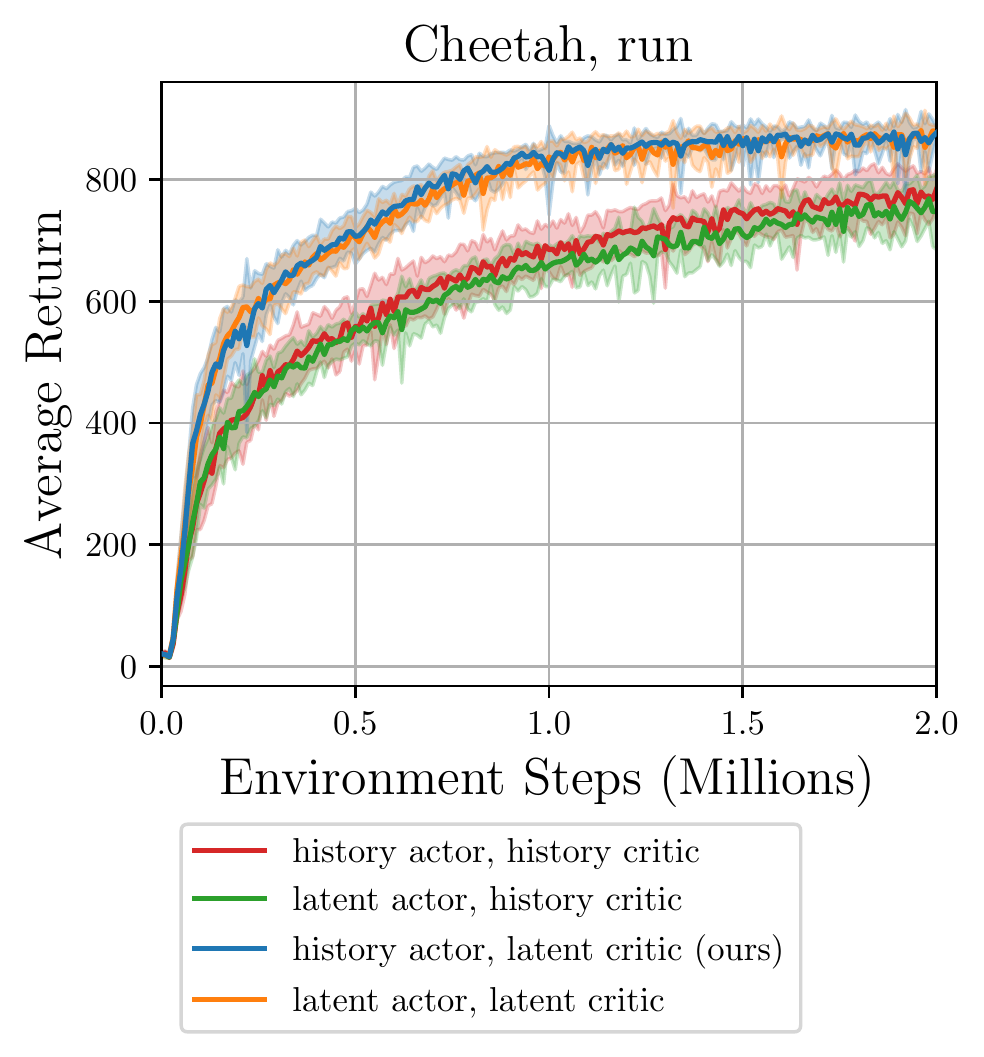}
    \caption{Actor and critic inputs}
    \label{fig:actor_input_critic_input_ablation}
  \end{subfigure}
  \hfill
  \begin{subfigure}[t]{0.235\textwidth}
    \centering
    \includegraphics[scale=0.35, trim={2mm 2.4mm 2.5mm 2.5mm}, clip]{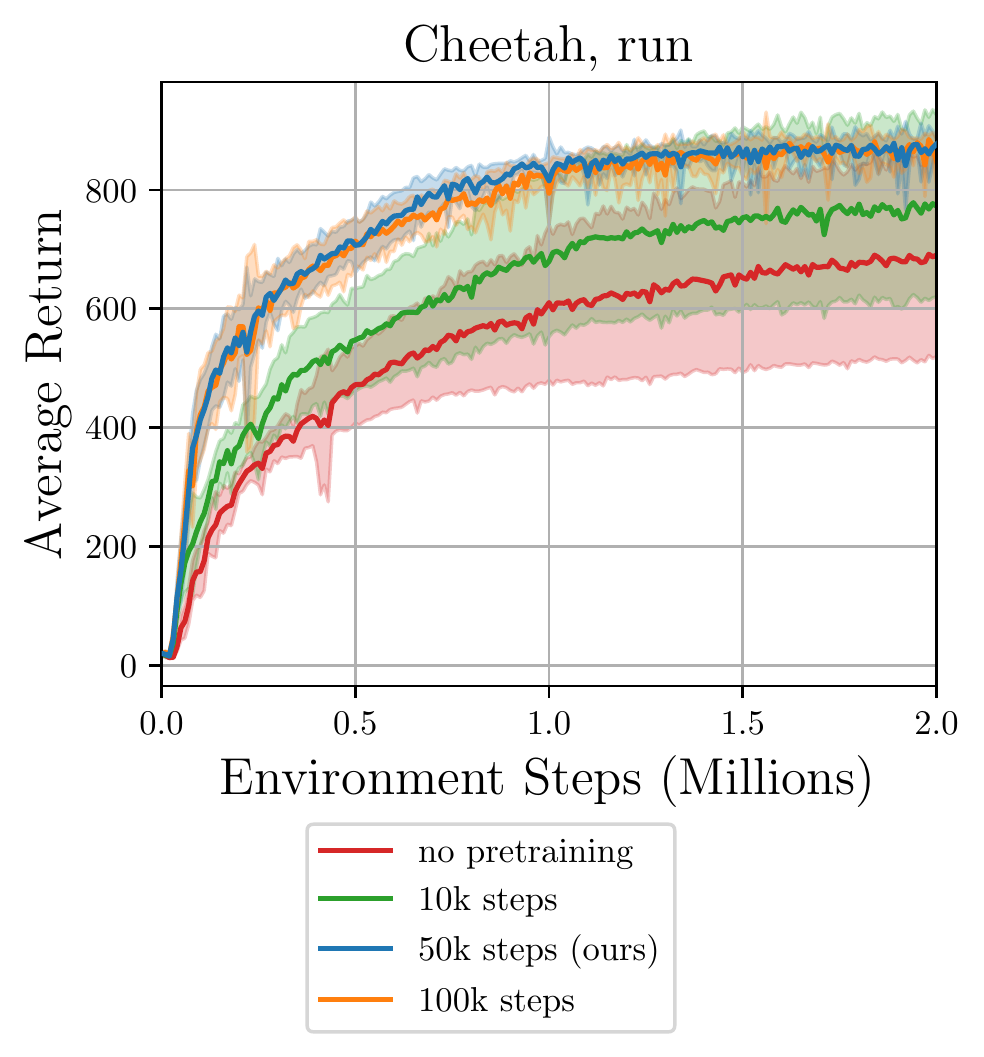}
    \caption{Model pretraining}
    \label{fig:pretraining_steps_ablation}
  \end{subfigure}
  \hfill
  \begin{subfigure}[t]{0.245\textwidth}
    \centering
    \includegraphics[scale=0.35, trim={2mm 2.4mm 2.5mm 2.5mm}, clip]{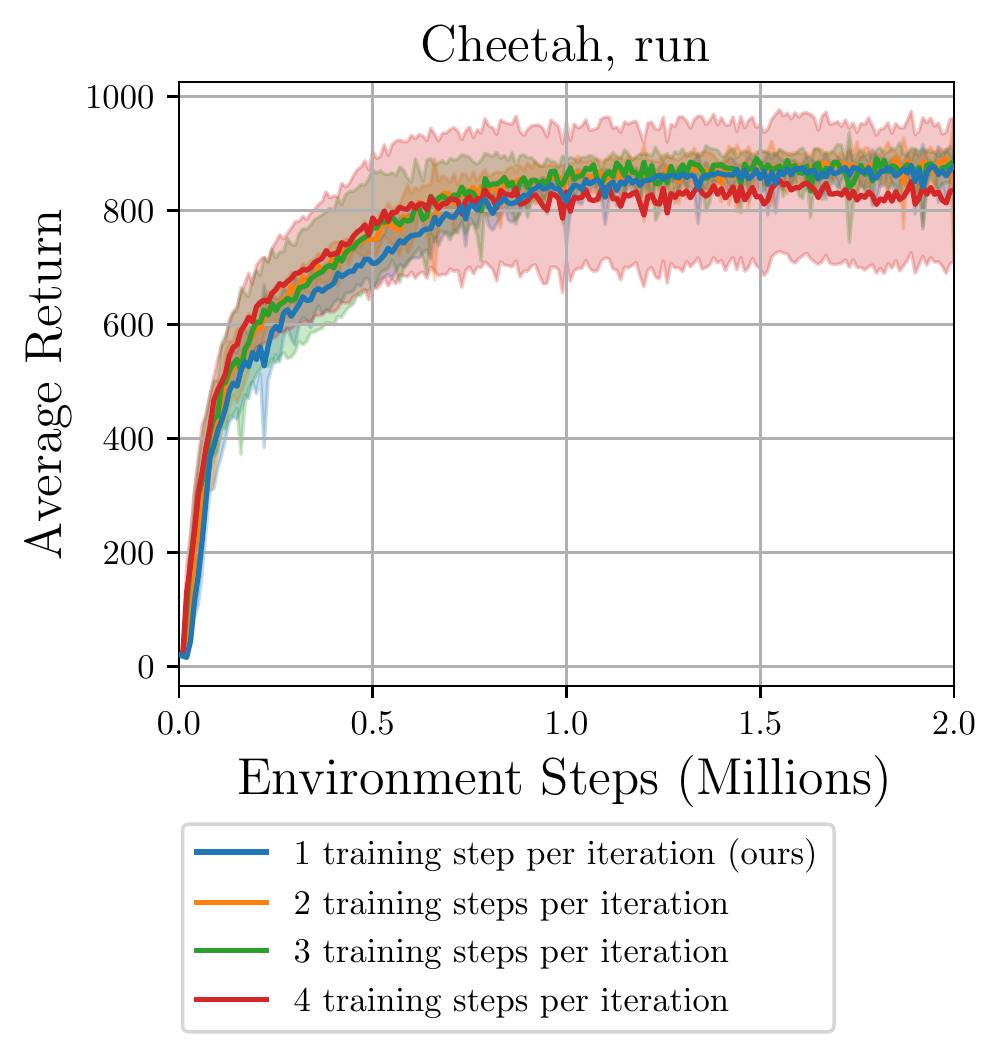}
    \caption{Train steps per iteration}
    \label{fig:train_steps_per_iteration_ablation}
  \end{subfigure}
  \vspace{-1.5mm}
  \caption{Comparison of different design choices for 
  (\subref{fig:model_ablation}) the latent variable model, 
  (\subref{fig:actor_input_critic_input_ablation}) the inputs given to the actor and critic, either the history of past observations and actions, or a latent sample, 
  (\subref{fig:pretraining_steps_ablation}) the number of model pretraining steps, and
  (\subref{fig:train_steps_per_iteration_ablation}) the number of training updates per iteration. 
  In all cases, we use the RL framework of SLAC. %
  See \autoref{fig:model_ablation_all}, \autoref{fig:actor_input_critic_input_ablation_all}, \autoref{fig:pretraining_steps_ablation_all}, and \autoref{fig:train_steps_per_iteration_ablation_all} for results on 5 additional tasks.
  }
  \vspace{-1.5mm}
  \label{fig:ablations}
\end{figure}

\textbf{Latent variable model.}
We study the tradeoffs between different design choices for the latent variable model in \autoref{fig:model_ablation} and \autoref{fig:model_ablation_all}. We compare our \emph{fully stochastic} model to a standard non-sequential \emph{VAE} model~\citep{kingma2013auto}, which has been used in multiple prior works for representation learning in RL~\citep{higgins2017darla, ha2018world, nair2018visual}, and a non-sequential \emph{factored VAE} model, which uses our autoregressive two-variable factorization but without any temporal dependencies.
We also compare to a sequential \emph{filtering} model that uses temporal dependencies but without the two-variable factorization, the partially stochastic model used by \emph{PlaNet}~\citep{hafner2019learning}, as well as two additional variants of our model:
a \emph{fully deterministic} model that removes all stochasticity from the hidden state dynamics, and a \emph{partially stochastic} model that adds deterministic paths in the transitions,
similar to the PlaNet model, but with our latent factorization and architecture.
All the models, except for the PlaNet model, are variants of our model that use the same architecture as our fully stochastic model, with minimal differences in the transitions or the latent variable factorization.
In all cases, we use the RL framework of SLAC and only vary the choice of model for representation learning.

Our fully stochastic model outperforms all the other models.
Contrary to the conclusions in prior work~\citep{hafner2019learning, buesing2018learning}, the fully stochastic model slightly outperforms the partially stochastic model, while retaining the appealing interpretation of a stochastic state space model.
We hypothesize that these prior works benefit from the deterministic paths (realized as an LSTM or GRU) because they use multi-step samples from the prior. In contrast, our method uses samples from the posterior, which are conditioned on same-step observations, and thus it is less sensitive to the propagation of the latent states through time.
The sequential variants of our model (including ours) outperform the non-sequential VAE models.
The models with the two-variable factorization perform similarly or better than their respective equivalents among the non-sequential VAE models and among the sequential stochastic models.
Overall, including temporal dependencies results in the largest improvement in performance, followed by the autoregressive latent variable factorization and using a fully stochastic model.

\textbf{Actor and critic inputs.}
We next investigate alternative choices for the actor and critic inputs as either the observation-action history or the latent sample.
In SLAC, the actor is conditioned on the observation-action history and the critic is conditioned on individual latent samples.
The images in the history are first compressed with the model's convolutional network before they are given to the networks.
However, the actor and critic losses do not propagate any gradient signal into the model nor its convolutional layers, i.e. the convolutional layers used for the observation-action history are only trained by the model loss.

\autoref{fig:actor_input_critic_input_ablation} and \autoref{fig:actor_input_critic_input_ablation_all} show that, in general, the performance is significantly worse when the critic input is the history instead of the latent sample, and indifferent to the choice for the actor input. This is consistent with our derivation---the critic should be given latent samples, but the actor can be conditioned on anything (since the policy is the variational posterior). However, we note that a latent-conditioned actor could lead to overconfident behaviors in uncertain environments. For generality, we choose to give the raw history directly to the actor.

\textbf{Model pretraining.}
We next study the effect of pretraining the model before the agent starts learning on the task. 
In our experiments, the agent first collects a small amount of data
by executing random actions, and then the model is pretrained with that data. The model is pretrained for 50000 iterations on the DeepMind Control Suite experiments, unless otherwise specified. \autoref{fig:pretraining_steps_ablation} and \autoref{fig:pretraining_steps_ablation_all} show that little or no pretraining results in slower learning and, in some cases, worse asymptotic performance.
There is almost no difference in performance when using 100000 instead of 50000 iterations, although the former resulted in higher variance across trials in some of the tasks.
Overall, these results show that the agent benefits from the supervision signal of the model even before the agent has made any progress on the task.

\textbf{Training updates per iteration.}
We next investigate the effect of the number of training updates per iteration, or equivalently, the number of training updates per environment step (we use 1 environment step per iteration in all of our experiments).
\autoref{fig:train_steps_per_iteration_ablation} and \autoref{fig:train_steps_per_iteration_ablation_all} show that, in general, more training updates per iteration speeds up learning slightly, but too many updates per iteration causes higher variance across trials and slightly worse asymptotic performance in some tasks.
Nevertheless, this drop in asymptotic performance (if any) is small, which indicates that our method is less susceptible to overfitting compared to methods in prior work. We hypothesize that using stochastic latent samples to train the critic provides some randomization, which limits overfitting.
The best tradeoff is achieved when using 2 training updates per iteration, however, in line with other works, we use 1 training update per iteration in all the other experiments.

%% file: 05_conclusion.tex
\vspace{-2mm}
\section{Conclusion}
\vspace{-2mm}

We presented SLAC, an efficient RL algorithm for learning from high-dimensional image inputs that combines efficient off-policy model-free RL with representation learning via a sequential stochastic state space model. Through representation learning in conjunction with effective task learning in the learned latent space, our method achieves improved sample efficiency and final task performance as compared to both prior model-based and model-free RL methods.

While our current SLAC algorithm is fully model-free, in that predictions from the model are not utilized to speed up training, a natural extension of our approach would be to use the model predictions themselves to generate synthetic samples. Incorporating this additional synthetic model-based data into a mixed model-based and model-free method could further improve sample efficiency and performance. More broadly, the use of explicit representation learning with RL has the potential to not only accelerate training time and increase the complexity of achievable tasks, but also enable reuse and transfer of our learned representation across tasks.

%% file: 06_broader_impact.tex
\vspace{-2mm}
\section*{Broader Impact}
\vspace{-2mm}

Despite the existence of automated robotic systems in controlled environments such as factories or labs, standard approaches to controlling systems still require precise and expensive sensor setups to monitor the relevant details of interest in the environment, such as the joint positions of a robot or pose information of all objects in the area. To instead be able to learn directly from the more ubiquitous and rich modality of vision would greatly advance the current state of our learning systems. Not only would this ability to learn directly from images preclude expensive real-world setups, but it would also remove the expensive need for human-engineering efforts in state estimation. While it would indeed be very beneficial for our learning systems to be able to learn directly from raw image observations, this introduces algorithm challenges of dealing with high-dimensional as well as partially observable inputs. In this paper, we study the use of explicitly learning latent representations to assist model-free reinforcement learning directly from raw, high-dimensional images. 

Standard end-to-end RL methods try to solve both representation learning and task learning together, and in practice, this leads to brittle solutions which are sensitive to hyperparameters but are also slow and inefficient. These challenges illustrate the predominant use of simulation in the deep RL community; we hope that with more efficient, stable, easy-to-use, and easy-to-train deep RL algorithms such as the one we propose in this work, we can help the field of deep RL to transition to more widespread use in real-world setups such as robotics.

From a broader perspective, there are numerous use cases and areas of application where autonomous decision making agents can have positive effects in our society, from automating dangerous and undesirable tasks, to accelerating automation and economic efficiency of society. That being said, however, automated decision making systems do introduce safety concerns, further exacerbated by the lack of explainability when they do make mistakes. Although this work does not explicitly address safety concerns, we feel that it can be used in conjunction with levels of safety controllers to minimize negative impacts, while drawing on its powerful deep reinforcement learning roots to enable automated and robust tasks in the real world.

%% file: 07_appendix.tex
\section{Derivation of the Evidence Lower Bound and SLAC Objectives}
\label{app:elbo_derivation}

\begin{wrapfigure}{r}{0.45\columnwidth}
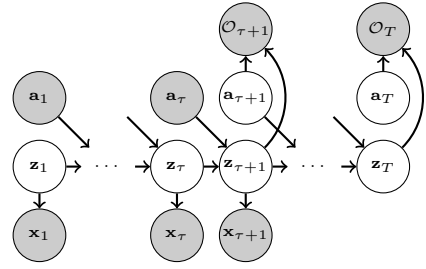

  \centering
  \vspace{-4mm}
  \includestandalone{figures/pgm}
  \vspace{-1mm}
  \caption{Graphical model of POMDP with optimality variables for $t \geq \tau+1$.}
  \label{fig:pgm_app}
  \vspace{-6mm}
\end{wrapfigure}
In this appendix, we discuss how the SLAC objectives can be derived from applying a variational inference scheme to the control as inference framework for reinforcement learning~\citep{levine2018reinforcement}. In this framework, the problem of finding the optimal policy is cast as an inference problem, conditioned on the evidence that the agent is behaving optimally. While \citet{levine2018reinforcement} derives this in the fully observed case, we present a derivation in the POMDP setting. For reference, we reproduce the probabilistic graphical model in \autoref{fig:pgm_app}.

\vspace{4mm}
We aim to maximize the marginal likelihood $p(\x{1:\tau+1}, \optim{\tau+1:T} | \a{1:\tau})$, where $\tau$ is the number of steps that the agent has already taken.
This likelihood reflects that the agent cannot modify the past $\tau$ actions and they might have not been optimal, but it can choose the future actions up to the end of the episode, such that the chosen future actions are optimal.
Notice that unlike the standard control as inference framework, in this work we not only maximize the likelihood of the optimality variables but also the likelihood of the observations, which provides additional supervision for the latent representation.
This does not come up in the MDP setting since the state representation is fixed and learning a dynamics model of the state would not change the model-free equations derived from the maximum entropy RL objective.

For reference, we restate the factorization of our variational distribution:
\begin{equation}
  \!\!q(\z{1:T}, \a{\tau+1:T} | \x{1:\tau+1}, \a{1:\tau}) = \prod_{t=0}^{\tau} q(\z{t+1} | \x{t+1}, \z{t}, \a{t}) \!\! \prod_{t=\tau+1}^{T-1} \!\! p(\z{t+1} | \z{t}, \a{t}) \!\! \prod_{t=\tau+1}^{T} \!\! \policy(\a{t} | \x{1:t}, \a{1:t-1}).\!\!\!
  \label{eq:our_posterior_app}
\end{equation}
As discussed by \citet{levine2018reinforcement}, the agent does not have control over the stochastic dynamics, so we use the dynamics $p(\z{t+1} | \z{t}, \a{t})$ for $t \geq \tau+1$ in the variational distribution in order to prevent the agent from choosing optimistic actions.

The joint likelihood is
\begin{equation}
  p(\x{1:\tau+1}, \optim{\tau+1:T}, \z{1:T}, \a{\tau+1:T} | \a{1:\tau})
  =\! \prod_{t=1}^{\tau+1} p(\x{t} | \z{t}) \! \prod_{t=0}^{T-1} p(\z{t+1} | \z{t}, \a{t}) \!\! \prod_{t=\tau+1}^T \!\! p(\optim{t} | \z{t}, \a{t}) \!\!\! \prod_{t=\tau+1}^T \!\!\! p(\a{t}).
  \label{eq:our_likelihood_app}
\end{equation}

We use the posterior from \autoref{eq:our_posterior_app}, the likelihood from \autoref{eq:our_likelihood_app}, and Jensen's inequality to obtain the ELBO of the marginal likelihood,
\begin{align}
  \MoveEqLeft \log p(\x{1:\tau+1}, \optim{\tau+1:T} | \a{1:\tau}) \nonumber\\
  &\!\!\!\!\!= \log \int\displaylimits_{\z{1:T}} \int\displaylimits_{\a{\tau+1:T}} p(\x{1:\tau+1}, \optim{\tau+1:T}, \z{1:T}, \a{\tau+1:T} | \a{1:\tau}) \diff \z{1:T} \diff \a{\tau+1:T} \\
  &\!\!\!\!\!\geq \! \E_{(\z{1:T}, \a{\tau+1:T}) \sim q} \! \left[ \vphantom{\sum} \log p(\x{1:\tau+1}, \optim{\tau+1:T}, \z{1:T}, \a{\tau+1:T} | \a{1:\tau}) - \log q(\z{1:T}, \a{\tau+1:T} | \x{1:\tau+1}, \a{1:\tau}) \right] \\[-1.5mm]
  &\!\!\!\!\!= \begin{multlined}[t]
    \E_{(\z{1:T}, \a{\tau+1:T}) \sim q} \left[ \vphantom{\sum_t^\tau}
      \smash{\underbrace{ 
        \sum_{t=0}^{\tau} \left( \vphantom{\sum} \log p(\x{t+1} | \z{t+1}) 
        - \kl{q(\z{t+1} | \x{t+1}, \z{t}, \a{t})}{p(\z{t+1} | \z{t}, \a{t})} \right)
      }_{\text{model objective terms}}}
    \right. \\[3.5mm]
    \left. \vphantom{\sum_t^\tau} {}+ 
      \smash{\underbrace{ 
        \sum_{t=\tau+1}^T \left( \vphantom{\sum} r(\z{t}, \a{t}) + \log p(\a{t}) - \log \policy(\a{t} | \x{1:t}, \a{1:t-1}) \right)
      }_{\text{policy objective terms}}}
    \right] \!,
  \end{multlined}
\end{align}

We are interested in the likelihood of optimal trajectories, so we use $\optim{t} = 1$ for $t \geq \tau+1$, and its distribution is given by $p(\optim{t} = 1 | \z{t}, \a{t}) = \exp(r(\z{t}, \a{t}))$ in the control as inference framework.
Notice that the dynamics terms $\log p(\z{t+1} | \z{t}, \a{t})$ for $t \geq \tau+1$ from the posterior and the prior cancel each other out in the ELBO.

The first part of the ELBO corresponds to the model objective. When using the parametric function approximators, the negative of it corresponds directly to the model loss in \autoref{eq:our_model_loss}.

The second part of the ELBO corresponds to the maximum entropy RL objective. We assume a uniform action prior, so the $\log p(\a{t})$ term is a constant term that can be omitted when optimizing this objective. We use message passing to optimize this objective, with messages defined as 
\vspace{-1mm}
\begin{align}
  Q(\z{t}, \a{t}) &= r(\z{t}, \a{t}) + \E_{\z{t+1} \sim q(\cdot | \x{t+1}, \z{t}, \a{t})} \left[ V(\z{t+1}) \vphantom{\sum} \right] \label{eq:Q} \\
  V(\z{t}) &= \log \int\displaylimits_{\a{t}} \exp(Q(\z{t}, \a{t})) \diff \a{t}. \label{eq:V}
\end{align}
\vspace{-1mm}
Then, the maximum entropy RL objective can be expressed in terms of the messages as
\begin{align}
  \MoveEqLeft \E_{(\z{\tau+1:T}, \a{\tau+1:T}) \sim q} \left[ \sum_{t=\tau+1}^T \left( \vphantom{\sum} r(\z{t}, \a{t}) - \log \policy(\a{t} | \x{1:t}, \a{1:t-1}) \right) \right] \nonumber\\[1mm]
  &\!\!\!\!\!= \E_{\z{\tau+1} \sim q(\cdot | \x{\tau+1}, \z{\tau}, \a{\tau})} \left[ \E_{\a{\tau+1} \sim \policy(\cdot | \x{1:\tau+1}, \a{1:\tau})} \left[ \vphantom{\sum} Q(\z{\tau+1}, \a{\tau+1}) - \log \policy(\a{\tau+1} | \x{1:\tau+1}, \a{1:\tau}) \right] \right] \label{eq:rl_objective_dynamic_programming}\\[1mm]
  &\!\!\!\!\!= \E_{\a{\tau+1} \sim \policy(\cdot | \x{1:\tau+1}, \a{1:\tau})} \left[ \E_{\z{\tau+1} \sim q(\cdot | \x{\tau+1}, \z{\tau}, \a{\tau})} \left[ \vphantom{\sum} Q(\z{\tau+1}, \a{\tau+1}) \right] - \log \policy(\a{\tau+1} | \x{1:\tau+1}, \a{1:\tau}) \right] \label{eq:rl_objective_expectation_swap} \\[1mm]
  &\!\!\!\!\!= -\kl{\policy(\a{\tau+1} | \x{1:\tau+1}, \a{1:\tau})}{\frac{\exp{ \left( \E_{\z{\tau+1} \sim q} \left[ Q(\z{\tau+1}, \a{\tau+1}) \right] \right) }}{\exp{ \left( \E_{\z{\tau+1} \sim q} \left[ V(\z{\tau+1}) \right] \right) }}} + \E_{\z{\tau+1} \sim q} \left[ \vphantom{\sum} V(\z{\tau+1}) \right],
  \label{eq:rl_objective_kl}
\end{align}
where the first equality is obtained from dynamic programming (see \citet{levine2018reinforcement} for details), the second equality is obtain by swapping the order of the expectations, the third from the definition of KL divergence, and $\E_{\z{t} \sim q} \left[ V(\z{t}) \right]$ is the normalization factor for $\E_{\z{t} \sim q} \left[ Q(\z{t}, \a{t}) \right]$ with respect to $\a{t}$.
Since the KL divergence term is minimized when its two arguments represent the same distribution, the optimal policy is given by
\vspace{-1.5mm}
\begin{equation}
  \policy(\a{t} | \x{1:t}, \a{1:t-1}) = \exp{ \left( \E_{\z{t} \sim q} \left[ \vphantom{\sum} Q(\z{t}, \a{t}) - V(\z{t}) \right] \right) }.
  \label{eq:optimal_policy}
\end{equation}
That is, the optimal policy is optimal with respect to the expectation over the belief of the Q value of the learned MDP. This is equivalent to the Q-MDP heuristic, which amounts to assuming that any uncertainty in the belief is gone after the next action~\citep{littman1995learning}.

Noting that the KL divergence term is zero for the optimal action, the equality from \autoref{eq:rl_objective_dynamic_programming} and \autoref{eq:rl_objective_kl} can be used in \autoref{eq:Q} to obtain
\vspace{-1.5mm}
\begin{multline}
  \!\!\! Q(\z{t}, \a{t}) = r(\z{t}, \a{t}) + \E_{\z{t+1} \sim q(\cdot | \x{t+1}, \z{t}, \a{t})} \left[ \E_{\a{t+1} \sim \policy(\cdot | \x{1:t+1}, \a{1:t})} \left[ \vphantom{\sum} Q(\z{t+1}, \a{t+1}) \right. \right.\\
  \left. \left. {}- \log \policy(\a{t+1} | \x{1:t+1}, \a{1:t}) \vphantom{\sum} \right] \vphantom{\E_{\a{t+1}}} \right] \!. \!\!\!\!
  \label{eq:soft_bellman_backup}
\end{multline}
This equation corresponds to the Bellman backup with a soft maximization for the value function.

As mentioned in \autoref{sec:slac}, our algorithm conditions the parametric policy in the history of observations and actions, which allows us to directly execute the policy without having to perform inference on the latent state at run time.
When using the parametric function approximators, the negative of the maximum entropy RL objective, written as in \autoref{eq:rl_objective_dynamic_programming}, corresponds to the policy loss in \autoref{eq:our_actor_loss}. Lastly, the Bellman backup of \autoref{eq:soft_bellman_backup} corresponds to the Bellman residual in \autoref{eq:our_critic_loss} when approximated by a regression objective.

We showed that the SLAC objectives can be derived from applying variational inference in the control as inference framework in the POMDP setting. This leads to the joint likelihood of the past observations and future optimality variables, which we aim to optimize by maximizing the ELBO of the log-likelihood. We decompose the ELBO into the model objective and the maximum entropy RL objective. We express the latter in terms of messages of Q-functions, which in turn are learned by minimizing the Bellman residual. These objectives lead to the model, policy, and critic losses.

\vspace{-2mm}
\section{Latent Variable Factorization and Network Architectures}
\vspace{-2mm}
\label{app:network_architectures}

\begin{wrapfigure}{r}{0.45\columnwidth}
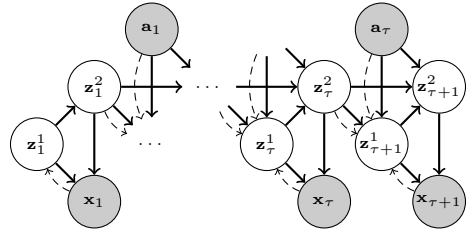

  \centering
  \vspace{-4mm}
  \includestandalone{figures/lvm_pgm}
  \caption{Diagram of our full model. Solid arrows show the generative model, dashed arrows show the inference model. Rewards are not shown for clarity.}
  \label{fig:lvm_pgm}
  \vspace{-4mm}
\end{wrapfigure}

In this section, we describe the architecture of our sequential latent variable model. Motivated by the recent success of autoregressive latent variables in VAEs~\citep{razavi2019vqvae2,maaloe2019biva}, we factorize the latent variable $\z{t}$ into two stochastic variables, $\z{t}^1$ and $\z{t}^2$, as shown in \autoref{fig:lvm_pgm}. This factorization results in latent distributions that are more expressive, and it allows for some parts of the prior and posterior distributions to be shared. We found this design to provide a good balance between ease of training and expressivity, producing good reconstructions and generations and, crucially, providing good representations for reinforcement learning. Note that the diagram in \autoref{fig:lvm_pgm} represents the \emph{Bayes net} corresponding to our full model. However, since all of the latent variables are stochastic, this visualization also presents the design of the computation graph. Inference over the latent variables is performed using amortized variational inference, with all training done via reparameterization. Hence, the computation graph can be deduced from the diagram by treating all solid arrows as part of the generative model and all dashed arrows as part of approximate posterior.

The generative model consists of the following probability distributions:
\vspace{-0.5mm}
\begin{align*}
  \z{1}^1 &\sim p(\z{1}^1) \\
  \z{1}^2 &\sim p_\mparams(\z{1}^2 | \z{1}^1) \\
  \z{t+1}^1 &\sim p_\mparams(\z{t+1}^1 | \z{t}^2, \a{t}) \\
  \z{t+1}^2 &\sim p_\mparams(\z{t+1}^2 | \z{t+1}^1, \z{t}^2, \a{t}) \\
  \x{t} &\sim p_\mparams(\x{t} | \z{t}^1, \z{t}^2) \\
  \r{t} &\sim p_\mparams(\r{t} | \z{t}^1, \z{t}^2, \a{t}, \z{t+1}^1, \z{t+1}^2).
\end{align*}
The initial distribution $p(\z{1}^1)$ is a multivariate standard normal distribution $\mathcal{N}(\vec{0}, \mat{I})$. All of the other distributions are conditional and parameterized by neural networks with parameters $\mparams{}$.
The networks for $p_\mparams(\z{1}^2 | \z{1}^1)$, $p_\mparams(\z{t+1}^1 | \z{t}^2, \a{t})$, $p_\mparams(\z{t+1}^2 | \z{t+1}^1, \z{t}^2, \a{t})$, and $p_\mparams(\r{t} | \z{t}, \a{t}, \z{t+1})$ consist of two fully connected layers, each with 256 hidden units, and a Gaussian output layer. The Gaussian layer is defined such that it outputs a multivariate normal distribution with diagonal variance, where the mean is the output of a linear layer and the diagonal standard deviation is the output of a fully connected layer with softplus non-linearity.
The pre-transformed standard deviation right before the softplus non-linearity is gradient clipped element-wise by value to within $[-10, 10]$ during the backward pass.
The observation model $p_\mparams(\x{t} | \z{t})$ consists of 5 transposed convolutional layers (256 $4 \times 4$, 128 $3\times 3$, 64 $3 \times 3$, 32 $3 \times 3$, and 3 $5 \times 5$ filters, respectively, stride 2 each, except for the first layer). The output variance for each image pixel is fixed to a constant $\sigma^2$, which is a hyperparameter $\sigma^2 \in \{0.04, 0.1, 0.4\}$ on DeepMind Control Suite and $\sigma^2 = 0.1$ on OpenAI Gym.

The variational distribution $q$, also referred to as the inference model or the posterior, is represented by the following factorization:
\vspace{-1mm}
\begin{align*}
  \z{1}^1 &\sim q_\mparams(\z{1}^1 | \x{1}) \\
  \z{1}^2 &\sim p_\mparams(\z{1}^2 | \z{1}^1) \\
  \z{t+1}^1 &\sim q_\mparams(\z{t+1}^1 | \x{t+1}, \z{t}^2, \a{t}) \\
  \z{t+1}^2 &\sim p_\mparams(\z{t+1}^2 | \z{t+1}^1, \z{t}^2, \a{t}).
\end{align*} 
The networks representing the distributions $q_\mparams(\z{1}^1 | \x{1})$ and $q_\mparams(\z{t+1}^1 | \x{t+1}, \z{t}^2, \a{t})$ both consist of 5 convolutional layers (32 $5 \times 5$, 64 $3 \times 3$, 128 $3 \times 3$, 256 $3 \times 3$, and 256 $4 \times 4$ filters, respectively, stride 2 each, except for the last layer), 2 fully connected layers (256 units each), and a Gaussian output layer. The parameters of the convolution layers are shared among both distributions.

Note that the variational distribution over $\z{1}^2$ and $\z{t+1}^2$ is intentionally chosen to exactly match the generative model $p$, such that this term does not appear in the KL-divergence within the ELBO, and a separate variational distribution is only learned over $\z{1}^1$ and $\z{t+1}^1$. In particular, the KL-divergence over $\z{t+1}$ simplifies to the KL-divergence over $\z{t+1}^1$:
\begin{align}
  \MoveEqLeft \kl{q(\z{t+1} | \x{t+1}, \z{t}, \a{t})}{p(\z{t+1} | \z{t}, \a{t})} \\[1mm]
  &\!\!\!\!\!= \E_{\z{t+1} \sim q(\cdot | \x{t+1}, \z{t}, \a{t})} \left[ \log q(\z{t+1} | \x{t+1}, \z{t}, \a{t}) - \log p(\z{t+1} | \z{t}, \a{t}) \vphantom{\sum} \right]\\
  &\!\!\!\!\!= \begin{multlined}[t]
    \E_{\z{t+1}^1 \sim q(\cdot | \x{t+1}, \z{t}^2, \a{t})} \! \left[ \E_{\z{t+1}^2 \sim p(\cdot | \z{t+1}^1, \z{t}^2, \a{t})} \! \left[
      \log q(\z{t+1}^1 | \x{t+1}, \z{t}^2, \a{t})
    \vphantom{\sum} \right. \right. \\[-1mm]
    \left. \left.
      {}+ \log p(\z{t+1}^2 | \z{t+1}^1, \z{t}^2, \a{t})
      - \log p(\z{t+1}^1 | \z{t}^2, \a{t})
      - \log p(\z{t+1}^2 | \z{t+1}^1, \z{t}^2, \a{t})
    \vphantom{\sum} \right] \vphantom{\E_{\z{t+1}^2}} \right]
  \end{multlined} \\
  &\!\!\!\!\!= \E_{\z{t+1}^1 \sim q(\cdot | \x{t+1}, \z{t}^2, \a{t})} \! \left[
    \log q(\z{t+1}^1 | \x{t+1}, \z{t}^2, \a{t})
    - \log p(\z{t+1}^1 | \z{t}^2, \a{t})
  \vphantom{\sum} \right] \\[1mm]
  &\!\!\!\!\!= \kl{\log q(\z{t+1}^1 | \x{t+1}, \z{t}^2, \a{t})}{\log p(\z{t+1}^1 | \z{t}^2, \a{t})}.
\end{align}
This intentional design decision simplifies the training process.

The latent variables have 32 and 256 dimensions, respectively, i.e. $\z{t}^1 \in \mathbb{R}^{32}$ and $\z{t}^2 \in \mathbb{R}^{256}$.
For the image observations, $\x{t} \in [0, 1]^{64 \times 64 \times 3}$.
All the layers, except for the output layers, use leaky ReLU non-linearities. Note that there are no deterministic recurrent connections in the network---all networks are feedforward, and the temporal dependencies all flow through the stochastic units $\z{t}^1$ and $\z{t}^2$.

For the reinforcement learning process, we use a critic network $Q_\Qparams{}$ consisting of 2 fully connected layers (256 units each) and a linear output layer. The actor network $\policy_\pparams{}$ consists of 5 convolutional layers, 2 fully connected layers (256 units each), a Gaussian layer, and a tanh bijector,  which constrains the actions to be in the bounded action space of $[-1, 1]$. The convolutional layers are shared with the ones from the latent variable model, but the parameters of these layers are only updated by the model objective and not by the actor objective.

\vspace{-1mm}
\section{Training and Evaluation Details}
\vspace{-1mm}
\label{app:train_eval}

Before the agent starts learning on the task, the model is first pretrained using a small amount of random data.
The DeepMind Control Suite experiments pretrains the model for 50000 iterations, using random data from 10 episodes, and random actions that are sampled from a tanh-transformed Gaussian distribution with zero mean and a scale of 2, i.e. $\a{} = \tanh{\tilde{\a{}}}$, where $\tilde{\a{}} \sim \mathcal{N}(0, 2^2)$.
The OpenAI Gym experiments pretrains the model for 100000 iterations, using random data from 10000 agent steps, and uniformly distributed random actions. Note that this data is taken into account in our plots.

The control portion of our algorithm uses the same hyperparameters as SAC~\citep{haarnoja2018soft}, except for a smaller replay buffer size of 100000 environment steps (instead of a million) due to the high memory usage of image observations.

The network parameters are initialized using the default initialization distributions. In the case of the DeepMind Control Suite experiments, the scale of the policy's pre-transformed Gaussian distribution is scaled by 2.
This, as well as the initial tanh-transformed Gaussian policy, contributes to trajectories with larger actions (i.e. closer to $-1$ and $1$) at the beginning of training. This didn't make a difference for the DeepMind Control Suite tasks \emph{except} for the walker task, where we observed that this initialization resulted in less variance across trials and avoided trials that would otherwise get stuck in local optima early in training.

All of the parameters are trained with the Adam optimizer~\citep{kingma2015adam}, and we perform 1 gradient step per environment step for DeepMind Control Suite and 3 gradient steps per environment step for OpenAI Gym.
The Q-function and policy parameters are trained with a learning rate of 0.0003 and a batch size of 256. The model parameters are trained with a learning rate of 0.0001 and a batch size of 32. We use fixed-length sequences of length 8, rather than all the past observations and actions within the episode.

We use action repeats for all the methods, except for D4PG for which we use the reported results from prior work~\citep{tassa2018deepmind}.
The number of environment steps reported in our plots correspond to the unmodified steps of the benchmarks. Note that the methods that use action repeats only use a fraction of the environment steps reported in our plots. For example, 1 million environment steps of the cheetah task correspond to 250000 samples when using an action repeat of 4.
The action repeats used in our experiments are given in \autoref{tab:action_repeats}.

Unlike in prior work~\citep{haarnoja2018soft,haarnoja2018applications}, we use the same stochastic policy as both the behavioral and evaluation policy since we found the deterministic greedy policy to be comparable or worse than the stochastic policy.

Our plots show results over multiple trials (i.e. seeds), and each trial computes average returns from 10 evaluation episodes. We used 10 trials for the DeepMind Control Suite experiments and 5 trials for the OpenAI Gym experiments.
In the case of the DeepMind Control Suite experiments, we sweep over $\sigma^2 \in \{0.04, 0.1, 0.4\}$ and plot the results corresponding to the hyperparameter $\sigma^2$ that achieves the best per-task average return across trials averaged over the first half a million environment steps. In \autoref{fig:dm_control_plots}, the best $\sigma^2$ values are 0.1, 0.4, 0.04, and 0.1 for the cheetah run, walker walk, ball-in-cup catch, and finger spin tasks, respectively.

\begin{table}
  \centering
  \begin{tabular}{l l l l l}
  \toprule
  Benchmark & Task & \makecell{Action \\ repeat} & \makecell{Original control \\ time step (s)} & \makecell{Effective control \\ time step (s)} \\
  \midrule
  \multirow{4}{*}{DeepMind Control Suite}
  & Cheetah, run       & 4 & 0.01  & 0.04 \\
  & Walker, walk       & 2 & 0.025 & 0.05 \\
  & Ball in cup, catch & 4 & 0.02  & 0.08 \\
  & Finger, spin       & 1 & 0.02  & 0.02 \\
  & Cartpole, swingup  & 4 & 0.01  & 0.04 \\
  & Reacher, easy      & 4 & 0.02  & 0.08 \\
  \cdashlinelr{1-5}
  \multirow{4}{*}{OpenAI Gym}
  & HalfCheetah-v2    & 1 & 0.05  & 0.05 \\
  & Walker2d-v2       & 4 & 0.008 & 0.032 \\
  & Hopper-v2         & 2 & 0.008 & 0.016 \\
  & Ant-v2            & 4 & 0.05  & 0.2 \\
  \bottomrule
  \end{tabular}
  \vspace{2mm}
  \caption{Action repeats and the corresponding agent's control time step used in our experiments.}
  \label{tab:action_repeats}
\end{table}

\section{Ablation Experiments}
\label{app:ablations}
\begin{figure}[H]
  \centering
  \includegraphics[scale=0.37, trim={2mm 8mm 2.5mm 2.5mm}, clip]{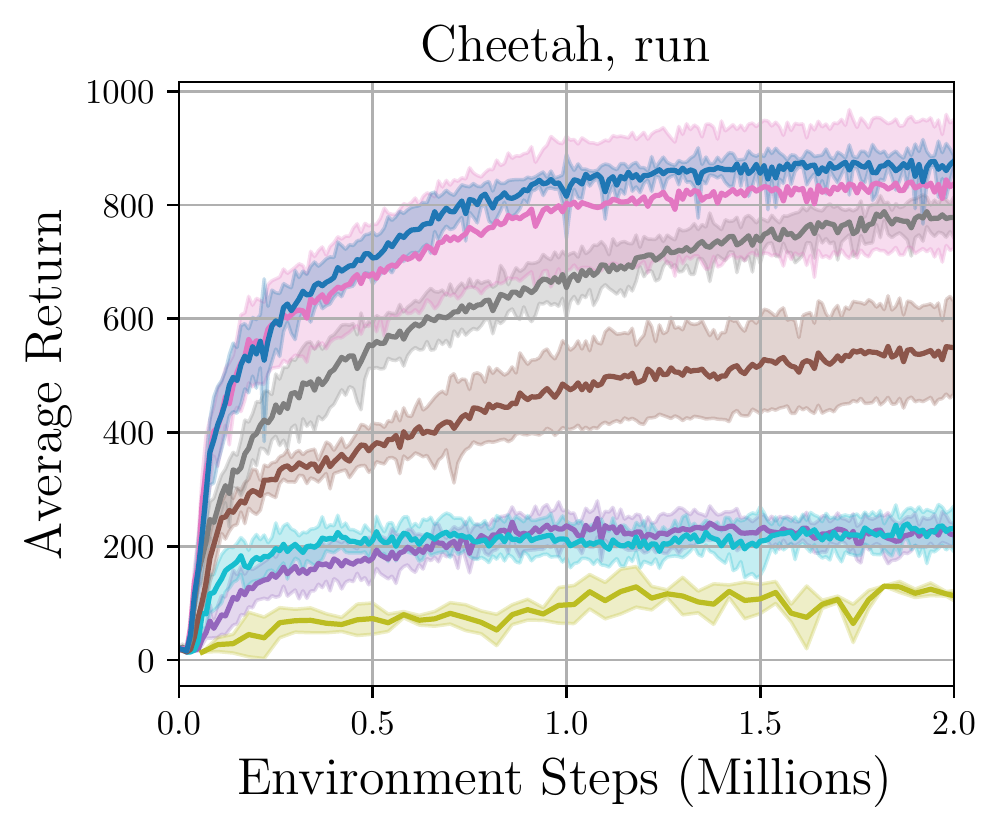} \hspace{1mm}
  \includegraphics[scale=0.37, trim={8.5mm 8mm 2.5mm 2.5mm}, clip]{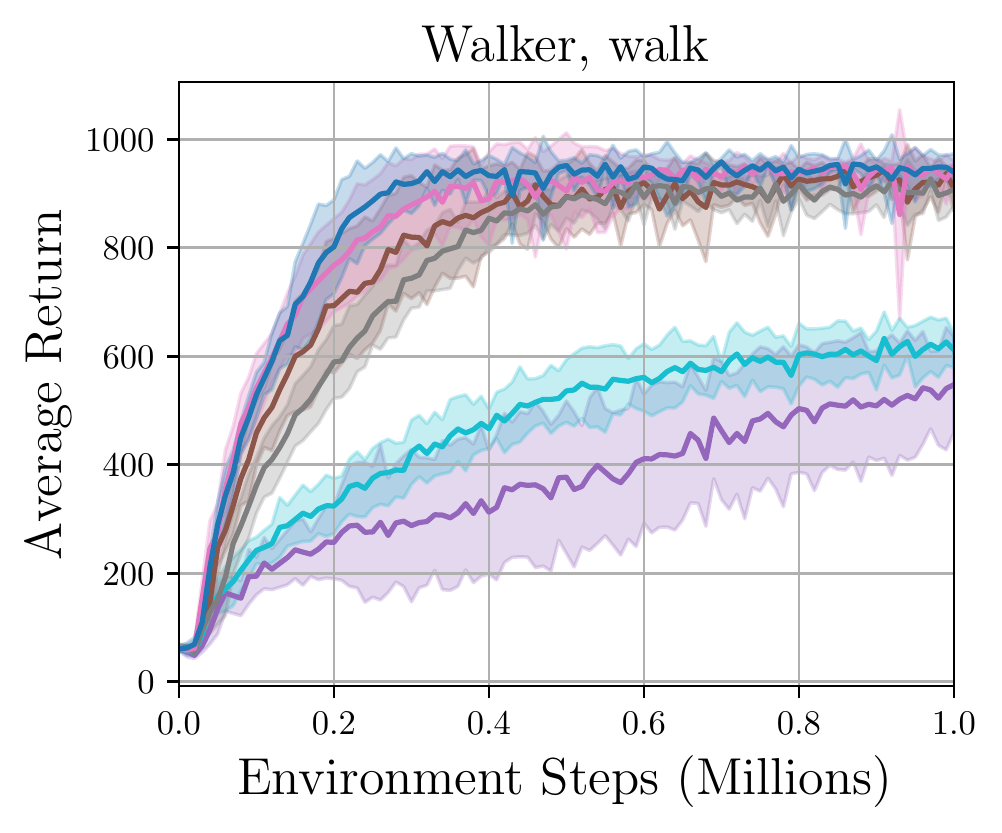} \hspace{1mm}
  \includegraphics[scale=0.37, trim={8.5mm 8mm 2.5mm 2.5mm}, clip]{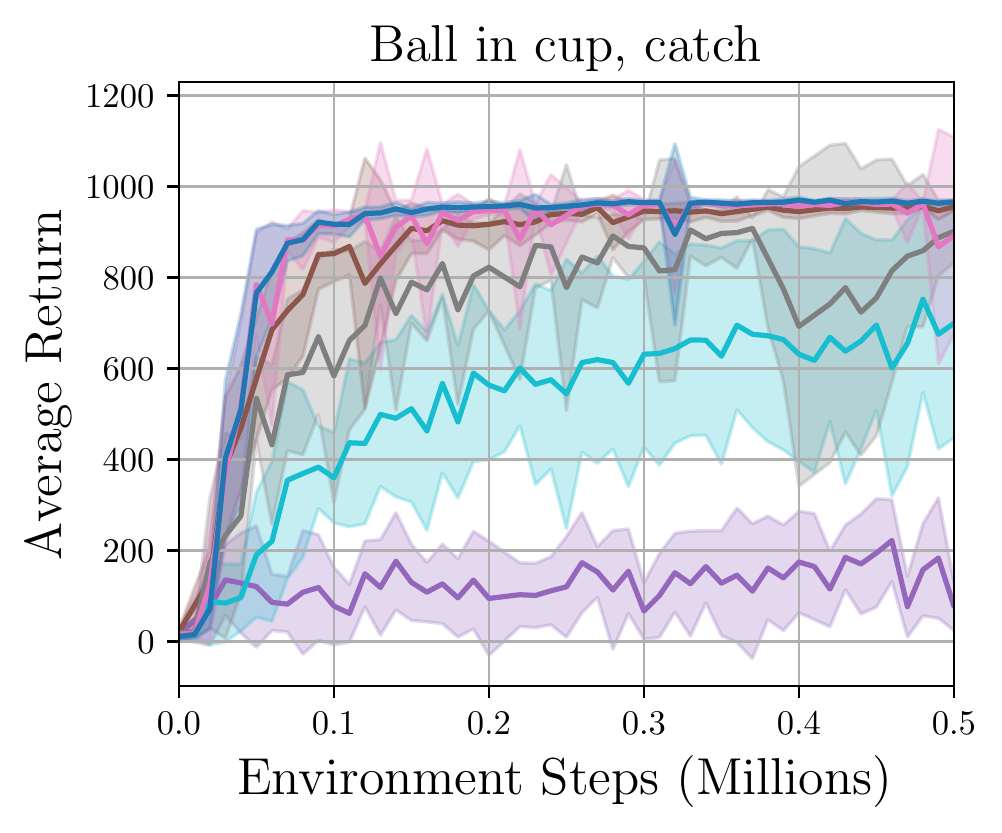} \\[1mm]
  \includegraphics[scale=0.37, trim={2mm 2.5mm 2.5mm 2.5mm}, clip]{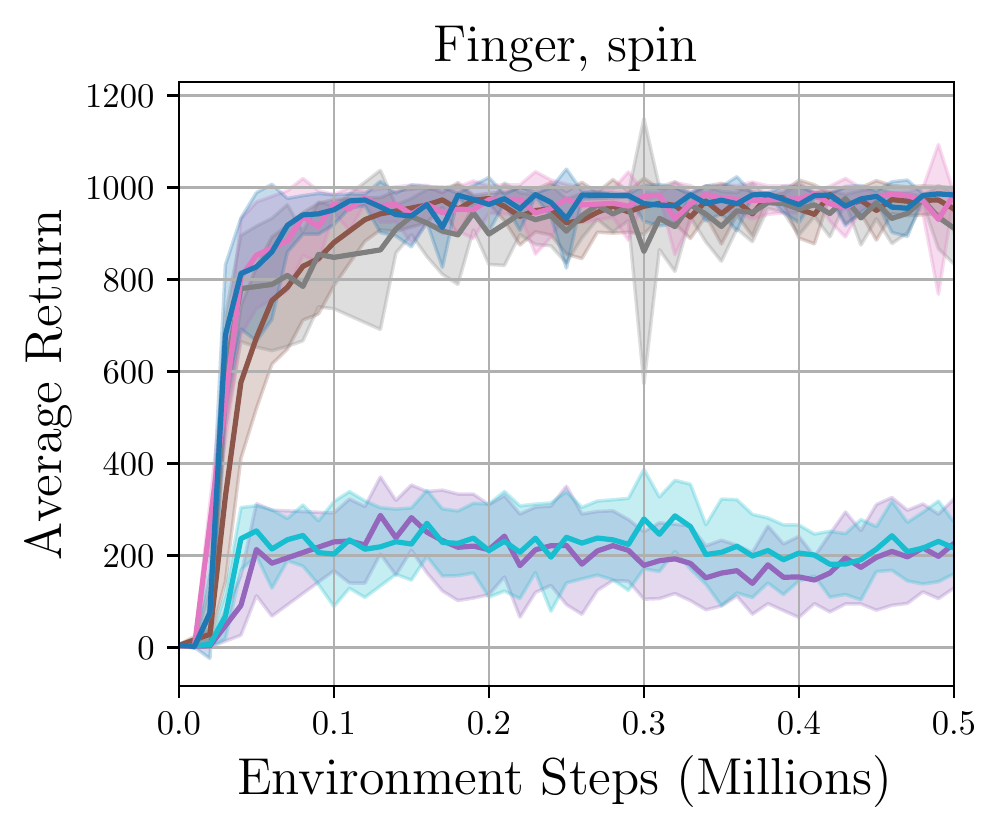} \hspace{1mm}
  \includegraphics[scale=0.37, trim={8.5mm 2.5mm 2.5mm 2.5mm}, clip]{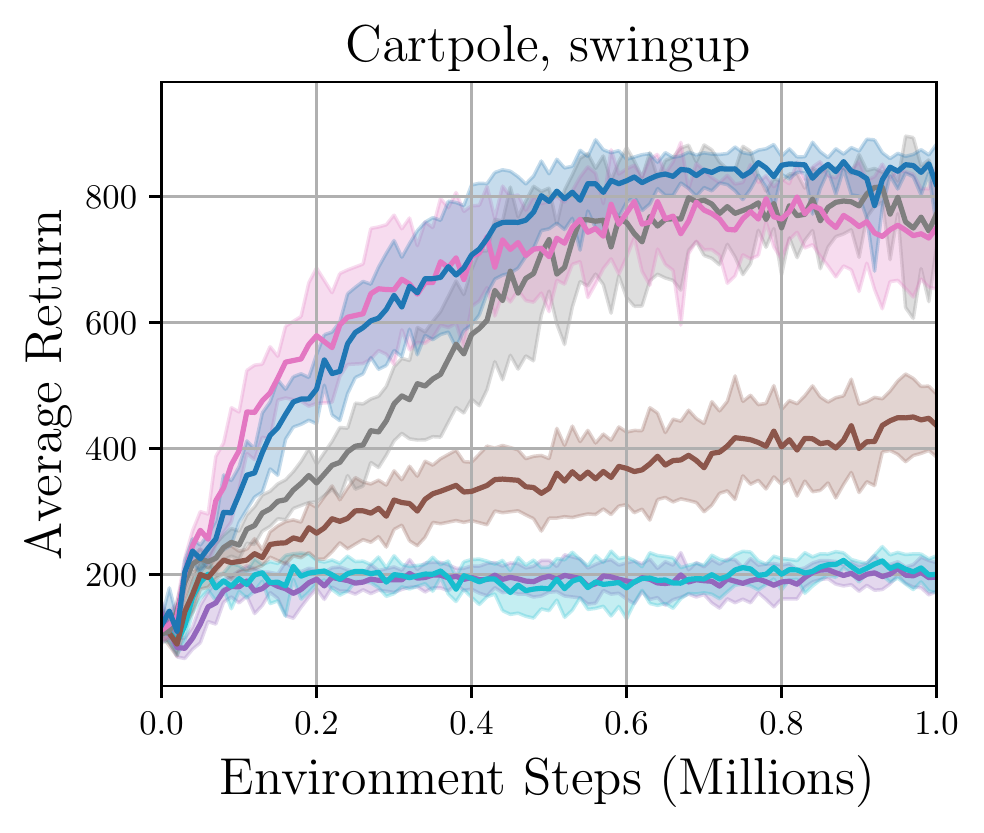} \hspace{1mm}
  \includegraphics[scale=0.37, trim={8.5mm 2.5mm 2.5mm 2.5mm}, clip]{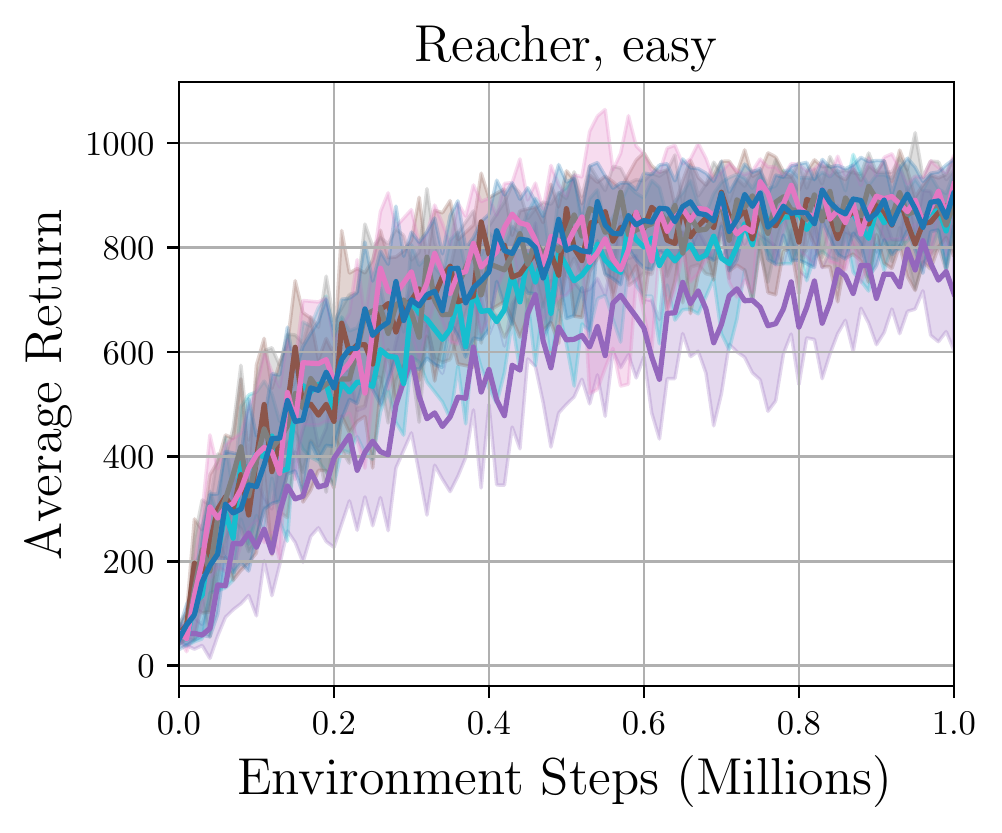} \\
  \includegraphics[scale=0.35]{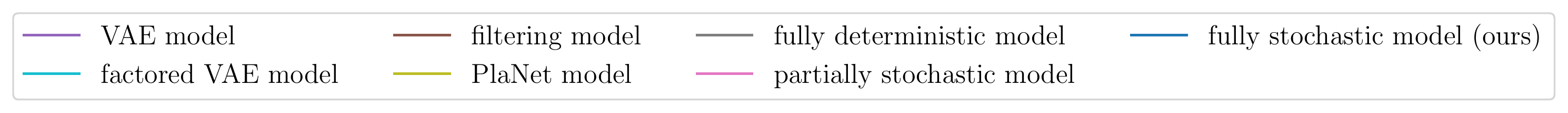}
  \caption{Comparison of different design choices for the latent variable model.
  In all cases, we use the RL framework of SLAC and only vary the choice of model for representation learning.
  These results show that including temporal dependencies leads to the largest improvement in performance, followed by the autoregressive latent variable factorization and using a fully stochastic model.
  }
  \label{fig:model_ablation_all}
\end{figure}

\begin{figure}[H]
  \centering
  \includegraphics[scale=0.37, trim={2mm 8mm 2.5mm 2.5mm}, clip]{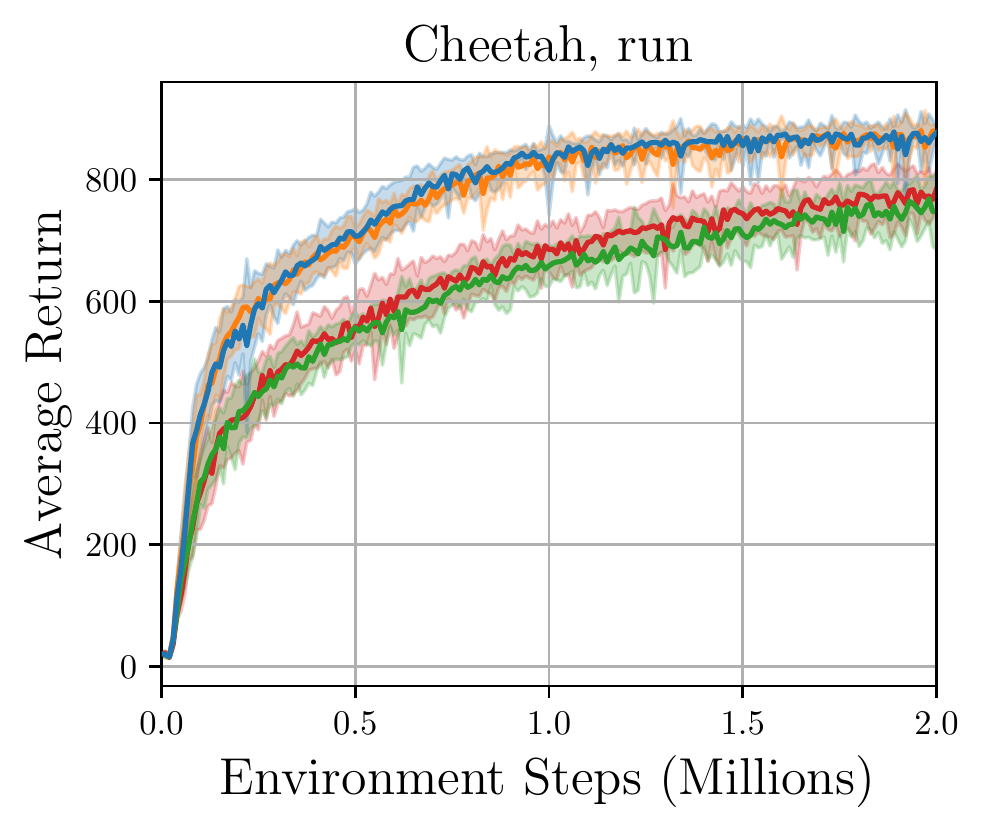} \hspace{1mm}
  \includegraphics[scale=0.37, trim={8.5mm 8mm 2.5mm 2.5mm}, clip]{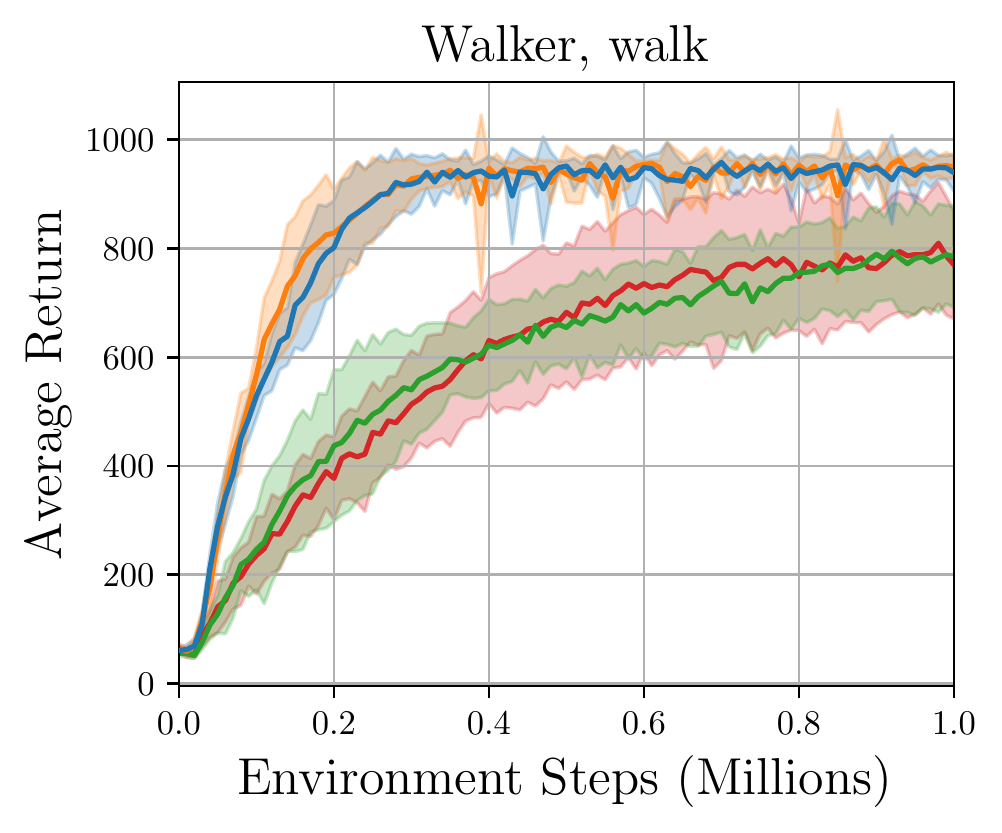} \hspace{1mm}
  \includegraphics[scale=0.37, trim={8.5mm 8mm 2.5mm 2.5mm}, clip]{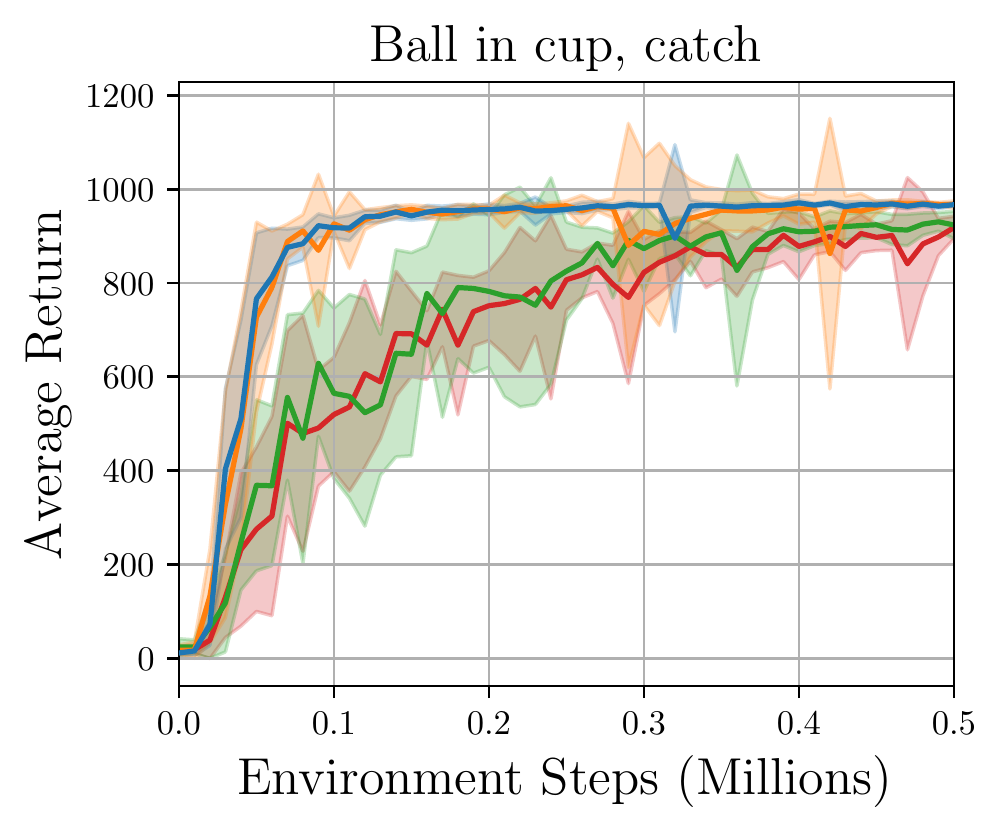} \\[1mm]
  \includegraphics[scale=0.37, trim={2mm 2.5mm 2.5mm 2.5mm}, clip]{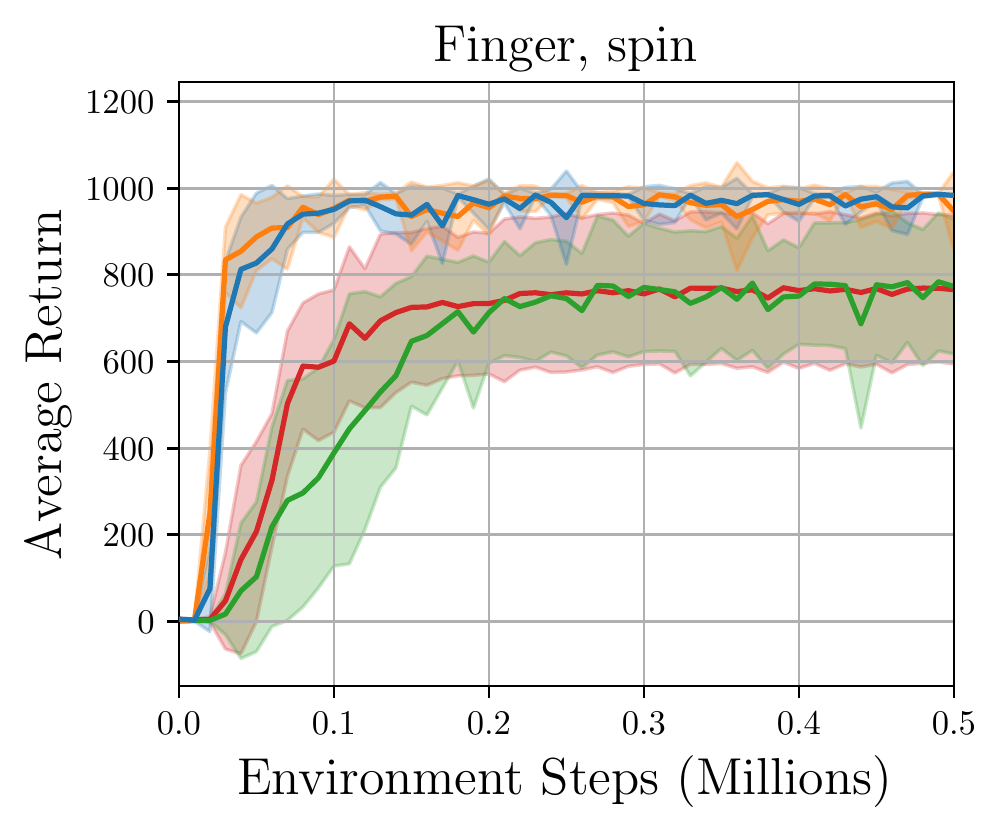} \hspace{1mm}
  \includegraphics[scale=0.37, trim={8.5mm 2.5mm 2.5mm 2.5mm}, clip]{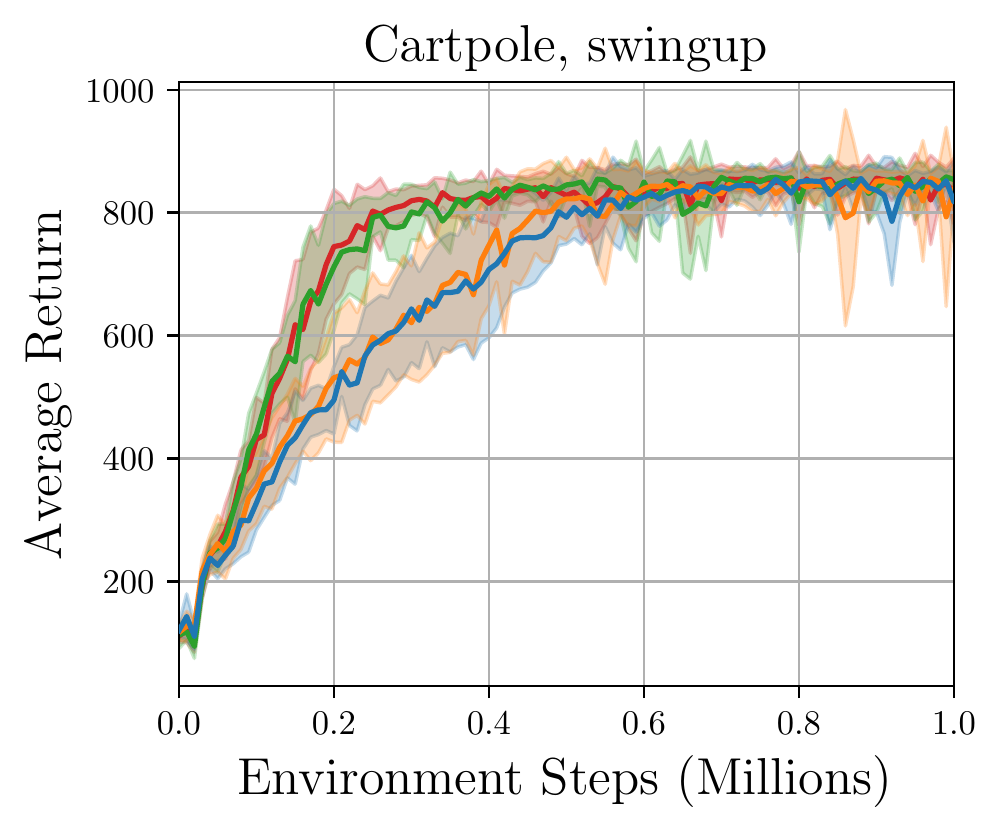} \hspace{1mm}
  \includegraphics[scale=0.37, trim={8.5mm 2.5mm 2.5mm 2.5mm}, clip]{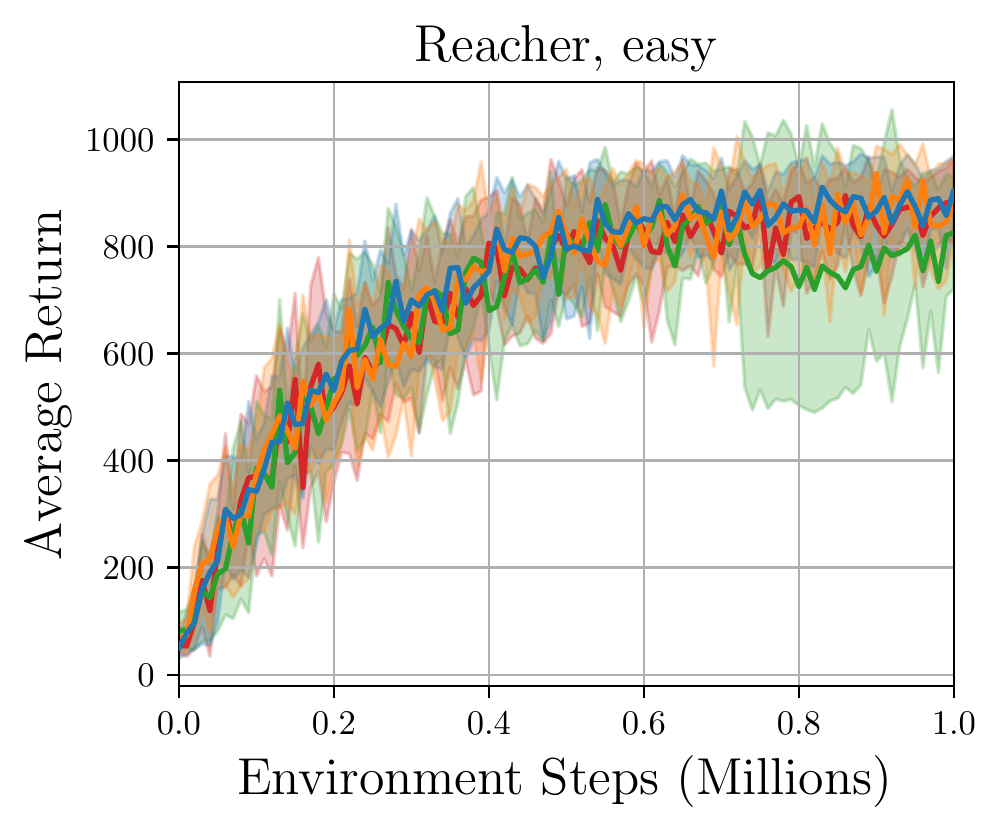} \\
  \includegraphics[scale=0.35]{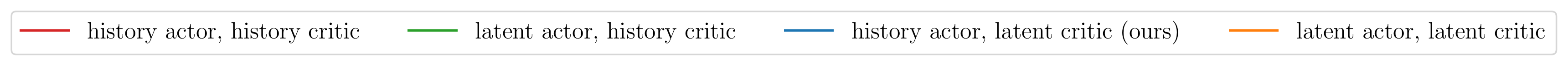}
  \caption{Comparison of alternative choices for the actor and critic inputs as either the observation-action history or the latent sample.
  With the exception of the cartpole swingup and reacher easy tasks, the performance is significantly worse when the critic input is the history instead of the latent sample, and indifferent to the choice for the actor input.
  }
  \label{fig:actor_input_critic_input_ablation_all}
\end{figure}

\begin{figure}[H]
  \centering
  \includegraphics[scale=0.37, trim={2mm 8mm 2.5mm 2.5mm}, clip]{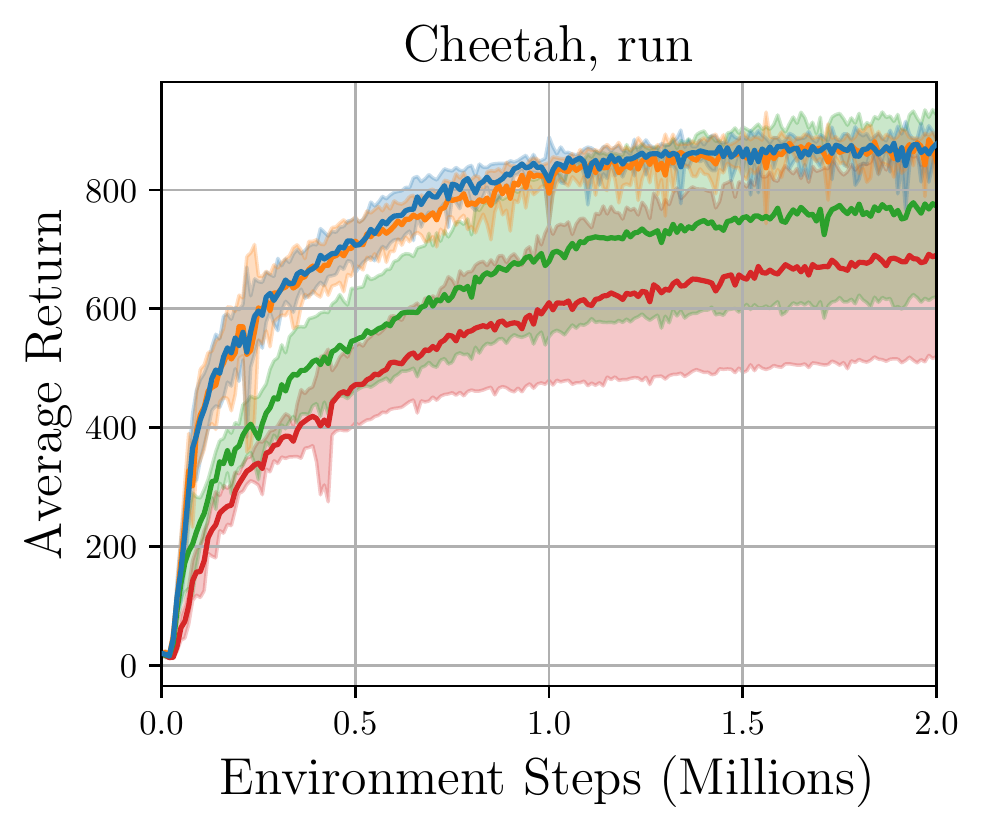} \hspace{1mm}
  \includegraphics[scale=0.37, trim={8.5mm 8mm 2.5mm 2.5mm}, clip]{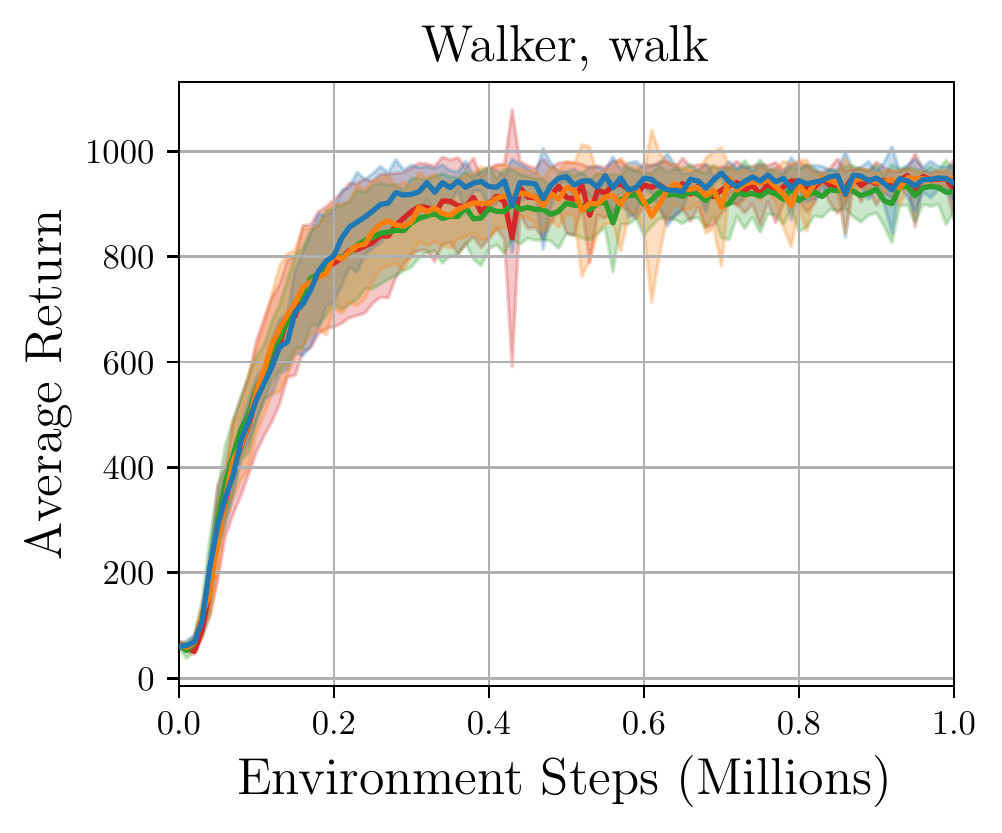} \hspace{1mm}
  \includegraphics[scale=0.37, trim={8.5mm 8mm 2.5mm 2.5mm}, clip]{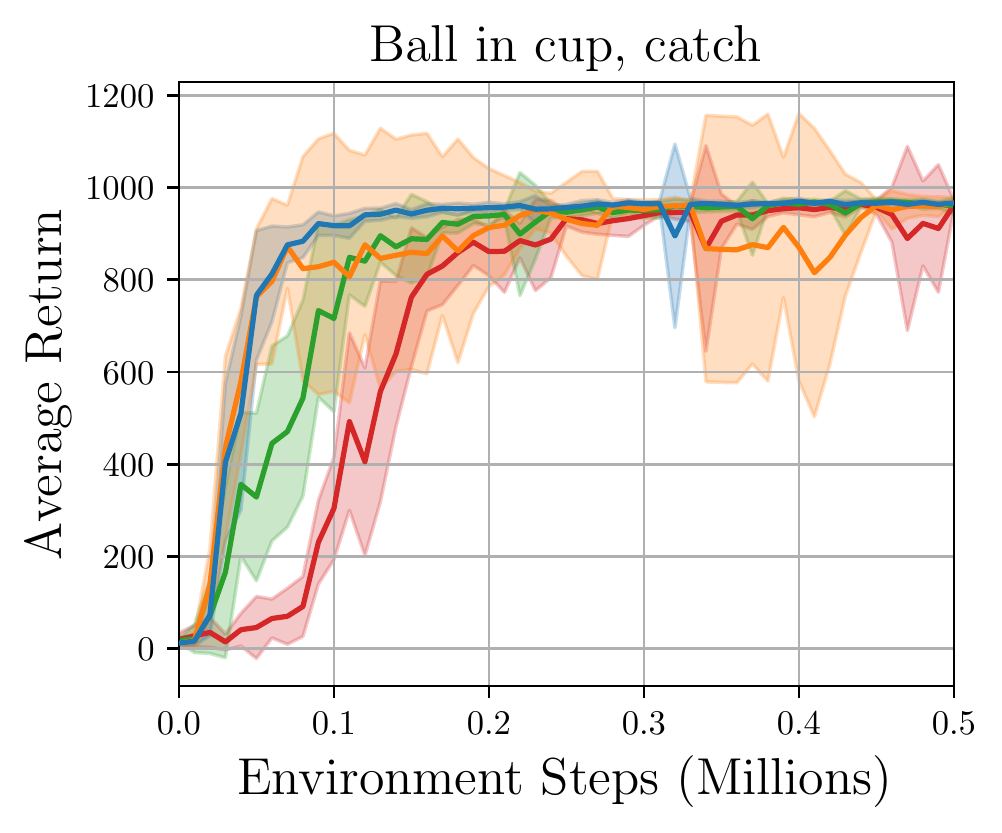} \\[1mm]
  \includegraphics[scale=0.37, trim={2mm 2.5mm 2.5mm 2.5mm}, clip]{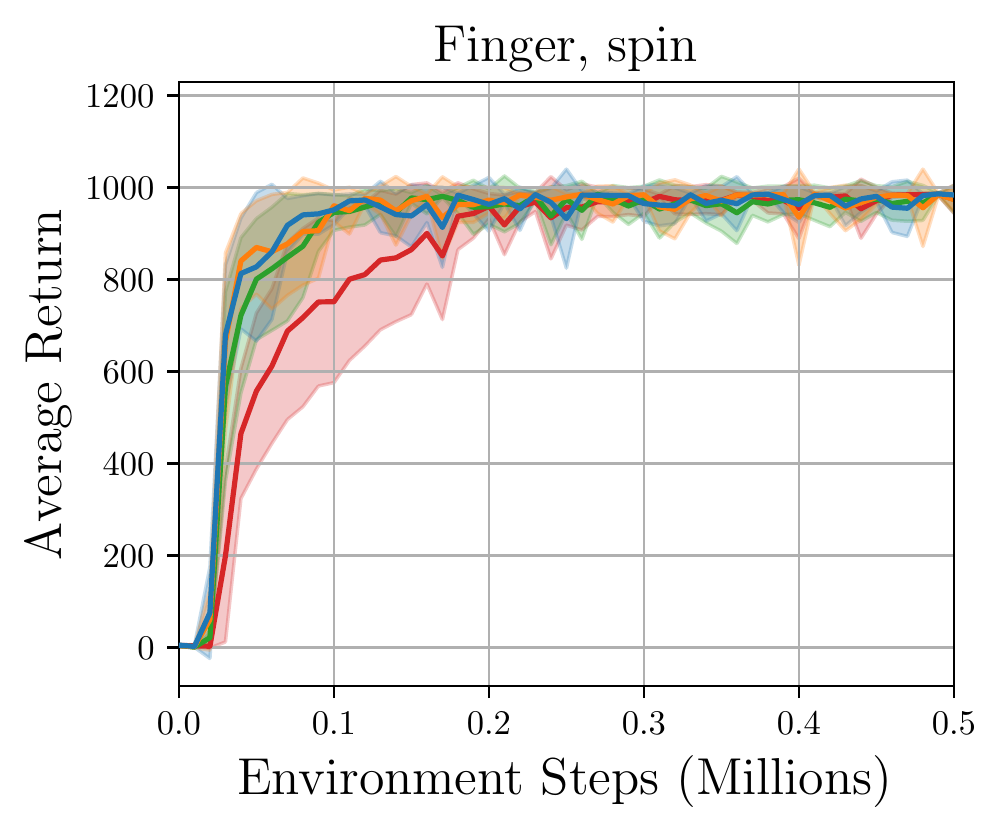} \hspace{1mm}
  \includegraphics[scale=0.37, trim={8.5mm 2.5mm 2.5mm 2.5mm}, clip]{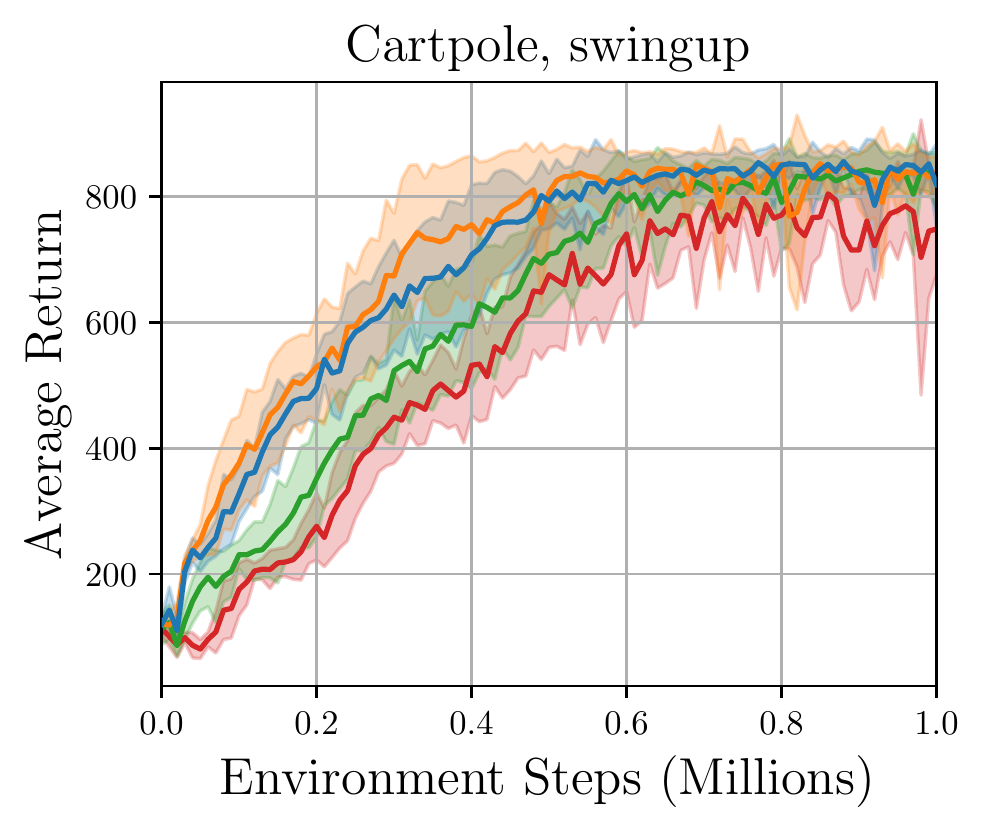} \hspace{1mm}
  \includegraphics[scale=0.37, trim={8.5mm 2.5mm 2.5mm 2.5mm}, clip]{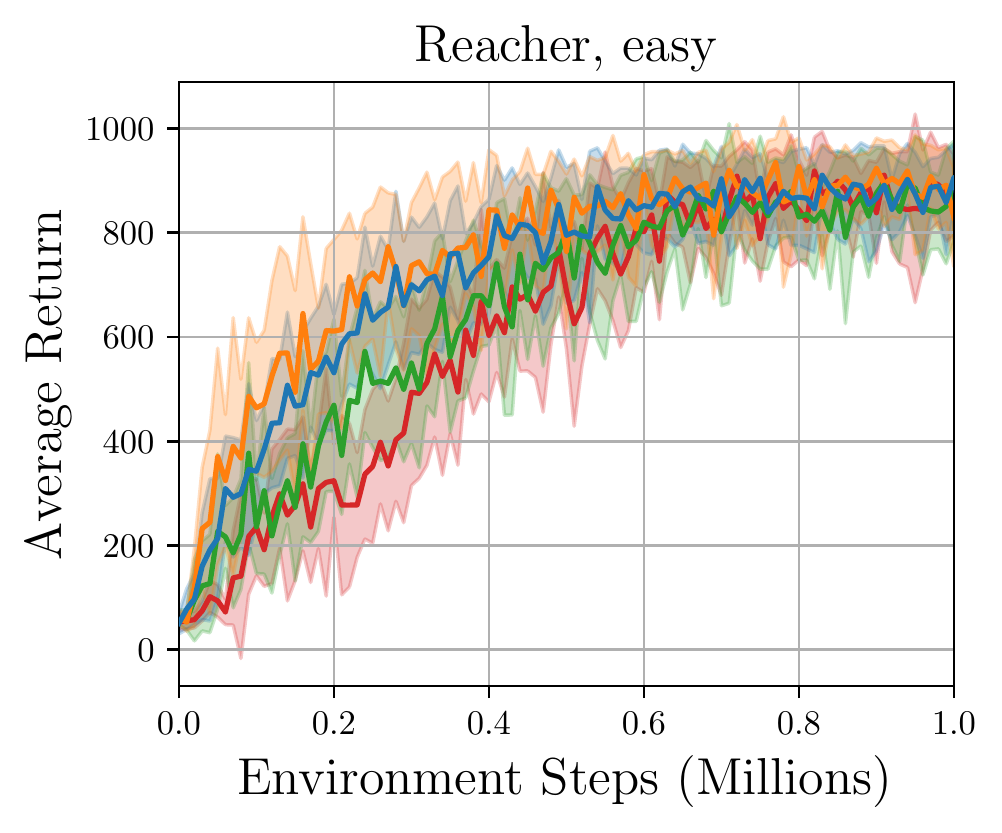} \\
  \includegraphics[scale=0.35]{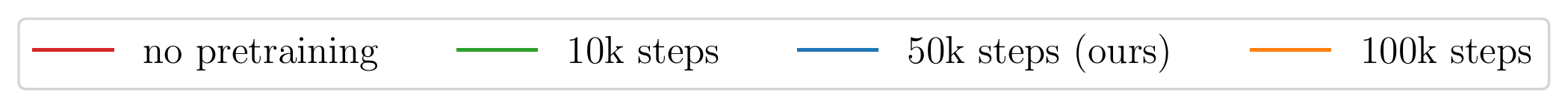}
  \caption{Comparison of the effect of pretraining the model before the agent starts learning on the task.
  These results show that the agent benefits from the supervision signal of the model even before making any progress on the task---little or no pretraining results in slower learning and, in some cases, worse asymptotic performance.
  }
  \label{fig:pretraining_steps_ablation_all}
\end{figure}

\begin{figure}[H]
  \centering
  \includegraphics[scale=0.37, trim={2mm 8mm 2.5mm 2.5mm}, clip]{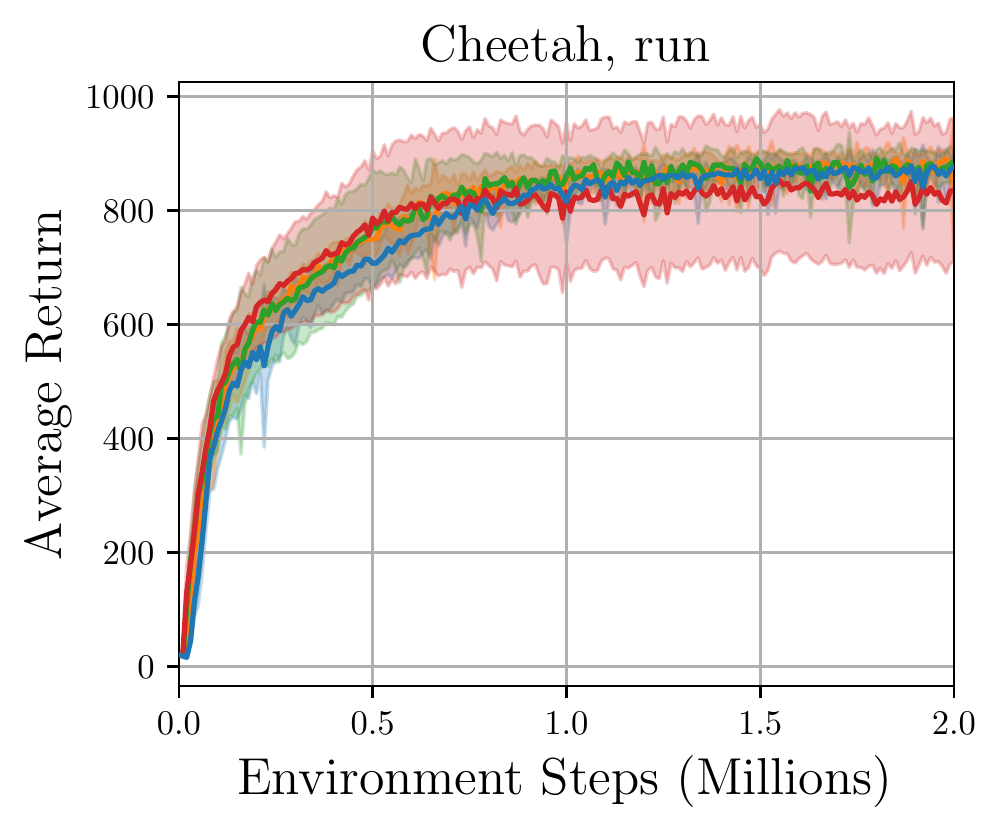} \hspace{1mm}
  \includegraphics[scale=0.37, trim={8.5mm 8mm 2.5mm 2.5mm}, clip]{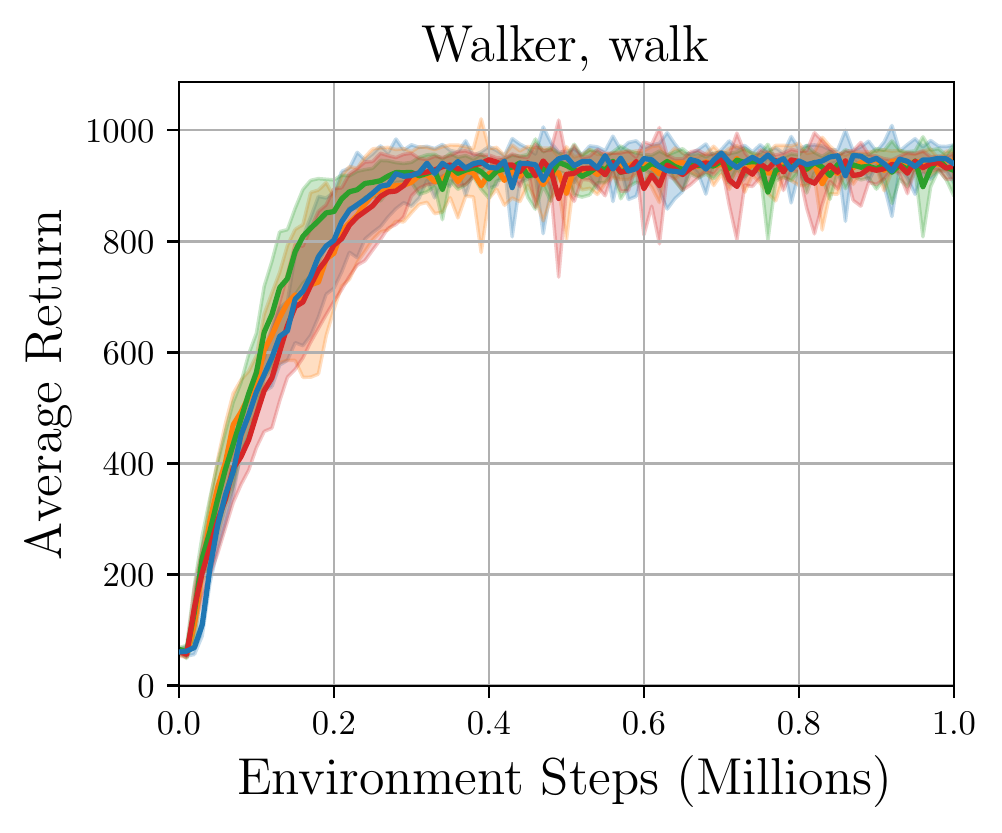} \hspace{1mm}
  \includegraphics[scale=0.37, trim={8.5mm 8mm 2.5mm 2.5mm}, clip]{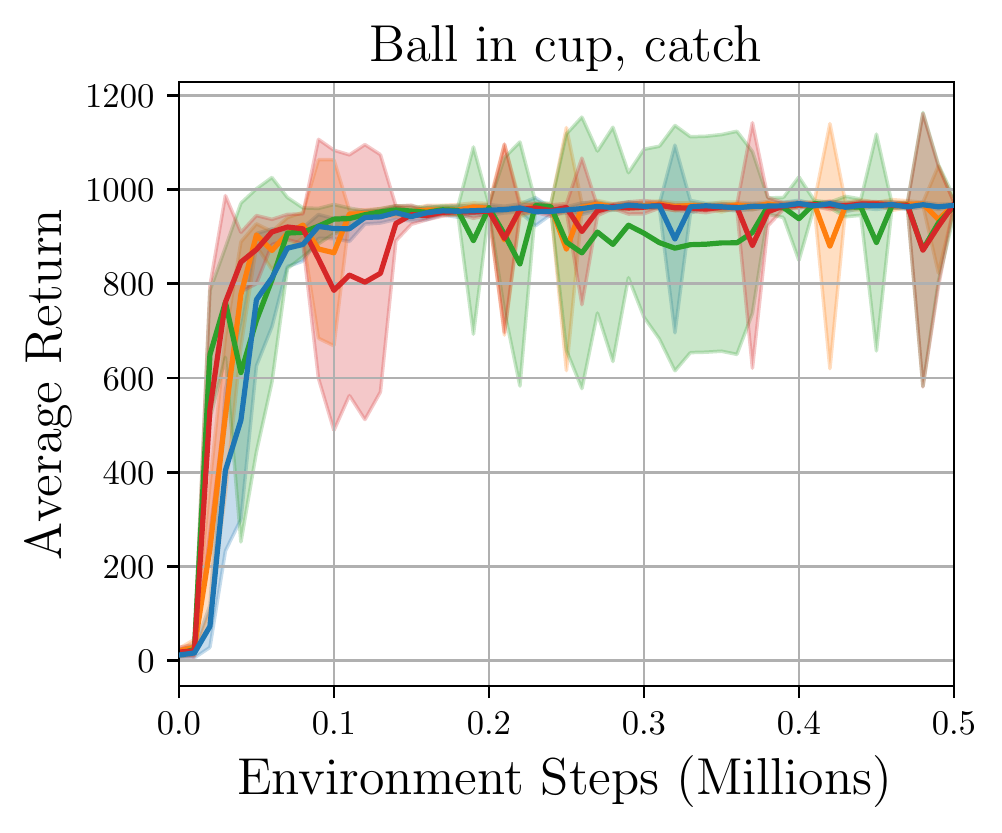} \\[1mm]
  \includegraphics[scale=0.37, trim={2mm 2.5mm 2.5mm 2.5mm}, clip]{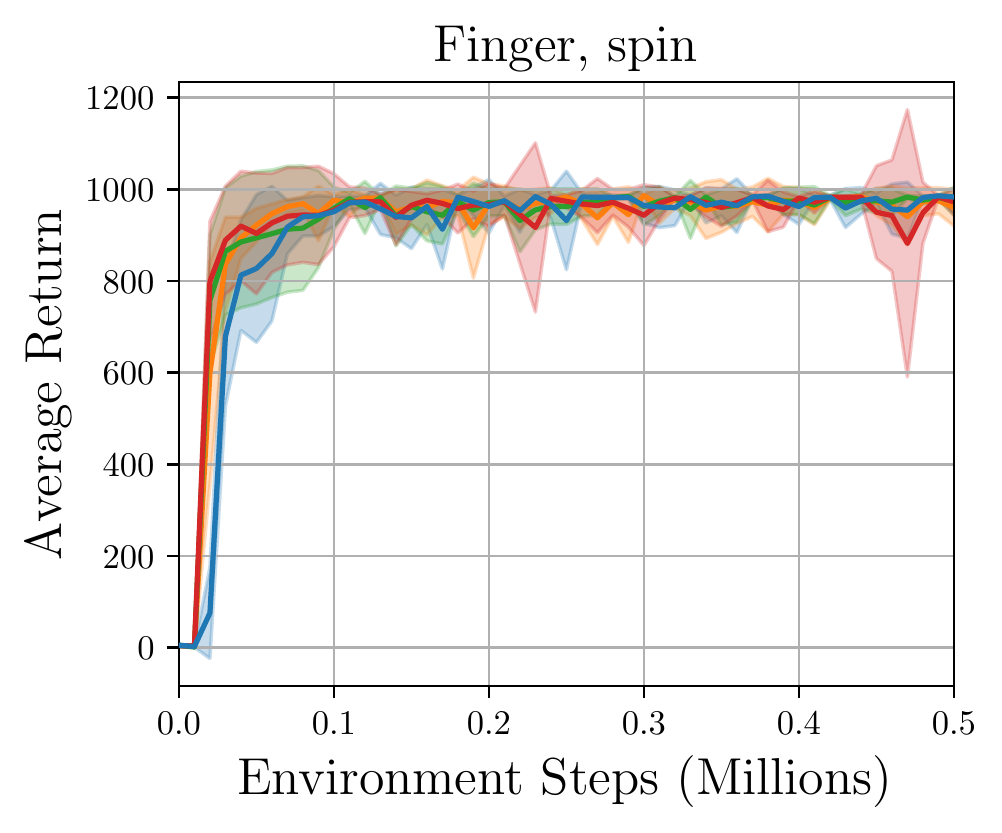} \hspace{1mm}
  \includegraphics[scale=0.37, trim={8.5mm 2.5mm 2.5mm 2.5mm}, clip]{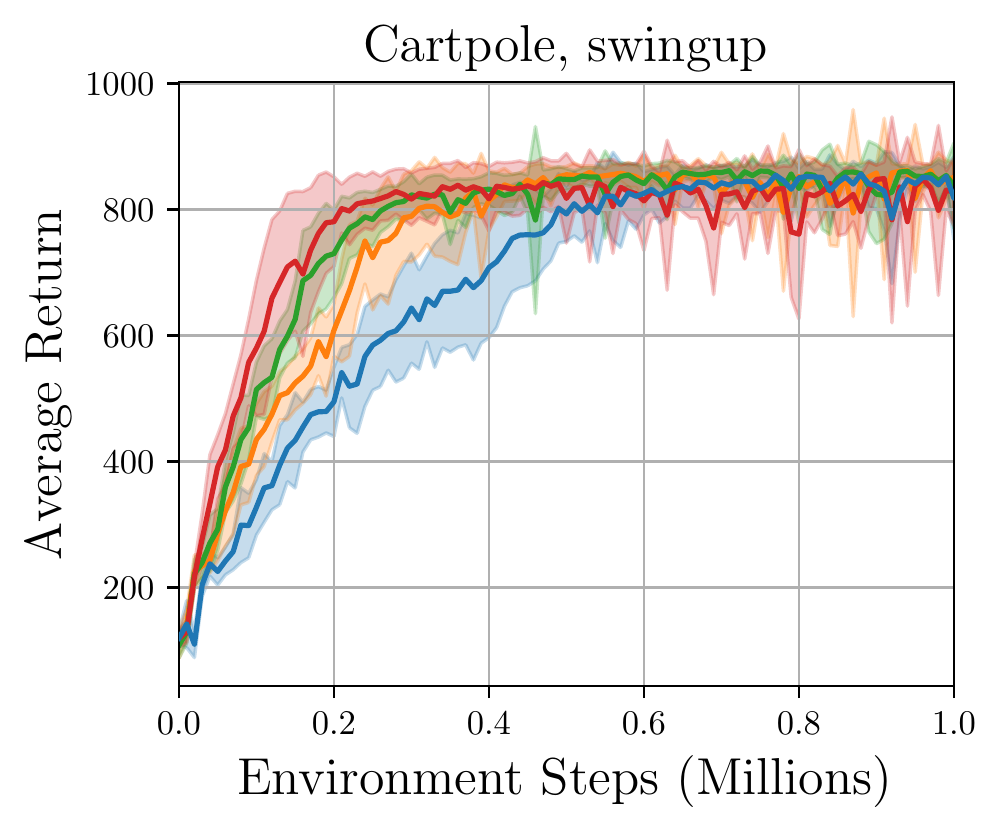} \hspace{1mm}
  \includegraphics[scale=0.37, trim={8.5mm 2.5mm 2.5mm 2.5mm}, clip]{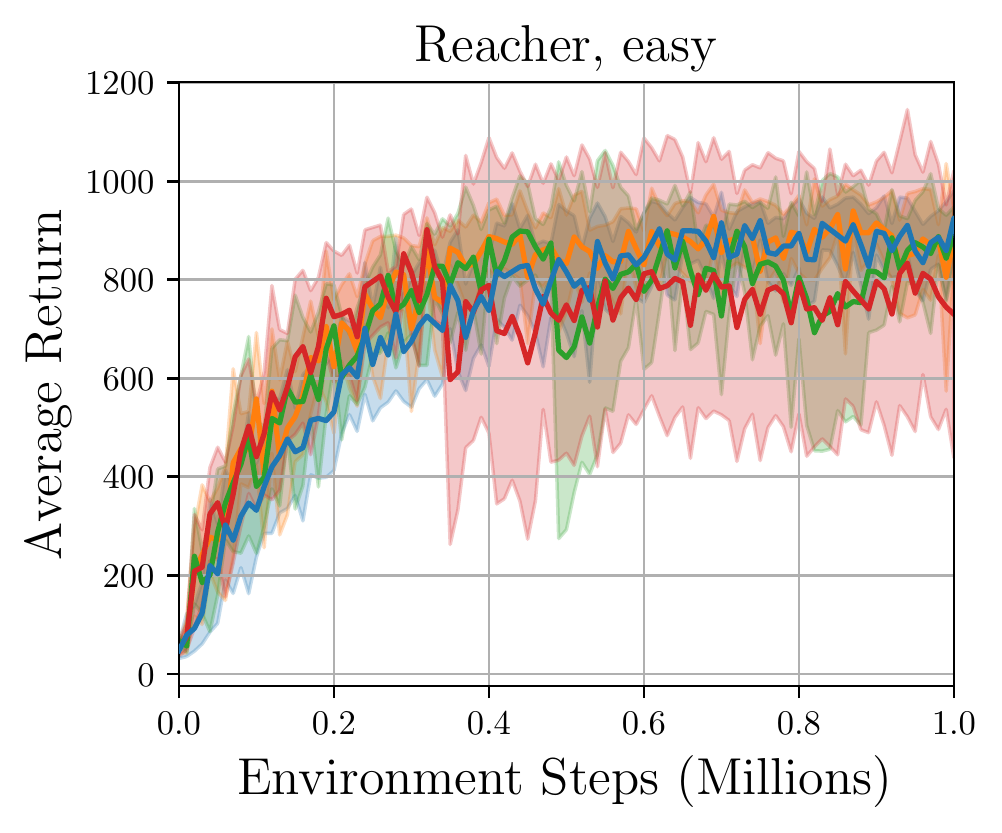} \\
  \includegraphics[scale=0.35]{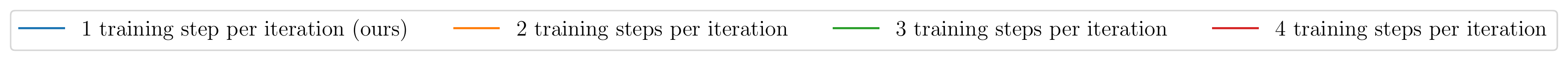}
  \caption{Comparison of the effect of the number of training updates per iteration (i.e. training updates per environment step).
  These results show that more training updates per iteration speeds up learning slightly, but too many updates per iteration causes higher variance across trials and, in some cases, slightly worse asymptotic performance.
  }
  \label{fig:train_steps_per_iteration_ablation_all}
\end{figure}

We show results for the ablation experiments from \autoref{sec:ablations} for additional environments.
\autoref{fig:model_ablation_all} compares different design choices for the latent variable model.
\autoref{fig:actor_input_critic_input_ablation_all} compares alternative choices for the actor and critic inputs as either the observation-action history or the latent sample.
\autoref{fig:pretraining_steps_ablation_all} compares the effect of pretraining the model before the agent starts learning on the task.
\autoref{fig:train_steps_per_iteration_ablation_all} compares the effect of the number of training updates per iteration.
In addition, we investigate the choice of the decoder output variance and using random cropping for data augmentation.

As in the main DeepMind Control Suite results, all the ablation experiments sweep over $\sigma^2 \in \{0.04, 0.1, 0.4\}$ and show results corresponding to the best per-task hyperparameter $\sigma^2$, unless otherwise specified.
This ensures a fairer and more informative comparison across the ablated methods.

\textbf{Output variance of the pixel decoder.}
The output variance $\sigma^2$ of the pixels in the image determines the relative weighting between the reconstruction loss and the KL-divergence.
The best weighting is determined by the complexity of the dataset~\citep{alemi2018fixing}, which in our case is dependent on the task.

\begin{figure}[H]
  \centering
  \includegraphics[scale=0.37, trim={2mm 8mm 2.5mm 2.5mm}, clip]{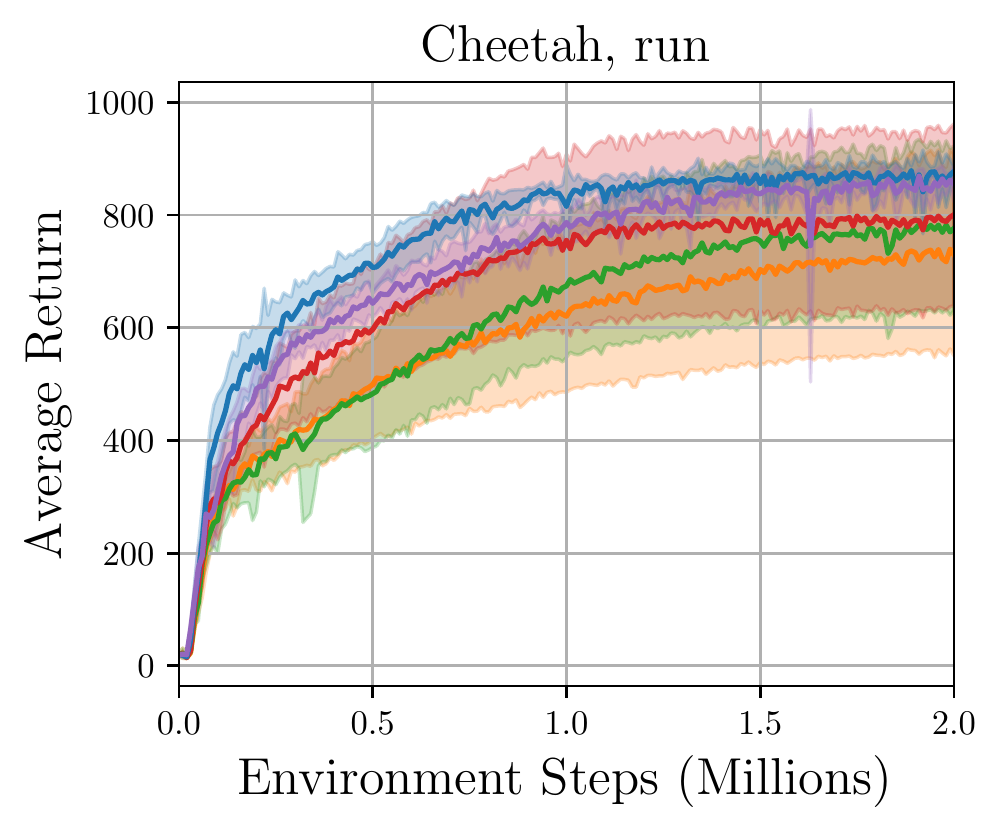} \hspace{1mm}
  \includegraphics[scale=0.37, trim={8.5mm 8mm 2.5mm 2.5mm}, clip]{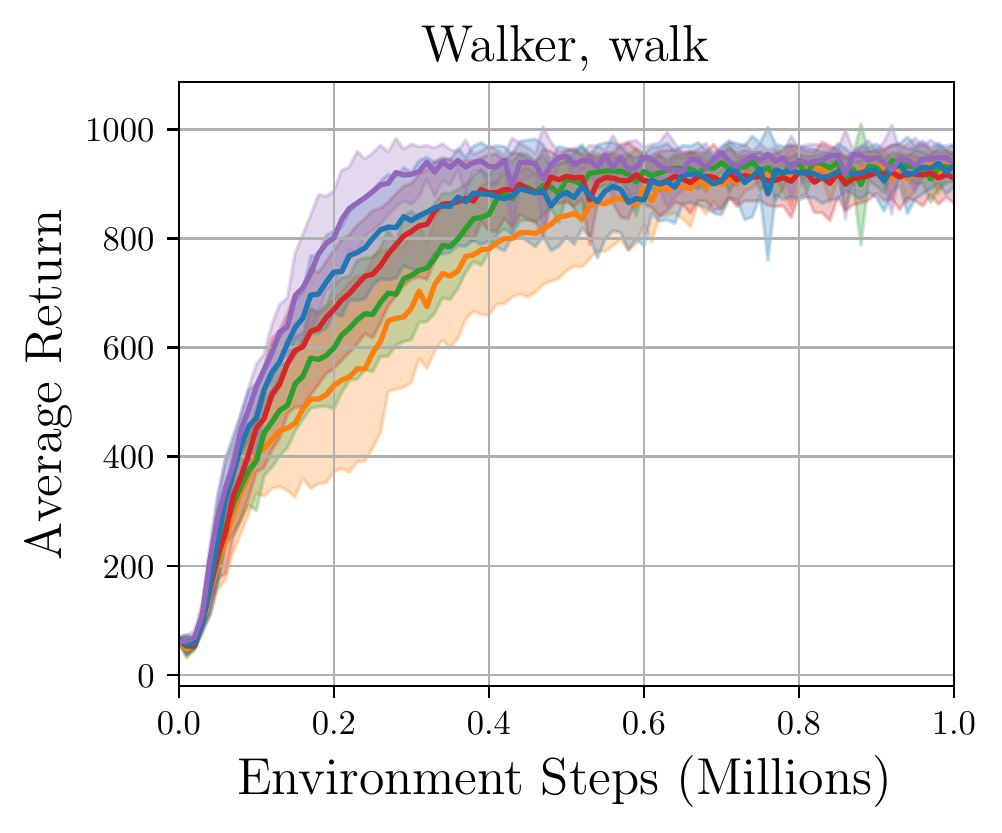} \hspace{1mm}
  \includegraphics[scale=0.37, trim={8.5mm 8mm 2.5mm 2.5mm}, clip]{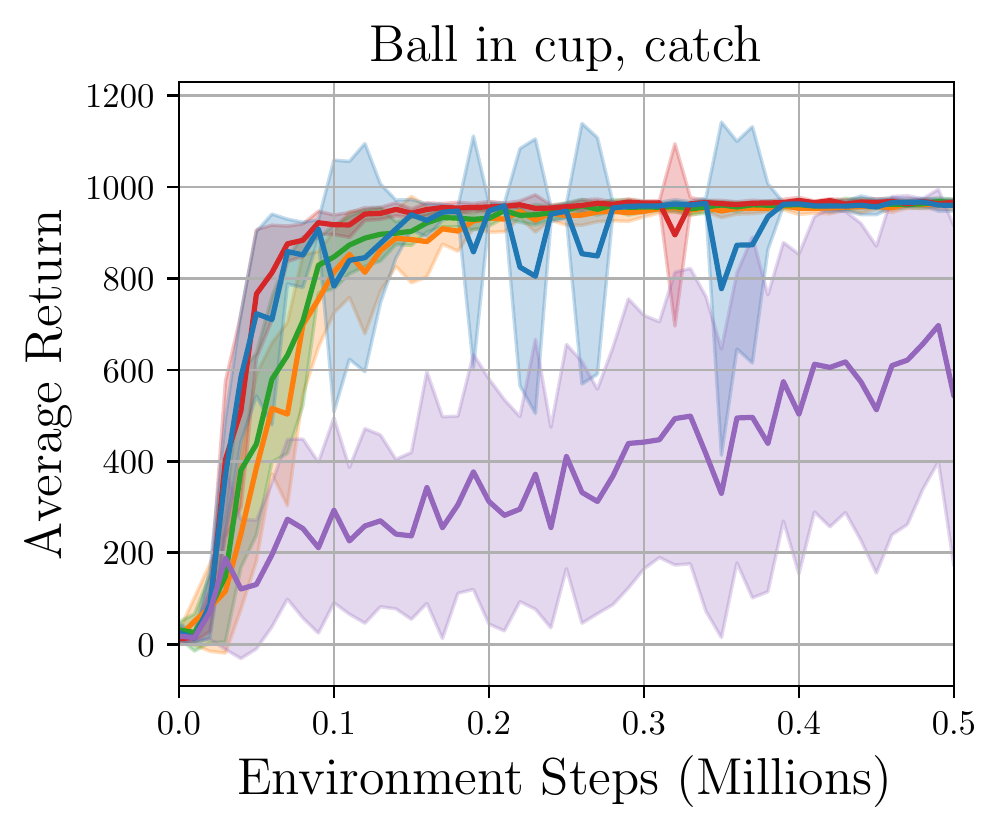} \\[1mm]
  \includegraphics[scale=0.37, trim={2mm 2.5mm 2.5mm 2.5mm}, clip]{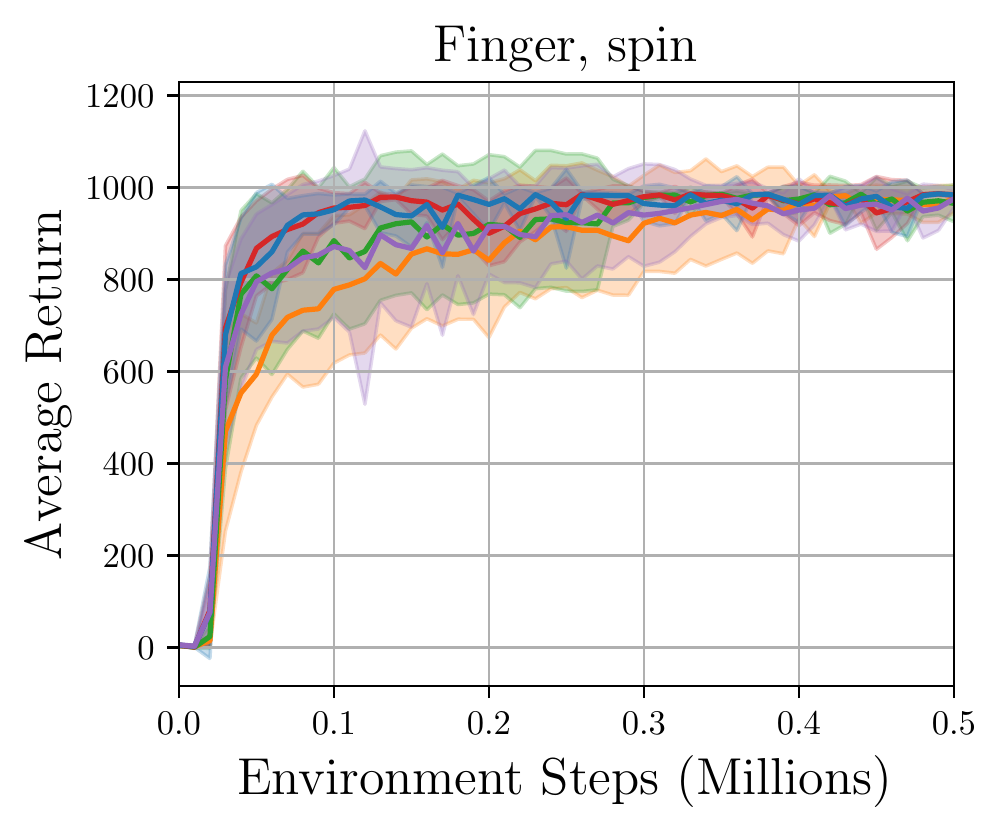} \hspace{1mm}
  \includegraphics[scale=0.37, trim={8.5mm 2.5mm 2.5mm 2.5mm}, clip]{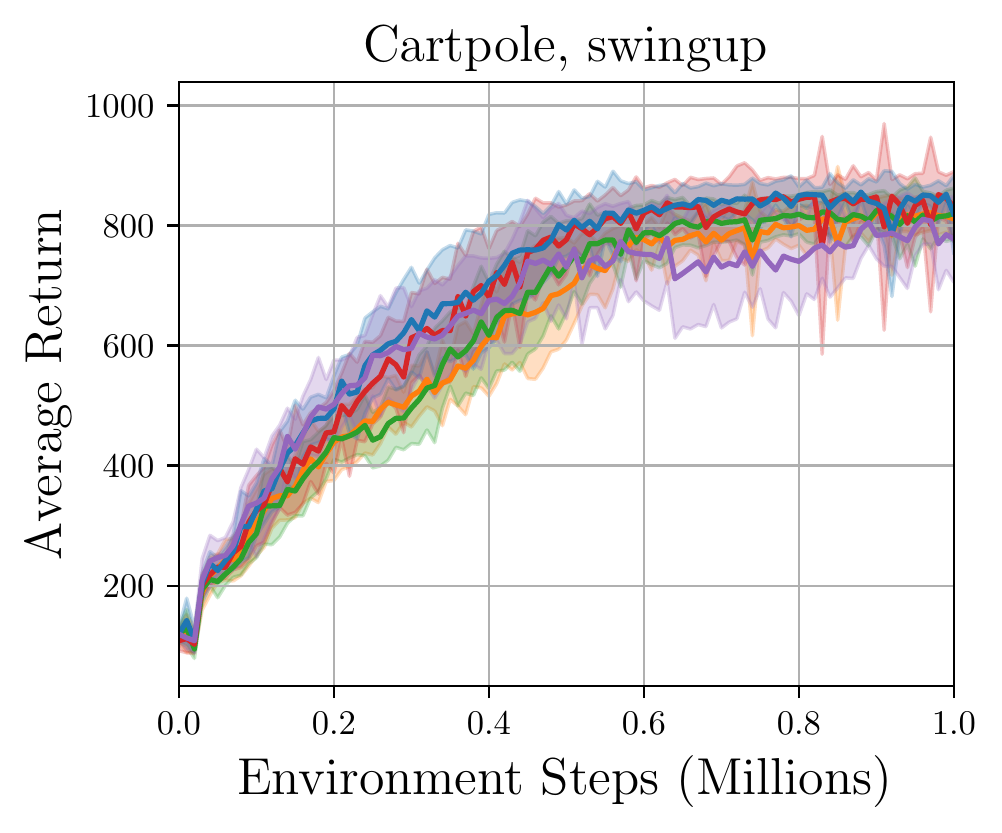} \hspace{1mm}
  \includegraphics[scale=0.37, trim={8.5mm 2.5mm 2.5mm 2.5mm}, clip]{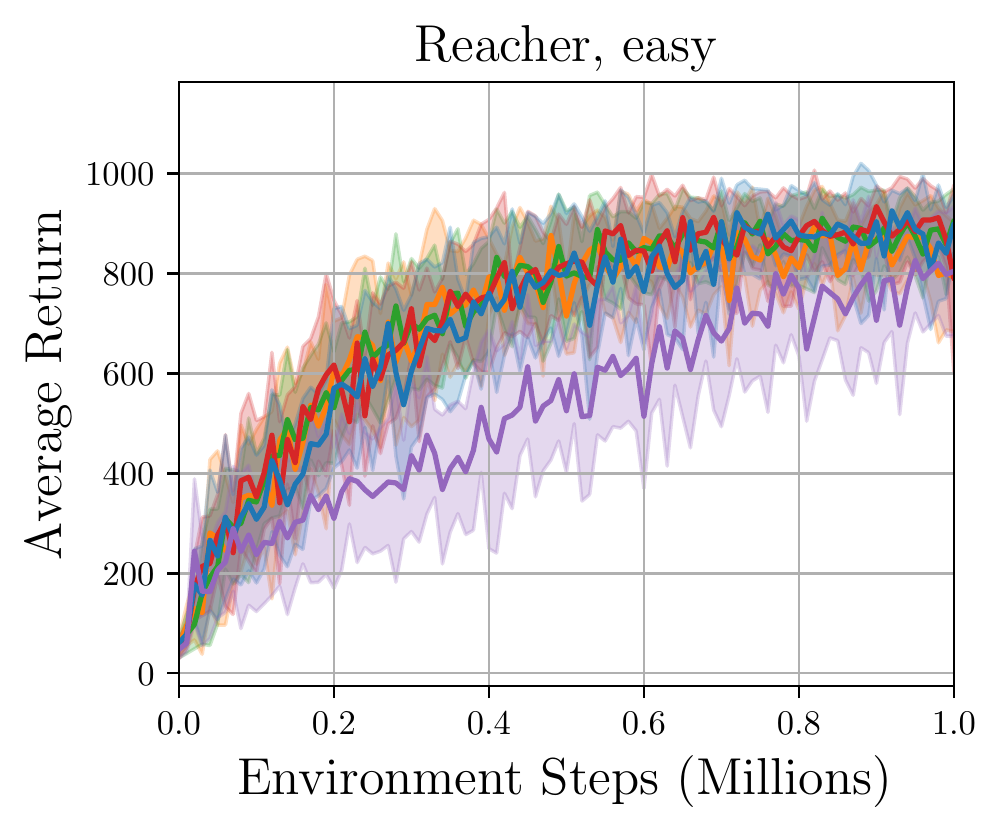} \\
  \includegraphics[scale=0.35]{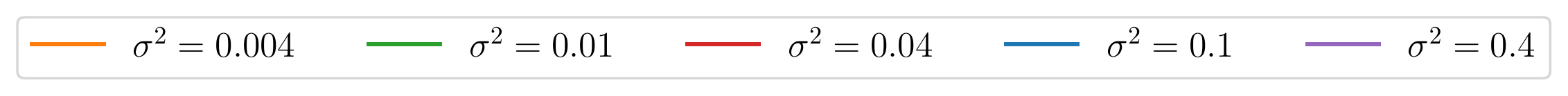}
  \caption{Comparison of different choices for the output variance of the pixel decoder.
  Good performance is achieved with $\sigma^2 = 0.1$, except for the tasks walker walk ($\sigma^2 = 0.4$) and ball-in-cup catch ($\sigma^2 = 0.04$).
  }
  \label{fig:decoder_var_ablation_all}
\end{figure}

As shown in \autoref{fig:decoder_var_ablation_all}, our model is sensitive to this hyperparameter, just as with any other VAE model.
Overall, a value of $\sigma^2 = 0.1$ gives good results, except for the walker walk and ball-in-cup catch tasks. The walker walk task benefits from a larger $\sigma^2 = 0.4$ likely because 
the images are harder to predict, a larger pixel area of the image changes over time, and the walker configuration varies considerably within an episode and throughout learning (e.g. when the walker falls over and bounces off the ground). On the other hand, the ball-in-cup catch task benefits from a smaller $\sigma^2 = 0.04$ likely because fewer pixels change over time.

\textbf{Random cropping.}
We next investigate the effect of using random cropping for data augmentation.
This augmentation consists of padding (replication) the $64 \times 64$ images by 4 pixels on each side, resulting in $72 \times 72$ images, and randomly sampling $64 \times 64$ crops from them. For training, we use these randomly translated images both as inputs to the model and the policy, whereas we use the original images as targets for the reconstruction loss of the model. For evaluation, we always use the original images as inputs to the policy.

\begin{figure}[H]
  \centering
  \includegraphics[scale=0.37, trim={2mm 8mm 2.5mm 2.5mm}, clip]{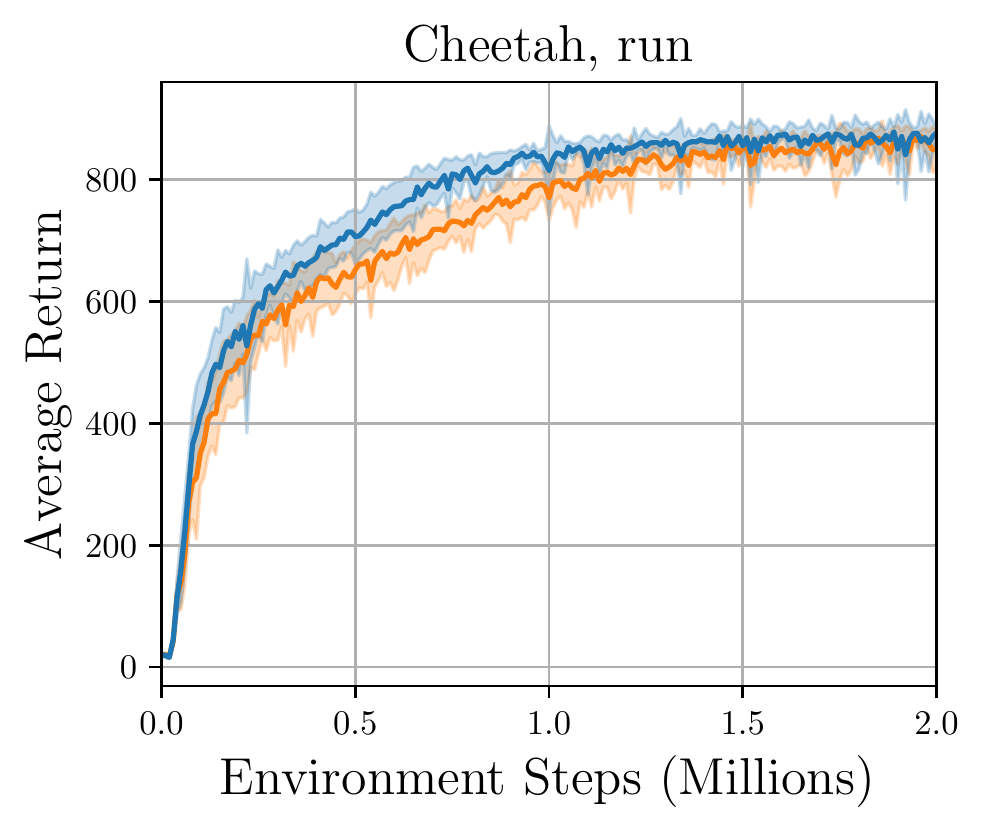} \hspace{1mm}
  \includegraphics[scale=0.37, trim={8.5mm 8mm 2.5mm 2.5mm}, clip]{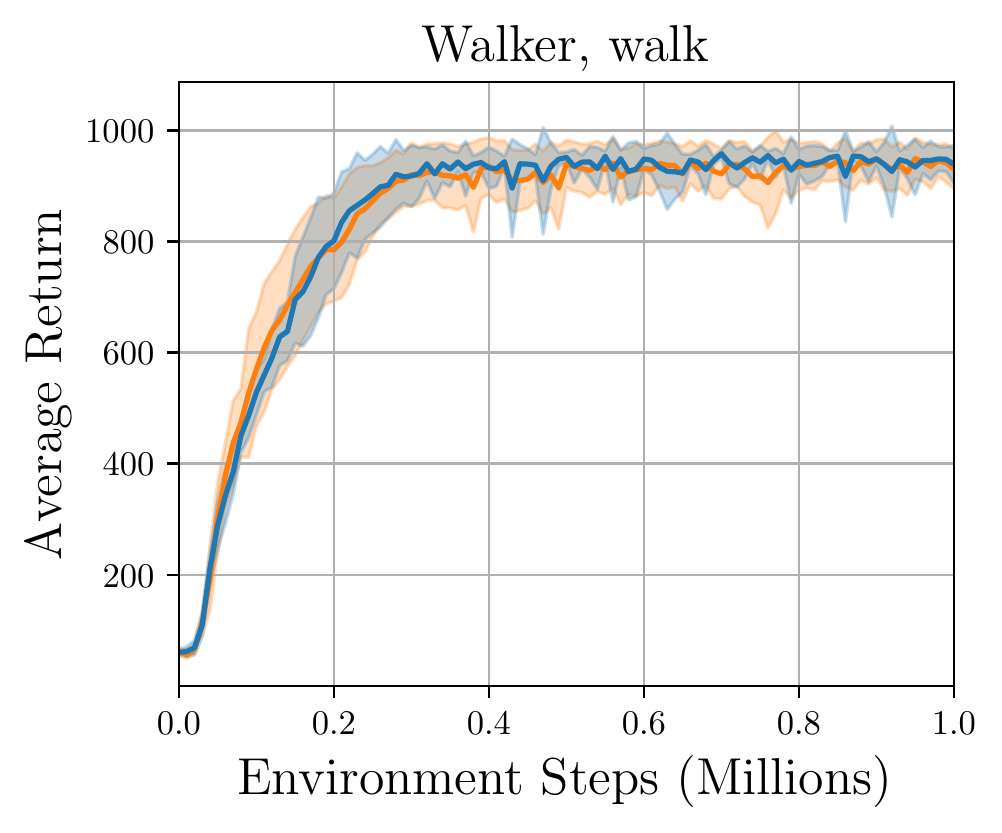} \hspace{1mm}
  \includegraphics[scale=0.37, trim={8.5mm 8mm 2.5mm 2.5mm}, clip]{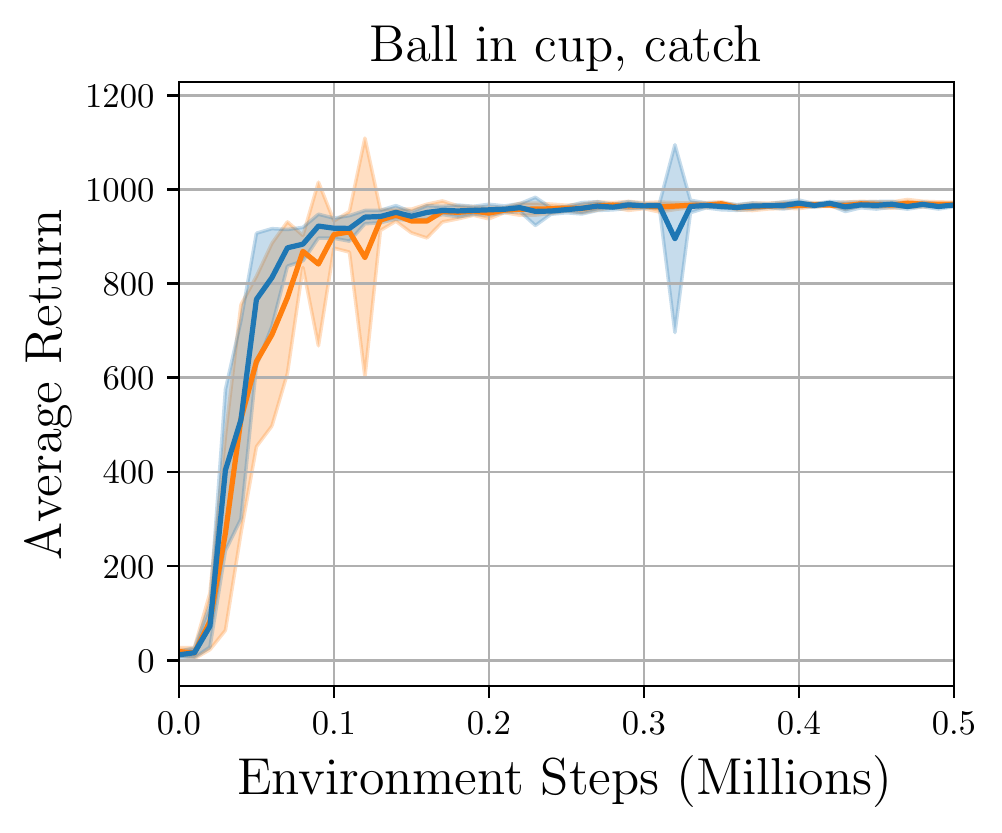} \\[1mm]
  \includegraphics[scale=0.37, trim={2mm 2.5mm 2.5mm 2.5mm}, clip]{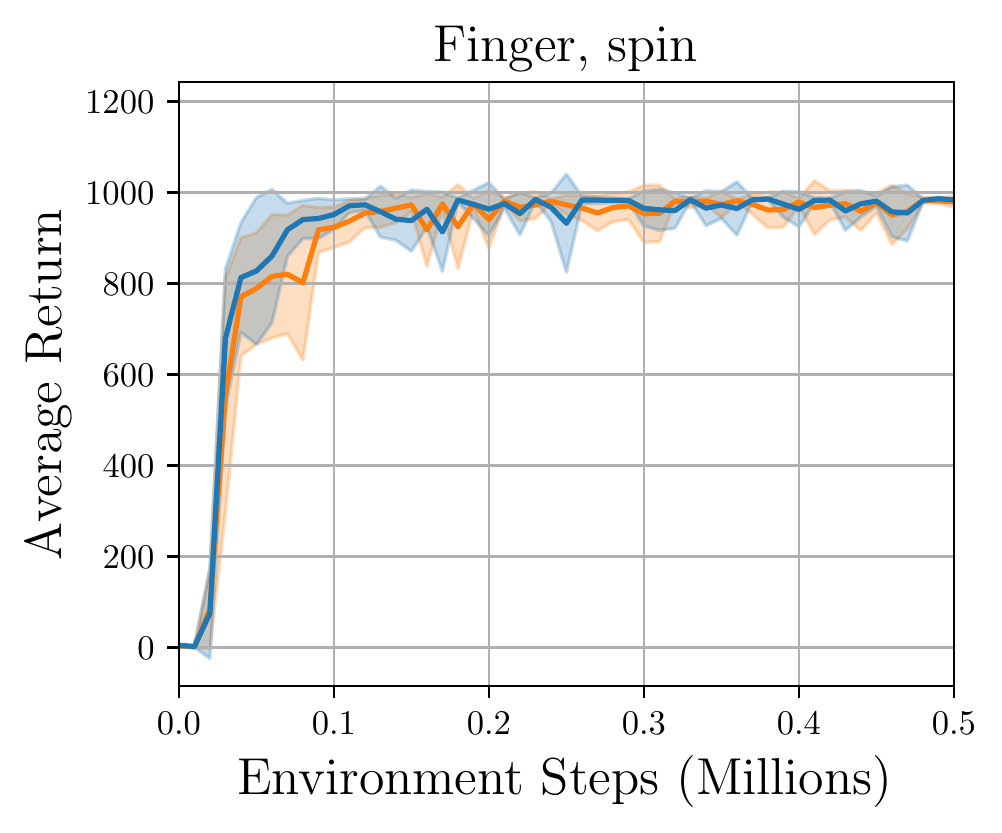} \hspace{1mm}
  \includegraphics[scale=0.37, trim={8.5mm 2.5mm 2.5mm 2.5mm}, clip]{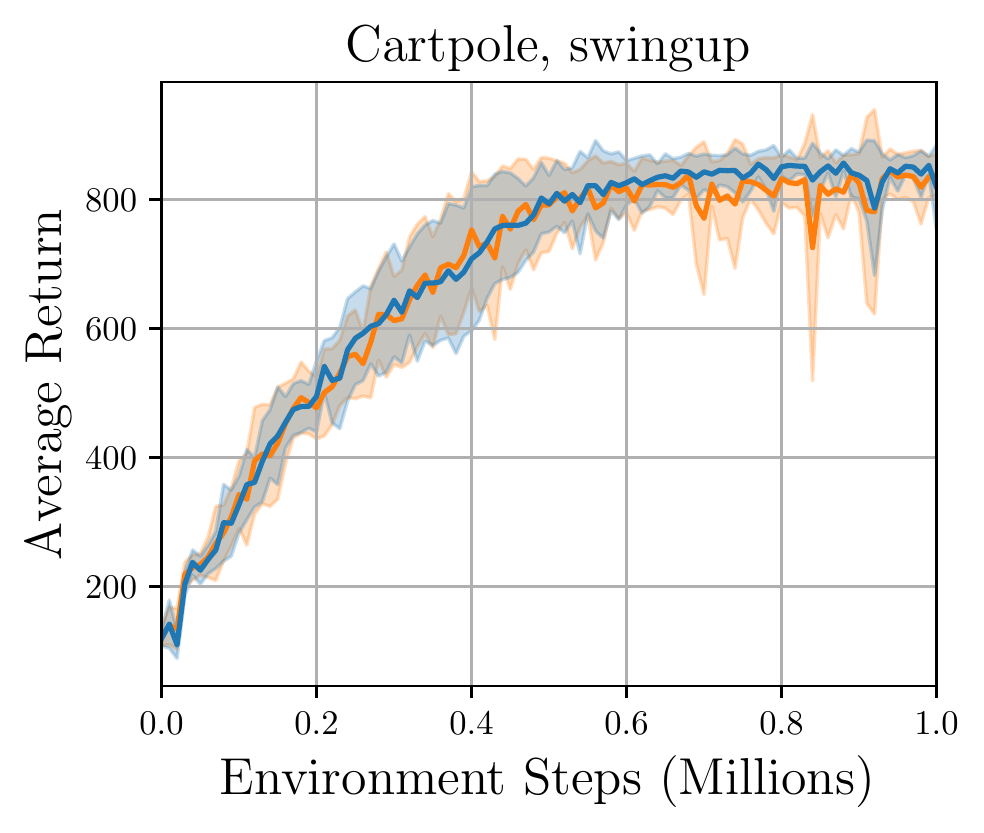} \hspace{1mm}
  \includegraphics[scale=0.37, trim={8.5mm 2.5mm 2.5mm 2.5mm}, clip]{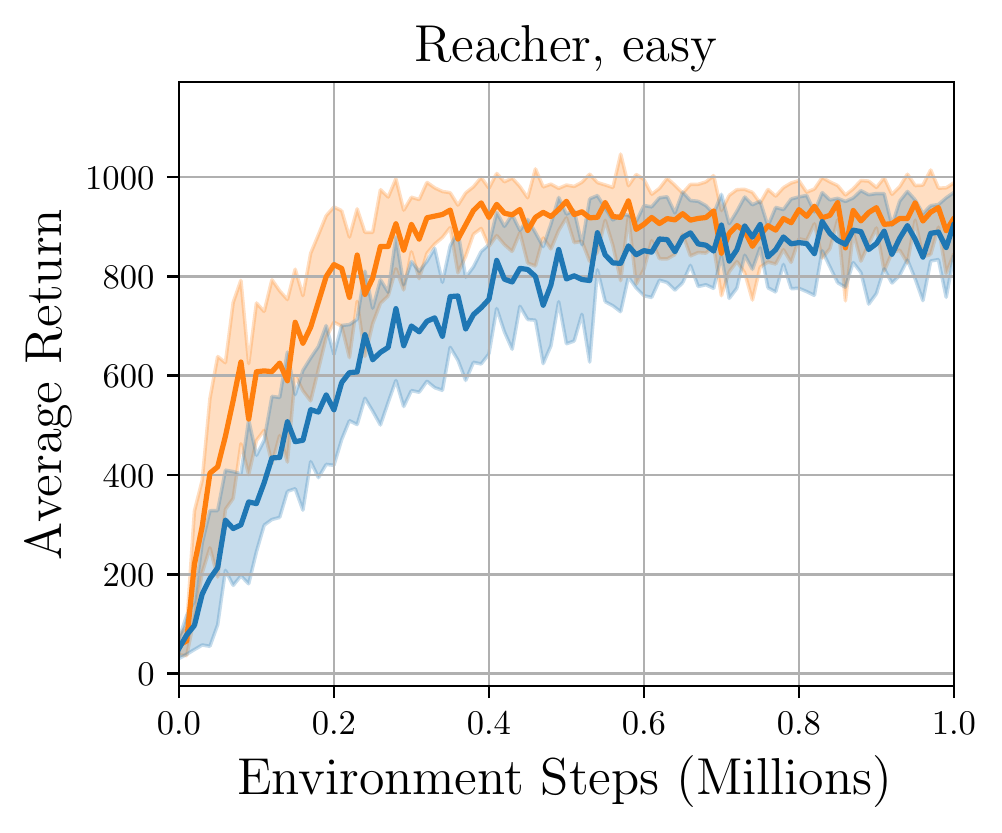} \\
  \includegraphics[scale=0.35]{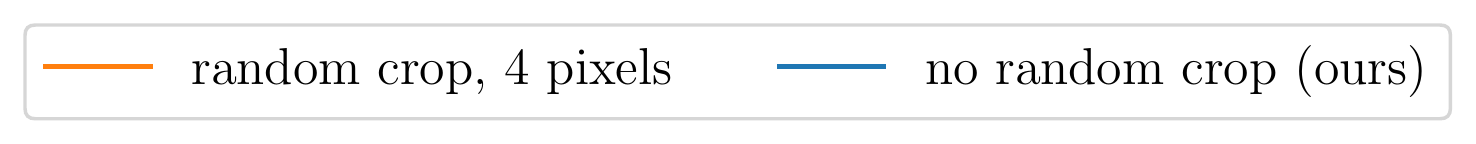}
  \caption{Comparison of using random cropping for data augmentation of the input images.
  The random cropping doesn't improve the learning performance except for the reacher easy task, in which this data augmentation results in faster learning and higher asymptotic performance.
  }
  \label{fig:random_crop_ablation_all}
\end{figure}

As shown in \autoref{fig:random_crop_ablation_all}, this random cropping doesn't improve the learning performance except for the reacher easy task, in which this data augmentation results in faster learning and higher asymptotic performance.

\vspace{-1mm}
\section{Predictions from the Latent Variable Model}
\vspace{-1mm}
\label{app:samples}

\begin{figure}
  \centering
  \begin{tabular}{c c l}
    \multirow{11}{12pt}{\rotatebox{90}{\makecell{\fontsize{9pt}{9pt}\selectfont \bfseries Cheetah, run}}} &
    \adjustbox{valign=c}{\rotatebox{90}{\scalebox{0.65}{\makecell{Ground \\ Truth}}}} &
    \adjustbox{valign=c,margin=0 1pt}{\includegraphics[width=0.76\linewidth]{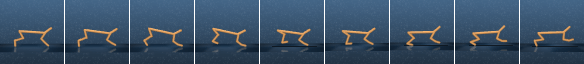}} \\
    &
    \adjustbox{valign=c}{\rotatebox{90}{\scalebox{0.65}{\makecell{Posterior \\ Sample}}}} &
    \adjustbox{valign=c,margin=0 1pt}{\includegraphics[width=0.76\linewidth]{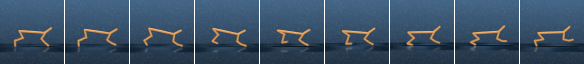}} \\
    &
    \adjustbox{valign=c}{\rotatebox{90}{\scalebox{0.65}{\makecell{Conditional \\ Prior Sample}}}} &
    \adjustbox{valign=c,margin=0 1pt}{\includegraphics[width=0.76\linewidth]{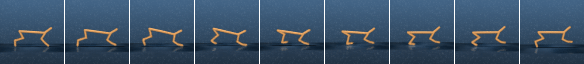}} \\
    &
    \adjustbox{valign=c}{\rotatebox{90}{\scalebox{0.65}{\makecell{Prior \\ Sample}}}} &
    \adjustbox{valign=c,margin=0 1pt}{\includegraphics[width=0.76\linewidth]{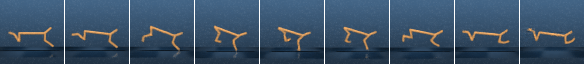}} \\[5.5mm]

    \multirow{11}{12pt}{\rotatebox{90}{\makecell{\fontsize{9pt}{9pt}\selectfont \bfseries Walker, walk}}} &
    \adjustbox{valign=c}{\rotatebox{90}{\scalebox{0.65}{\makecell{Ground \\ Truth}}}} &
    \adjustbox{valign=c,margin=0 1pt}{\includegraphics[width=0.76\linewidth]{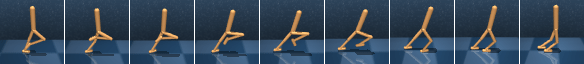}} \\
    &
    \adjustbox{valign=c}{\rotatebox{90}{\scalebox{0.65}{\makecell{Posterior \\ Sample}}}} &
    \adjustbox{valign=c,margin=0 1pt}{\includegraphics[width=0.76\linewidth]{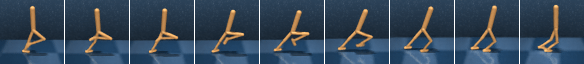}} \\
    &
    \adjustbox{valign=c}{\rotatebox{90}{\scalebox{0.65}{\makecell{Conditional \\ Prior Sample}}}} &
    \adjustbox{valign=c,margin=0 1pt}{\includegraphics[width=0.76\linewidth]{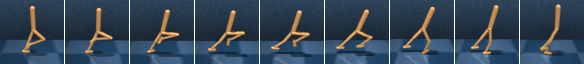}} \\
    &
    \adjustbox{valign=c}{\rotatebox{90}{\scalebox{0.65}{\makecell{Prior \\ Sample}}}} &
    \adjustbox{valign=c,margin=0 1pt}{\includegraphics[width=0.76\linewidth]{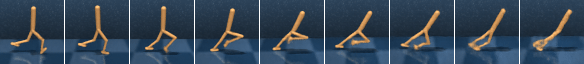}} \\[5.5mm]

    \multirow{11}{12pt}{\rotatebox{90}{\makecell{\fontsize{9pt}{9pt}\selectfont \bfseries Ball in cup, catch}}} &
    \adjustbox{valign=c}{\rotatebox{90}{\scalebox{0.65}{\makecell{Ground \\ Truth}}}} &
    \adjustbox{valign=c,margin=0 1pt}{\includegraphics[width=0.76\linewidth]{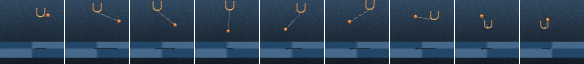}} \\
    &
    \adjustbox{valign=c}{\rotatebox{90}{\scalebox{0.65}{\makecell{Posterior \\ Sample}}}} &
    \adjustbox{valign=c,margin=0 1pt}{\includegraphics[width=0.76\linewidth]{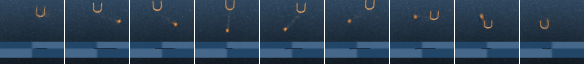}} \\
    &
    \adjustbox{valign=c}{\rotatebox{90}{\scalebox{0.65}{\makecell{Conditional \\ Prior Sample}}}} &
    \adjustbox{valign=c,margin=0 1pt}{\includegraphics[width=0.76\linewidth]{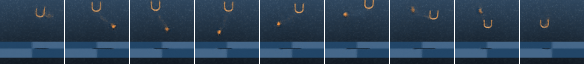}} \\
    &
    \adjustbox{valign=c}{\rotatebox{90}{\scalebox{0.65}{\makecell{Prior \\ Sample}}}} &
    \adjustbox{valign=c,margin=0 1pt}{\includegraphics[width=0.76\linewidth]{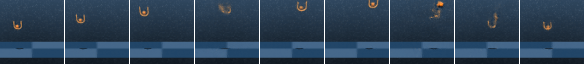}} \\[5.5mm]

    \multirow{11}{12pt}{\rotatebox{90}{\makecell{\fontsize{9pt}{9pt}\selectfont \bfseries Finger, spin}}} &
    \adjustbox{valign=c}{\rotatebox{90}{\scalebox{0.65}{\makecell{Ground \\ Truth}}}} &
    \adjustbox{valign=c,margin=0 1pt}{\includegraphics[width=0.76\linewidth]{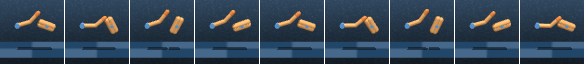}} \\
    &
    \adjustbox{valign=c}{\rotatebox{90}{\scalebox{0.65}{\makecell{Posterior \\ Sample}}}} &
    \adjustbox{valign=c,margin=0 1pt}{\includegraphics[width=0.76\linewidth]{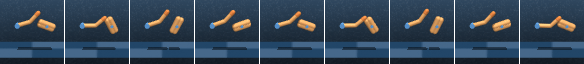}} \\
    &
    \adjustbox{valign=c}{\rotatebox{90}{\scalebox{0.65}{\makecell{Conditional \\ Prior Sample}}}} &
    \adjustbox{valign=c,margin=0 1pt}{\includegraphics[width=0.76\linewidth]{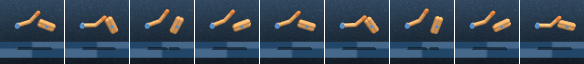}} \\
    &
    \adjustbox{valign=c}{\rotatebox{90}{\scalebox{0.65}{\makecell{Prior \\ Sample}}}} &
    \adjustbox{valign=c,margin=0 1pt}{\includegraphics[width=0.76\linewidth]{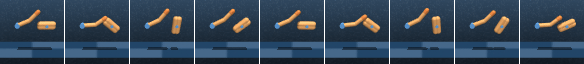}}
  \end{tabular}
  \caption{
  Example image sequences for the four DeepMind Control Suite tasks (first rows), along with corresponding posterior samples (reconstruction) from our model (second rows), and generated predictions from the generative model (last two rows). The second to last row is conditioned on the first frame (i.e., the posterior model is used for the first time step while the prior model is used for all subsequent steps), whereas the last row is not conditioned on any ground truth images. Note that all of these sampled sequences are conditioned on the same action sequence, and that our model produces highly realistic samples, even when predicting via the generative model.
  }
  \label{fig:dm_control_samples_observations}
\end{figure}

\begin{figure}
  \centering
  \begin{tabular}{c c l}
    \multirow{11}{12pt}{\rotatebox{90}{\makecell{\fontsize{9pt}{9pt}\selectfont \bfseries HalfCheetah-v2}}} &
    \adjustbox{valign=c}{\rotatebox{90}{\scalebox{0.65}{\makecell{Ground \\ Truth}}}} &
    \adjustbox{valign=c,margin=0 1pt}{\includegraphics[width=0.76\linewidth]{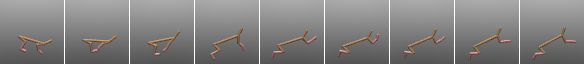}} \\
    &
    \adjustbox{valign=c}{\rotatebox{90}{\scalebox{0.65}{\makecell{Posterior \\ Sample}}}} &
    \adjustbox{valign=c,margin=0 1pt}{\includegraphics[width=0.76\linewidth]{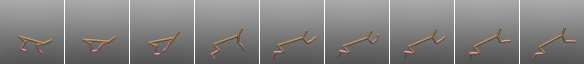}} \\
    &
    \adjustbox{valign=c}{\rotatebox{90}{\scalebox{0.65}{\makecell{Conditional \\ Prior Sample}}}} &
    \adjustbox{valign=c,margin=0 1pt}{\includegraphics[width=0.76\linewidth]{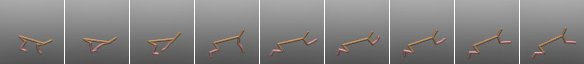}} \\
    &
    \adjustbox{valign=c}{\rotatebox{90}{\scalebox{0.65}{\makecell{Prior \\ Sample}}}} &
    \adjustbox{valign=c,margin=0 1pt}{\includegraphics[width=0.76\linewidth]{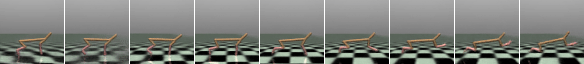}} \\[5.5mm]

    \multirow{11}{12pt}{\rotatebox{90}{\makecell{\fontsize{9pt}{9pt}\selectfont \bfseries Walker2d-v2}}} &
    \adjustbox{valign=c}{\rotatebox{90}{\scalebox{0.65}{\makecell{Ground \\ Truth}}}} &
    \adjustbox{valign=c,margin=0 1pt}{\includegraphics[width=0.76\linewidth]{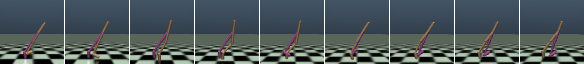}} \\
    &
    \adjustbox{valign=c}{\rotatebox{90}{\scalebox{0.65}{\makecell{Posterior \\ Sample}}}} &
    \adjustbox{valign=c,margin=0 1pt}{\includegraphics[width=0.76\linewidth]{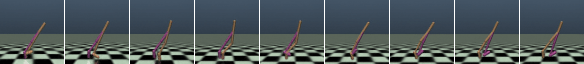}} \\
    &
    \adjustbox{valign=c}{\rotatebox{90}{\scalebox{0.65}{\makecell{Conditional \\ Prior Sample}}}} &
    \adjustbox{valign=c,margin=0 1pt}{\includegraphics[width=0.76\linewidth]{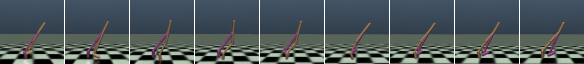}} \\
    &
    \adjustbox{valign=c}{\rotatebox{90}{\scalebox{0.65}{\makecell{Prior \\ Sample}}}} &
    \adjustbox{valign=c,margin=0 1pt}{\includegraphics[width=0.76\linewidth]{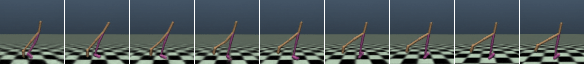}} \\[5.5mm]

    \multirow{11}{12pt}{\rotatebox{90}{\makecell{\fontsize{9pt}{9pt}\selectfont \bfseries Hopper-v2}}} &
    \adjustbox{valign=c}{\rotatebox{90}{\scalebox{0.65}{\makecell{Ground \\ Truth}}}} &
    \adjustbox{valign=c,margin=0 1pt}{\includegraphics[width=0.76\linewidth]{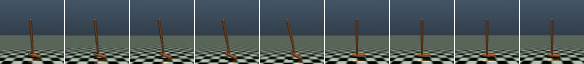}} \\
    &
    \adjustbox{valign=c}{\rotatebox{90}{\scalebox{0.65}{\makecell{Posterior \\ Sample}}}} &
    \adjustbox{valign=c,margin=0 1pt}{\includegraphics[width=0.76\linewidth]{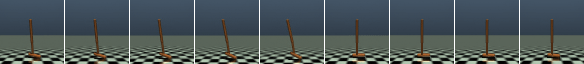}} \\
    &
    \adjustbox{valign=c}{\rotatebox{90}{\scalebox{0.65}{\makecell{Conditional \\ Prior Sample}}}} &
    \adjustbox{valign=c,margin=0 1pt}{\includegraphics[width=0.76\linewidth]{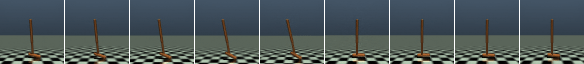}} \\
    &
    \adjustbox{valign=c}{\rotatebox{90}{\scalebox{0.65}{\makecell{Prior \\ Sample}}}} &
    \adjustbox{valign=c,margin=0 1pt}{\includegraphics[width=0.76\linewidth]{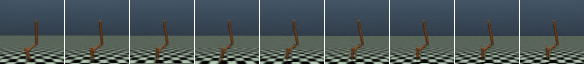}} \\[5.5mm]

    \multirow{11}{12pt}{\rotatebox{90}{\makecell{\fontsize{9pt}{9pt}\selectfont \bfseries Ant-v2}}} &
    \adjustbox{valign=c}{\rotatebox{90}{\scalebox{0.65}{\makecell{Ground \\ Truth}}}} &
    \adjustbox{valign=c,margin=0 1pt}{\includegraphics[width=0.76\linewidth]{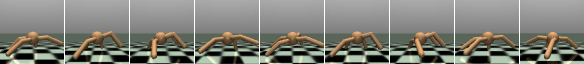}} \\
    &
    \adjustbox{valign=c}{\rotatebox{90}{\scalebox{0.65}{\makecell{Posterior \\ Sample}}}} &
    \adjustbox{valign=c,margin=0 1pt}{\includegraphics[width=0.76\linewidth]{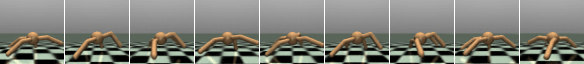}} \\
    &
    \adjustbox{valign=c}{\rotatebox{90}{\scalebox{0.65}{\makecell{Conditional \\ Prior Sample}}}} &
    \adjustbox{valign=c,margin=0 1pt}{\includegraphics[width=0.76\linewidth]{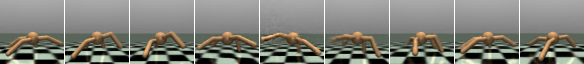}} \\
    &
    \adjustbox{valign=c}{\rotatebox{90}{\scalebox{0.65}{\makecell{Prior \\ Sample}}}} &
    \adjustbox{valign=c,margin=0 1pt}{\includegraphics[width=0.76\linewidth]{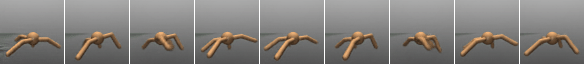}}
  \end{tabular}
  \caption{
  Example image sequences for the four OpenAI Gym tasks (first rows), along with corresponding posterior samples (reconstruction) from our model (second rows), and generated predictions from the generative model (last two rows). The second to last row is conditioned on the first frame (i.e., the posterior model is used for the first time step while the prior model is used for all subsequent steps), whereas the last row is not conditioned on any ground truth images. Note that all of these sampled sequences are conditioned on the same action sequence, and that our model produces highly realistic samples, even when predicting via the generative model.
  }
  \label{fig:gym_samples_observations}
\end{figure}

We show example image samples from our learned sequential latent variable model in \autoref{fig:dm_control_samples_observations} and \autoref{fig:gym_samples_observations}.
Samples from the posterior show the images $\x{t}$ as constructed by the decoder $p_\mparams(\x{t}|\z{t})$, using a sequence of latents $\z{t}$ that are encoded and sampled from the posteriors, $q_\mparams(\z{1} | \x{1})$ and $q_\mparams(\z{t+1} | \x{t+1}, \z{t}, \a{t})$.
Samples from the prior, on the other hand, use a sequence of latents where $\z{1}$ is sampled from $p_\mparams(\z{1})$ and all remaining latents $\z{t}$ are from the propagation of the previous latent state through the latent dynamics $p_\mparams(\z{t+1} | \z{t}, \a{t})$. Note that these prior samples do not use any image frames as inputs, and thus they do not correspond to any ground truth sequence.
We also show samples from the conditional prior, which is conditioned on the first image from the true sequence: for this, the sampling procedure is the same as the prior, except that $\z{1}$ is encoded and sampled from the posterior $q_\mparams(\z{1} | \x{1})$, rather than being sampled from $p_\mparams(\z{1})$.
We notice that the generated images samples can be sharper and more realistic by using a smaller variance for $p_\mparams(\x{t}|\z{t})$ when training the model, but at the expense of a representation that leads to lower returns. Finally, note that we do not actually use the samples from the prior for training.

\newpage

\newpage

%% file: figures/lvm_pgm.tex
\begin{tikzpicture}[
  node distance=0.5mm, auto,
  empty/.style={shape=circle, minimum size=7mm, inner sep=0mm},
  obs/.style={shape=circle, draw=black, fill=black!20, minimum size=7mm, inner sep=0mm},
  unobs/.style={shape=circle, draw=black, minimum size=7mm, inner sep=0mm},
  gen/.style={->, draw=black, thick},
  inf/.style={->, draw=black, densely dashed, bend right=22.5},
]
{\tiny
  \node[empty] (z2_05)   []                 {};
  \node[unobs]   (z2_1)    [right=of z2_05]   {$\z{1}^2$};
  \node[empty] (z2_15)   [right=of z2_1]    {};
  \node[empty] (z2_2)    [right=of z2_15]   {$\cdots$};
  \node[empty] (z2_25)   [right=of z2_2]    {};
  \node[unobs]   (z2_t)    [right=of z2_25]   {$\z{\tau}^2$};
  \node[empty] (z2_tp05) [right=of z2_t]    {};
  \node[unobs]   (z2_tp1)  [right=of z2_tp05] {$\z{\tau+1}^2$};

  \node[unobs]   (z1_1)    [below=of z2_05]   {$\z{1}^1$};
  \node[empty] (z1_15)   [below=of z2_1]    {};
  \node[empty] (z1_2)    [below=of z2_15]   {$\cdots$};
  \node[empty] (z1_25)   [below=of z2_2]    {};
  \node[unobs]   (z1_t)    [below=of z2_25]   {$\z{\tau}^1$};
  \node[empty] (z1_tp05) [below=of z2_t]    {};
  \node[unobs]   (z1_tp1)  [below=of z2_tp05] {$\z{\tau+1}^1$};
  \node[empty] (z1_tp15) [below=of z2_tp1]  {};

  \node[obs]   (x_1)     [below=of z1_15]   {$\x{1}$};
  \node[obs]   (x_t)     [below=of z1_tp05] {$\x{\tau}$};
  \node[obs]   (x_tp1)   [below=of z1_tp15] {$\x{\tau+1}$};

  \node[obs]   (a_1)     [above=of z2_15]   {$\a{1}$};
  \node[empty] (a_2)     [above=of z2_25]   {};
  \node[obs]   (a_t)     [above=of z2_tp05] {$\a{\tau}$};
}
  \path (z2_1) edge [gen] (z2_2)
        (z2_2) edge [gen] (z2_t)
        (z2_t) edge [gen] (z2_tp1)

        (z2_1) edge [gen] (x_1)
        (z2_t) edge [gen] (x_t)
        (z2_tp1) edge [gen] (x_tp1)

        (z1_1) edge [gen] (z2_1)
        (z1_t) edge [gen] (z2_t)
        (z1_tp1) edge [gen] (z2_tp1)

        (z2_1) edge [gen] (z1_2)
        (z2_2) edge [gen] (z1_t)
        (z2_t) edge [gen] (z1_tp1)

        (z1_1) edge [gen] (x_1)
        (z1_t) edge [gen] (x_t)
        (z1_tp1) edge [gen] (x_tp1)

        (a_1) edge [gen] (z2_2)
        (a_1) edge [gen] (z1_2)
        (a_2) edge [gen] (z2_t)
        (a_2) edge [gen] (z1_t)
        (a_t) edge [gen] (z2_tp1)
        (a_t) edge [gen] (z1_tp1)

        (x_1) edge [inf, bend right=-22.5] (z1_1)
        (a_1) edge [inf] (z1_2)
        (z2_1) edge [inf] (z1_2)
        (x_t) edge [inf, bend right=-22.5] (z1_t)
        (a_2) edge [inf] (z1_t)
        (z2_2) edge [inf] (z1_t)
        (x_tp1) edge [inf, bend right=-22.5] (z1_tp1)
        (a_t) edge [inf] (z1_tp1)
        (z2_t) edge [inf] (z1_tp1);
\end{tikzpicture}